%% file: main_iclr_2027.tex
\documentclass{article}

\makeatletter
\def\input@path{{iclr2027/}}
\makeatother
\PassOptionsToPackage{table}{xcolor}
\usepackage{iclr2027/iclr2027_conference,times}

\input{iclr2027/math_commands.tex}

\usepackage[utf8]{inputenc}
\usepackage[T1]{fontenc}
\usepackage[colorlinks=true,allcolors=blue]{hyperref}
\usepackage{url}
\usepackage{booktabs}
\usepackage{microtype}
\usepackage{graphicx}
\usepackage{multirow}
\usepackage{amssymb}
\usepackage{mathtools}
\usepackage{amsthm}
\usepackage[capitalize,noabbrev]{cleveref}
\usepackage{wrapfig}

\definecolor{grape}{rgb}{0.43, 0.17, 0.71}
\definecolor{blush}{rgb}{0.87, 0.36, 0.51}
\definecolor{aquua}{HTML}{689d6a}

\newif\ifdraft
\draftfalse

\ifdraft
  \usepackage[textsize=tiny]{todonotes}
  \newcommand{\ak}[1]{\todo[color=blush,size=\footnotesize]{\textbf{AK:} #1}}
  \newcommand{\ya}[1]{\todo[color=grape,size=\footnotesize]{\textbf{YA:} #1}}
  \newcommand{\ka}[1]{\todo[color=yellow,size=\footnotesize]{\textbf{KA:} #1}}
  \newcommand{\nc}[1]{\todo[color=green,size=\footnotesize]{\textbf{NC:} #1}}
  \newcommand{\anna}[1]{{\color{blush} Anna: #1}}
  \newcommand{\yedidia}[1]{{\color{grape} Yedidia: #1}}
  \newcommand{\karteek}[1]{{\color{blue} Karteek: #1}}
  \newcommand{\nicolas}[1]{{\color{green} Nicolas: #1}}
\else
  \newcommand{\ak}[1]{}
  \newcommand{\ya}[1]{}
  \newcommand{\ka}[1]{}
  \newcommand{\nc}[1]{}
  \newcommand{\anna}[1]{}
  \newcommand{\yedidia}[1]{}
  \newcommand{\karteek}[1]{}
  \newcommand{\nicolas}[1]{}
\fi

\newcommand{\ragtruth}{RAGTruth\,}
\newcommand{\hallulens}{HalluLens\,}
\newcommand{\longwiki}{LongWiki\,}
\newcommand{\precisewiki}{PreciseWikiQA\,}
\newcommand{\nonexistent}{NonExistentRefusal\,}

\title{How Do Uncertainty Estimators Align with Hallucination Occurrence in LLMs?}

\author{%
  Yedidia~\textsc{Agnimo}\thanks{Correspondence to yedidia.agnimo@ekimetrics.com.} \\
  Ekimetrics, France \\
  Inria Grenoble \\
  \And
  Anna \textsc{Korba} \\
  CREST-ENSAE, France \\
  \And
  Annabelle \textsc{Blangero} \\
  Ekimetrics, France \\
  \And
  Nicolas \textsc{Chesneau} \\
  Ekimetrics, France \\
  \And
  Karteek \textsc{Alahari} \\
  Inria Grenoble, France \\
}

\iclrfinalcopy
\begin{document}

\maketitle
\lhead{Preprint}

\begin{abstract}
Large language models (LLMs) can hallucinate, but do model uncertainty estimators reliably reveal when they do? Although uncertainty estimators are increasingly used to identify unreliable generations, their relationship with hallucination remains poorly understood and is often taken for granted. We systematically evaluate 46 uncertainty estimators across complementary hallucination settings, covering both intrinsic hallucinations, where outputs are unsupported by the provided context, and extrinsic hallucinations, where claims are unsupported by expected model knowledge. Across four benchmarks and multiple LLMs, we find that the association between uncertainty and hallucination varies widely and is often weak, depending on the hallucination setting and the model under evaluation. These results challenge the use of uncertainty as a direct signal of hallucination and clarify when it provides useful information.
\end{abstract}

\section{Introduction}\label{sec:introduction}
\input{sections/introduction}

\section{Background on uncertainty estimators}\label{sec:background}
\input{sections/background}

\input{sections/datasets_methodology}

\input{sections/evaluation_methodology}

\section{Results}\label{sec:results}

\input{sections/results}

\section{Conclusion}
\input{sections/conclusion}

\subsection*{AI use statement}
LLMs were used as experimental components to generate question--answer pairs and to evaluate answer quality, claim support, and abstention in the HalluLens tasks, as described in \Cref{sec:datasets,sec:evaluation,apx:target_processing}. Separately, generative AI tools helped identify literature during the initial review, understand research papers and code implementations, refactor and review code, improve plot presentation, construct LaTeX tables from CSV/JSON results, and edit prose and mathematical notation. They also helped interpret the reported results and organise the presentation. The authors retain responsibility for the accuracy of the content of the paper.

\subsection*{Ethics statement}
This study evaluates hallucination-related failures using existing human-annotated and generated datasets; the reported experiments do not include a new human-annotation study. The principal risk is overinterpreting a benchmark score as evidence that a model response is safe or correct. Uncertainty can miss unsupported content, and LLM judges can introduce systematic errors. Our conclusions are restricted to the evaluated tasks, models, and target definitions, and do not establish reliability in high-stakes applications. We distinguish reference support, answer quality, and abstention so that improvements on one target are not mistaken for a general guarantee against hallucination.

\subsection*{Reproducibility statement}
\Cref{sec:datasets,sec:evaluation} describe the datasets, response models, data splits, logit recovery and evaluation protocol. \Cref{apx:uncertainty_estimators} provides estimator definitions and implementation details; \Cref{apx:additive_methodology_details} specifies dataset evaluation, reference texts and metric computation. The complementary results include the procedures and evidence for length, judge and target sensitivity (\Cref{apx:additive_existing_results}). The code repository is linked in \Cref{apx:code_availability}. The accompanying code separates heavy GPU-intensive model-dependent score computation from CPU-based evaluation and regeneration of the paper figures and tables, with pinned inputs and configuration files for the latter.

\subsection*{Acknowledgements}
Karteek Alahari was supported in part by the Institute of Information \& Communications Technology Planning \& Evaluation (IITP) grant funded by the Korean Government (MSIT) (No. RS-2024-00457882, National AI Research Lab Project).

\bibliography{references}
\bibliographystyle{iclr2027/iclr2027_conference}

\appendix
\input{appendices/additive_navigation}

\section{Related works}\label{sec:rel_work}
\input{sections/related_works}

\section{Description and implementation details on uncertainty estimators}
\label{apx:uncertainty_estimators}
\input{appendices/ue_methods_detailed}

\section{Additional methodology}\label{apx:additive_methodology_details}
\input{appendices/methodology_detailed}

\section{Additional results}\label{apx:additive_existing_results}
\input{appendices/results_detailed}

\end{document}

%% file: iclr2027/math_commands.tex
\usepackage{amsmath,amsfonts,bm}

\def\eqref#1{equation~\ref{#1}}

\def\1{\bm{1}}
\newcommand{\train}{\mathcal{D}}

\newcommand{\test}{\mathcal{D_{\mathrm{test}}}}

\def\vone{{\bm{1}}}
\def\vmu{{\bm{\mu}}}
\def\vtheta{{\bm{\theta}}}

\def\ve{{\bm{e}}}

\def\vu{{\bm{u}}}
\def\vv{{\bm{v}}}

\def\mC{{\bm{C}}}
\def\mD{{\bm{D}}}
\def\mE{{\bm{E}}}

\def\mI{{\bm{I}}}
\def\mJ{{\bm{J}}}
\def\mK{{\bm{K}}}
\def\mL{{\bm{L}}}

\def\mS{{\bm{S}}}

\def\mW{{\bm{W}}}

\def\mSigma{{\bm{\Sigma}}}

\DeclareMathAlphabet{\mathsfit}{\encodingdefault}{\sfdefault}{m}{sl}
\SetMathAlphabet{\mathsfit}{bold}{\encodingdefault}{\sfdefault}{bx}{n}

\newcommand{\R}{\mathbb{R}}

\newcommand{\softmax}{\mathrm{softmax}}

\newcommand{\KL}{D_{\mathrm{KL}}}

\DeclareMathOperator{\Tr}{Tr}

%% file: sections/introduction.tex
Large language models (LLMs) have demonstrated strong performance across a wide range of natural language generation tasks \citep{zhao_survey_2025}. However, fluent and coherent responses are not necessarily supported by the evidence the model is expected to rely on. Such unsupported statements, commonly referred to as hallucinations, remain an important obstacle to the reliable deployment of LLMs \citep{bang-etal-2025-hallulens,fan_halluhard_2026}, particularly in settings where errors can have serious consequences \citep{bengio_international_2025,sahoo_comprehensive_2024}.

Assessing hallucination first requires specifying the relevant source of evidence. Recent work distinguishes hallucination from factuality, which concerns correctness relative to real-world truth \citep{bang-etal-2025-hallulens,augenstein_factuality_2024}. Following \citet{bang-etal-2025-hallulens}, we distinguish intrinsic hallucinations, where model responses are unsupported by information supplied at inference time, from extrinsic hallucinations, where responses are unsupported by knowledge the model is expected to have encountered during pretraining. These notions are related but not equivalent: a statement can be factually correct while remaining unsupported by the context it was expected to use. This distinction matters when evaluating methods intended to identify hallucinated responses.

Substantial effort has been devoted to uncertainty estimation (UE) for language models, with the aim of quantifying confidence in generated responses \citep{vashurin-etal-2025-benchmarking,kang2025uncertainty}. Existing methods rely on different sources of information, including token probabilities and likelihoods \citep{fomicheva2020unsupervised,malinin2020uncertainty}, variability across sampled generations \citep{kuhn2023semantic,lin2023generating}, internal model representations, and model self-evaluation \citep{kadavath2022language,tian_just_2023}. In practice, such signals can support selective prediction or abstention, route uncertain responses to additional verification, or trigger retrieval and refinement mechanisms \citep{abbasi-yadkori_believe_2024,yang_uncertainty-aware_2023,yao_learning_2024,moskvoretskii_adaptive_2025}.

Although uncertainty estimators are not specifically designed to detect hallucinations, they are widely used for this purpose. Yet, the extent to which their scores reflect hallucination remains insufficiently understood. Most prior evaluations assess uncertainty against broader notions of correctness, generation quality, robustness, or calibration rather than against explicit hallucination criteria \citep{vashurin-etal-2025-benchmarking,bakman-etal-2025-reconsidering,wang_ubench_2025}. Moreover, uncertainty estimators capture different signals, and their ability to detect one type of failure does not necessarily extend to other forms of hallucination. Recent work has also shown that estimator comparisons are sensitive to evaluation choices, including correctness functions, and to confounders such as response length  \citep{santilli_revisiting_2025,ielanskyi_addressing_2025}. Consequently, existing results provide limited insights into when and under what conditions uncertainty scores reliably indicate hallucination.

In this paper, we directly evaluate the relationship between uncertainty estimators and hallucination. We compare 46 estimators across 4 datasets (tasks) and 3 open-weight instruction-tuned models. The tasks cover context faithfulness in retrieval-augmented generation (intrinsic) and extrinsic hallucinations, encompassing short-form question answering, long-form generation, and abstention on false premises. This design allows us to examine not only whether uncertainty is associated with hallucination, but also how this association changes across task types and models.

Our study makes three main contributions. First, we systematically evaluate a broad set of uncertainty estimators for hallucination detection across intrinsic and extrinsic hallucination settings. Second, we characterize the relationship between uncertainty and hallucination across tasks and models. We find substantial variation in this relationship: uncertainty provides a strong signal across multiple estimators in some settings, whereas in others, it is useful only for a smaller subset of estimators or varies considerably across models. Third, we investigate estimator selection beyond individual performance by examining the stability of estimator rankings across tasks and models and the relationships between competitive uncertainty scores. We find that estimator rankings vary substantially across settings, although several poorly performing estimators consistently rank near the bottom. Moreover, some competitive estimators produce strongly correlated response rankings despite differing computational requirements. Together, these findings show that the utility of uncertainty for hallucination detection depends on the evaluation setting and highlight the importance of evaluating estimators under the specific conditions in which they are intended to be used.

%% file: sections/background.tex
 \paragraph{Notation.}
We consider an autoregressive language model $p_{\vtheta}(y \mid x)$, where $x$ denotes the input (query and optional context), $y = (y_1,\dots,y_T)$ is a generated response sequence, and $\vtheta$ parametrizes the model. At each step $t$, the model produces a token-level probability distribution $p_{\vtheta}(\cdot \mid x,y_{<t})$, so that $p_{\vtheta}(y \mid x)=\prod_{t=1}^T p_{\vtheta}(y_t \mid x,y_{<t})$ for a response sequence of length $T$, with $y_{<t} = (y_{1},\dots,y_{t-1})$.  %

        \subsection{Uncertainty definition and scope}
There is no universally accepted formal definition of uncertainty in language generation \citep{zhang_luq_2024,bakman-etal-2025-reconsidering}. In this work, uncertainty is treated as a query-level property of the predictive distribution \(p_{\vtheta}(\cdot \mid x)\). A generic query-level target is denoted \(U^\star(x)=\Phi(p_{\vtheta}(\cdot \mid x))\), where \(\Phi\) is a property of the predictive distribution \citep{chen_query-level_2025}. Different estimators may approximate different aspects of this target and need not share the same probabilistic foundation.

Terminology in the literature is not fully consistent: some works use 
\emph{uncertainty} and \emph{confidence} interchangeably \citep{qiu_semantic_2024,zhang_luq_2024}, while others reserve \emph{confidence} for response-level uncertainty $y$ \citep{lin2023generating}. Here, \emph{uncertainty} refers by default to query-level uncertainty. Scores of the form \(U(x)\) target this quantity directly, whereas response-conditioned scores \(U(x,y)\) are treated as single-response proxies. Our empirical evaluation associates each score with the quality of the observed response; it does not estimate \(U^\star(x)\) independently. %

    \subsection{Uncertainty estimators studied} \label{sec:ue_families}
We organize uncertainty estimators by their access requirements: white-box [WB] methods require access to the evaluated model's logits or internal states, whereas black-box [BB] methods operate on generated text alone. Within each access regime, we group estimators by the information used to compute uncertainty and their underlying methodology. Our implementation follows \textsc{LM-Polygraph} \citep{fadeeva2023lm,vashurin-etal-2025-benchmarking}; see detailed definitions in \Cref{apx:uncertainty_estimators}.

\textbf{[WB] Information/logit-based methods} estimate uncertainty from token-level predictive distributions \(p_{\vtheta}(\cdot \mid x,y_{<t})\), without requiring additional sampled responses. They differ in what information they extract from these distributions and how they aggregate it across tokens. The simplest methods, such as Maximum Sequence Probability (MSP), use only the probabilities assigned to generated tokens \citep{fomicheva2020unsupervised}. Other methods exploit the full token distribution, measuring uncertainty through token entropy \citep{fomicheva2020unsupervised} or measures of deviation from the uniform distribution, such as Rényi or Fisher--Rao divergences \citep{darrin2023rainproof}. Pointwise mutual information methods compare conditional and unconditional token probabilities to quantify the influence of the input on the generated response \citep{van_der_poel_mutual_2022}. Finally, some methods incorporate token importance: TokenSAR weights token-level uncertainty by semantic importance \citep{duan_shifting_2024}, while Claim-Conditioned Probability (CCP) assesses uncertainty based on whether alternative tokens preserve or alter the meaning of the current claim \citep{fadeeva_fact-checking_2024}.

\textbf{[WB] Sample-based methods} require multiple responses sampled from the evaluated model, together with their token-level probabilities. They estimate uncertainty from the variability of sampled responses, using either likelihood-based or semantic information. Monte Carlo methods approximate predictive uncertainty through sample-based estimates of entropy or related quantities \citep{malinin2020uncertainty}. Semantic Entropy \citep{kuhn2023semantic} groups responses into semantic equivalence classes and measures uncertainty over these classes, whereas Semantic Density \citep{qiu_semantic_2024} uses continuous pairwise semantic similarities. SentenceSAR \citep{duan_shifting_2024} combines sequence likelihoods with semantic similarities to weight sampled responses, while Cocoa \citep{vashurin_cocoa_2025} estimators combine logit-based uncertainty with semantic variability across samples.

\textbf{[WB] Internal-state methods} require access to the model's attention weights or hidden representations. They estimate uncertainty from internal processing signals rather than solely from output probabilities. AttentionScore uses self-attention patterns \citep{sriramanan_llm-check_2024}, RAUQ models uncertainty propagation across the generated sequence \citep{vazhentsev_uncertainty-aware_2025}, and attention-based saliency methods reweight token probabilities according to their contextual importance \citep{lin_contextualized_2024}.

\textbf{[WB] Training-based methods} require internal representations and a separate dataset to fit a reference distribution. They estimate uncertainty by measuring how far an input or response representation lies from the fitted distribution, typically using Gaussian density models and Mahalanobis distance \citep{vazhentsev_hybrid_2023,lee2018simple}. Unlike the preceding methods, they require a fitting stage before evaluation.

\textbf{[WB] Reflexive methods} estimate uncertainty from the model's explicit assessment of its own responses, typically elicited through a follow-up prompt. For example, $P(\mathrm{True})$ uses the probability assigned to a positive answer when the model is asked to assess the correctness of its generated response \citep{kadavath2022language}. Such methods rely on the model's self-assessment rather than directly aggregating the original response's predictive probabilities.

\textbf{[BB] Black-box methods} require only generated text, without access to the evaluated model's logits or internal states. Most rely on multiple sampled responses and estimate uncertainty from their lexical or semantic agreement \citep{lin2023generating,nikitin_kernel_2024}. Pairwise similarities, computed using lexical overlap or semantic similarity measures such as natural language inference, are aggregated into uncertainty scores through clustering, graph-based statistics, or spectral analysis. Unlike white-box methods, they can be applied without access to the evaluated model's internal computations.

%% file: sections/datasets_methodology.tex
\section{Benchmark setup}

\subsection{Datasets}\label{sec:datasets}

Our evaluation covers both intrinsic hallucinations, assessed against evidence provided with the input, and extrinsic hallucinations, assessed against knowledge expected to be available from pretraining. We consider four tasks targeting different aspects of hallucination: context faithfulness, short-answer correctness, factual support in long-form generation, and abstention on false premises. Each task defines a response-level quality target \(Q(x,y)\) based on its specific evaluation criterion. We use these targets to assess whether higher uncertainty is associated with lower response quality.

{\bf Intrinsic hallucination: \ragtruth.} \ragtruth \citep{niu-etal-2024-ragtruth} evaluates context faithfulness across summarization, passage-grounded question answering, and data-to-text generation. Human annotators identify response spans that conflict with or are unsupported by the provided context. We use the processed release\footnote{\href{https://huggingface.co/datasets/wandb/RAGTruth-processed}{huggingface.co/datasets/wandb/RAGTruth-processed}} and define \(Q(x,y)=0\) if the response contains at least one annotated hallucinated span, and \(Q(x,y)=1\) otherwise. Under this criterion, a statement may be considered hallucinated even if it is factually correct, provided that it is not supported by the supplied context.

{\bf Extrinsic hallucination: \hallulens.} We consider three tasks from \hallulens \citep{bang-etal-2025-hallulens}, which use Wikipedia as a proxy for knowledge expected to be available from pretraining and synthetic nonexistent entities to evaluate abstention when the requested information does not exist. We follow the HalluLens evaluation procedures, using Llama-3.1-70B-Instruct\footnote{\texttt{hugging-quants/Meta-Llama-3.1-70B-Instruct-AWQ-INT4}.} as the judge. Label processing, quality formulas and evaluated response counts are provided in \Cref{apx:target_processing}.

\textbf{PreciseWikiQA} evaluates short-answer correctness in closed-book question answering. Questions and reference answers are generated from sampled Wikipedia pages using an LLM. The judge assesses the correctness of each response against its reference answer. We retain responses with a valid correctness judgment and no abstention flag, assigning \(Q(x,y)=1\) to correct answers and \(Q(x,y)=0\) to incorrect answers. 
\textbf{LongWiki} evaluates the factual support and quantity of information in long-form responses. An LLM decomposes each response into atomic claims, which are then evaluated against retrieved Wikipedia passages using the judge. We measure response quality using F1@$K$, the harmonic mean of claim-level precision and recall@$K$. Precision measures the proportion of generated claims that are supported by the reference evidence, while recall@$K$ measures the number of supported claims relative to a fixed quota of $K$ claims. Following \citet{bang-etal-2025-hallulens}, we use $Q(x,y)=\mathrm{F1@}K$, with $K=32$ throughout our experiments. Further details are provided in \Cref{apx:target_processing}. %
\textbf{NonExistentRefusal} evaluates whether a model abstains when asked about a nonexistent entity. The judge assesses whether the response rejects the false premise, assigning $Q(x,y)=1$ when the model abstains and $Q(x,y)=0$ when it accepts the premise. The quality target therefore measures failure to abstain rather than directly identifying fabricated statements.

\subsection{Choice of models and computation of uncertainty estimators}

{\bf Models.}
The model selection is constrained by \ragtruth, which provides human-annotated responses generated by six language models \citep{niu-etal-2024-ragtruth}. We retain the three small open-weight models: \texttt{Mistral-7B-Instruct-v0.2}, \texttt{Llama-2-7b-chat-hf}, and \texttt{Llama-2-13b-chat-hf}. Their open weights provide access to token-level probabilities, enabling the evaluation of white-box uncertainty estimators under a common protocol.

{\bf Recovering token logits for annotated \ragtruth responses.} 
Because \ragtruth provides responses generated during dataset construction and subsequently annotated for hallucinations, we must evaluate uncertainty estimators on these original responses. Regenerating responses would break their correspondence with the human annotations. However, the dataset does not provide token logits, which are required to compute information-based and certain white-box sampling-based estimators. We therefore recover the necessary token-level probabilities by scoring each annotated response under the corresponding causal language model using teacher forcing \citep{williams_learning_1989}. Given an input--response pair $(x,y)$, we run the model on the concatenated sequence $x \oplus y$ and extract the logits at the response positions. This yields the autoregressive conditional distributions $p_{\vtheta}(\cdot \mid x,y_{<t})$, including the probabilities $p_{\vtheta}(y_t \mid x,y_{<t})$ assigned to the observed tokens, which are used to compute the uncertainty estimators on the annotated responses.

{\bf Dataset splits.} 
Following \citet{vashurin-etal-2025-benchmarking}, \ragtruth is partitioned into $1000$ training and $2000$ evaluation queries, keeping all responses associated with a query in the same split. Each \hallulens dataset has $1000$ training and $2000$ evaluation examples. The training split is used only by estimators requiring fitting or training-set statistics. Summary statistics are reported in \Cref{table:dataset_statistics}.

\input{tables/dataset_summary_statistics}

{\bf Response-length baseline.} We include generated token count as a baseline to assess whether uncertainty estimators provided additional predictive information beyond response length alone.

{\bf Auxiliary reference-based score: AlignScore.} AlignScore AlignScore \citep{zha2023alignscore} is a trained text-pair model that estimates whether one text is supported by another. We use it as a comparator to contextualize uncertainty estimator performance against a method with access to external reference information. Reference choices and scoring directions are detailed in \Cref{apx:alignscore_comparator}.

%% file: tables/dataset_summary_statistics.tex
\begin{table*}%
\caption{Summary statistics of the benchmark datasets. Token counts with \texttt{Llama-2}'s tokenizer.
$^\ast$For \textsc{RAGTruth}, the reported target length corresponds to the average generated response length. \textsc{NonExistentRefusal} has no reference target answer because the desired behavior is abstention.}
\label{table:dataset_statistics}
\centering
\small
\resizebox{0.9\textwidth}{!}{%
\begin{tabular}{lccc}
\toprule
\textbf{Dataset} & \textbf{\begin{tabular}[c]{@{}c@{}}Train + eval  queries\end{tabular}}  
& \textbf{\begin{tabular}[c]{@{}c@{}}Avg. document  length\end{tabular}}  
& \textbf{Avg. target length} \\
\midrule
\textsc{PreciseWikiQA}      & $1000 + 2000$ & 37.9  & 5.8   \\
\textsc{LongWiki}           & $1000 + 2000$ & 38.4  & 235.0 \\
\textsc{NonExistentRefusal} & $1000 + 2000$ & 17.6  & --    \\
\textsc{RAGTruth}           & $1000 + 1960$ & 698.2 & 194.4$^\ast$ \\
\bottomrule
\end{tabular}%
}
\end{table*}

%% file: sections/evaluation_methodology.tex
\section{Evaluation methodology}
\label{sec:evaluation}

Let $\test=\{(x_j,y_j)\}_{j=1}^n$ denote an evaluation set, where $x_j$ is the input and $y_j$ the model response. Each dataset defines a response-level quality target $Q(x_j,y_j)$, with higher values indicating better responses (\Cref{sec:datasets}). Each uncertainty estimator produces a scalar score $u_j$ for the same instance. We assess whether higher uncertainty is associated with lower quality through discrimination, selective prediction, and rank calibration, then compare estimator rankings and the information their scores capture.
For each panel, we estimate metric variability using 1,000 bootstrap resamples and report the resulting standard deviation. For paired AUROC differences, we report 95\% percentile confidence intervals using identical resamples for both estimators (\Cref{apx:additive_length}).

\subsection{Detection and selective prediction}

{\bf Discriminating errors from supported responses.} The area under the receiver operating characteristic curve (AUROC) measures the probability that an error receives higher uncertainty than a non-error; a random estimator has an expected AUROC of 0.5, and 1 indicates perfect discrimination. We define errors by $Q=0$ for binary targets and $\mathrm{F1@32}<0.5$ for LongWiki. Sensitivity to the F1@32 threshold is examined in \Cref{apx:longwiki_components}; continuous $\mathrm{F1@32}$ is retained for selective prediction and rank calibration. %
We also report the area under the precision--recall curve (AUPRC) to complement AUROC when errors and non-errors are imbalanced. As responses are selected from highest to lowest uncertainty, precision is the fraction of selected responses that are errors, and recall is the fraction of all errors selected. A constant score yields AUPRC equal to error prevalence. %

{\bf Selective prediction.} A complementary axis evaluates whether rejecting high-uncertainty responses improves the quality of those retained. Responses are ordered from lowest to highest uncertainty. At rejection rate $r$, we retain the lowest-uncertainty fraction $1-r$ and compute its mean quality, defining a quality--retention curve. The oracle instead orders responses by decreasing quality, perfectly rejecting the lowest-quality responses first. Following prior work \citep{malinin2020uncertainty,vashurin-etal-2025-benchmarking}, we summarize the curve with the prediction--rejection ratio (PRR): \[ \mathrm{PRR}=\frac{\mathrm{AUC}_{U}-\mathrm{AUC}_{\mathrm{random}}} {\mathrm{AUC}_{\mathrm{oracle}}-\mathrm{AUC}_{\mathrm{random}}}. \] Here $\mathrm{AUC}_{U}$, $\mathrm{AUC}_{\mathrm{random}}$, and $\mathrm{AUC}_{\mathrm{oracle}}$ are the areas under the quality--retention curves for the uncertainty estimator, a random estimator, and the oracle, respectively. All three use rejection rates from 0 to 1. PRR measures the proportion of the oracle's improvement over random rejection that the uncertainty estimator achieves: 1 corresponds to oracle-level performance, 0 to random rejection, and negative values to worse-than-random performance.

{\bf Rank calibration.} Discrimination and selective prediction assess separability and rejection utility; rank calibration asks whether uncertainty is consistently ordered against expected response quality. Rank-calibration error (RCE; \citealp{huang_uncertainty_2024}) provides a rank-based diagnostic of this alignment. It compares scores on different scales without requiring them to be probabilities or binarizing continuous quality targets. Lower RCE is better. %

\subsection{Ranking stability and estimator redundancy}
After evaluating individual performance, we examine two distinct questions: whether estimator rankings remain stable across evaluation settings, and whether different estimators assign similar rankings to the same responses.

{\bf Ranking stability across settings.} For each dataset--model pair, referred to as a panel, we rank estimators by their evaluation performance. We then compare these rankings between panels using Kendall's $\tau$, which measures agreement in the ordering of estimator pairs. Holding the model fixed assesses transfer across datasets; holding the dataset fixed assesses stability across models.

{\bf Estimator-level correlation.} Within each panel, we compute pairwise Spearman correlations between uncertainty scores to measure how similarly estimators rank the same responses. We average these correlations across panels and apply hierarchical clustering to identify groups with similar behavior. This analysis concerns overlap in response rankings; it does not establish whether combining estimators would improve detection.

%% file: sections/results.tex
We report results on four datasets covering different hallucination types, \ragtruth (RT), \precisewiki (PWQA), \longwiki (LW), and \nonexistent (NR), and three models: \texttt{Mistral-7B-Instruct-v0.2} (M7B), \texttt{Llama-2-7B-chat} (L7B), and \texttt{Llama-2-13B-chat} (L13B). This yields 12 dataset-model pairs, referred to as \emph{panels}. Unless stated otherwise, estimator rankings exclude AlignScore variants, which are used as reference-based comparators. We use AUROC to compare detection performance, then examine how rankings vary across settings, whether competitive scores are redundant, and how the comparisons depend on targets and controls.

\begin{figure}[htbp]
    \centering
    \includegraphics[scale=0.8]{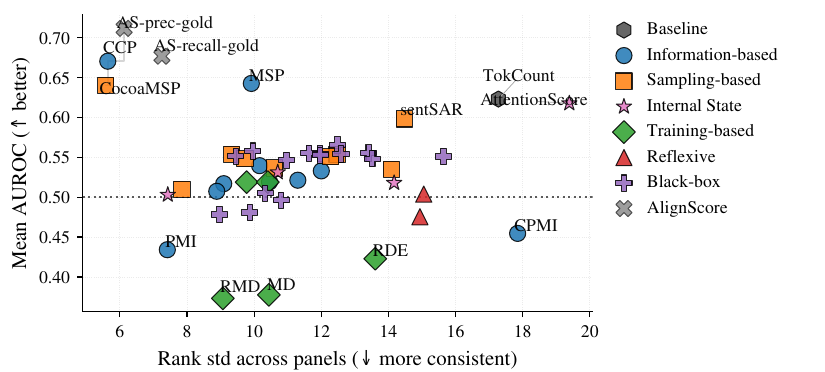}
    \caption{Mean AUROC against rank variability across
    the 12 panels. Each point represents an uncertainty estimator, the token-count baseline,
    or an AlignScore variant. Rank variability is the standard deviation of an estimator's within-panel rank across panels; lower values indicate a more stable cross-panel ranking. AlignScore variants are reference-based comparators.}
    \label{fig:stability_perf_vs_rank_std_pooled}
\end{figure}

{\bf Average performance and consistency across panels.} 
\Cref{fig:stability_perf_vs_rank_std_pooled} compares each estimator's mean AUROC across panels with the variability of its within-panel rank. Estimators in the upper part of the plot discriminate hallucinations more strongly on average; estimators on the left rank more consistently across panels. CCP (logit/information-based) and CocoaMSP (white-box sample-based) provide the best compromise between the two criteria. MSP (logit/information-based) and AttentionScore (internal-state) achieve comparable or higher mean AUROC but exhibit greater rank variability, indicating that their averages are driven by panels where they peak rather than by uniform performance across panels. Uncertainty is informative on some panels, but no estimator is uniformly reliable. We next compare rank variation across datasets and models. %

{\bf The dataset is the primary source of heterogeneity in estimator rankings.} Estimator rankings vary more with dataset changes than with model changes. Kendall's $\tau$ heatmaps in \Cref{apx:ranking_transfer} (see appendix) quantify this pattern: when the model is fixed and the dataset changes, mean estimator-ranking agreement is low ($\tau = 0.09$ to $0.13$); when the dataset is fixed and the model changes, agreement is higher but uneven, with stronger transfer on LW ($\tau = 0.71$) and PWQA ($\tau = 0.66$) than on RT ($\tau = 0.34$) and NR ($\tau = 0.30$).

\begin{figure}[htbp] 
    \centering 
    \includegraphics[scale=0.8]{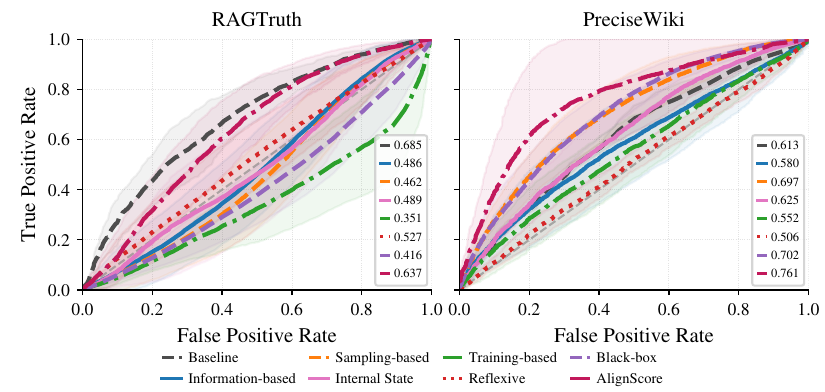}
    \caption{Family-level ROC aggregates per task, averaged across the three generators. For each task--model panel, estimator ROC curves are first averaged within each family by taking the mean true positive rate at each false positive rate. For each task, these model-level family means are then averaged across models. Shaded bands show one standard deviation of these model-level family-mean true positive rates at each false positive rate; they reflect cross-model variability.} 
    \label{fig:family_roc_agg_per_task} 
\end{figure}

Family-level comparisons summarize this difference across groups of related estimators.
\Cref{fig:family_roc_agg_per_task} reports family-level predictive performance of uncertainty estimators for two contrasting hallucination settings: PWQA, where most estimators exhibit moderate to strong discriminative power, and RT, where it is sparsest (see \Cref{sec:ue_families} for the family definitions). On PWQA, all six uncertainty estimator families lie above the $0.5$ baseline of an uninformative score, with the information-based and white-box sample-based families reaching the highest family-aggregate AUROCs. In contrast, family-level ROC curves on RT sit closer to the $0.5$ baseline, suggesting that uncertainty estimators %
carry limited information for discriminating hallucinations in this setting. The two remaining datasets, LW and NR, are reported in \Cref{fig:family_roc_agg_per_task_lw_nr} (see appendix).

{\bf Model differences still affect which estimators perform best.} Although task differences dominate overall, the model still changes which uncertainty signal is most informative on a given dataset. On RT with M7B, the leading estimators include the information-based MSP and CCP, as well as the white-box, sample-based CocoaMSP (see \Cref{tab:metrics_mistral_7b_instruct} in appendixev). This pattern does not persist on the Llama models. On RT with L13B, by contrast, the output-length baseline is the strongest signal (AUROC $0.75$), followed by AttentionScore ($0.69$) and the reflexive estimators, in which the language model is prompted to assess its own uncertainty (Black-box PTrue $0.67$, PTrue $0.63$); the reflexive estimators are uninformative on the other models. Thus, even within a fixed task setting, estimator choice remains model-dependent. %

{\bf Some families consistently perform poorly.} The bottom of the panel ranking in \Cref{fig:stability_perf_vs_rank_std_pooled} is more stable than the top across datasets and models (see \Cref{fig:stability_auroc_across_models_by_task} in the appendix). The training-based density estimators Mahalanobis Distance (MD), Relative MD, and Robust Density Estimation (RDE) occur among the bottom five uncertainty estimators in 10, 10, and 7 of the 12 panels, respectively, with the pointwise mutual information variants (PMI and CPMI) also frequently near the bottom. A plausible interpretation is that density-based estimators measure how typical a generated response is under a fitted reference representation distribution. Since hallucinated responses are still sampled from the generator, they need not be atypical in representation space. Performance may also depend on modeling choices such as the training data, background corpus, representation layer and dimensionality reduction.  These results suggest that training-based density estimators and PMI/CPMI should not be used as default uncertainty signals without dataset-specific validation.

\begin{wrapfigure}[21]{r}{0.5\textwidth}
\vspace{-1em}
    \centering
    \includegraphics[width=\linewidth]{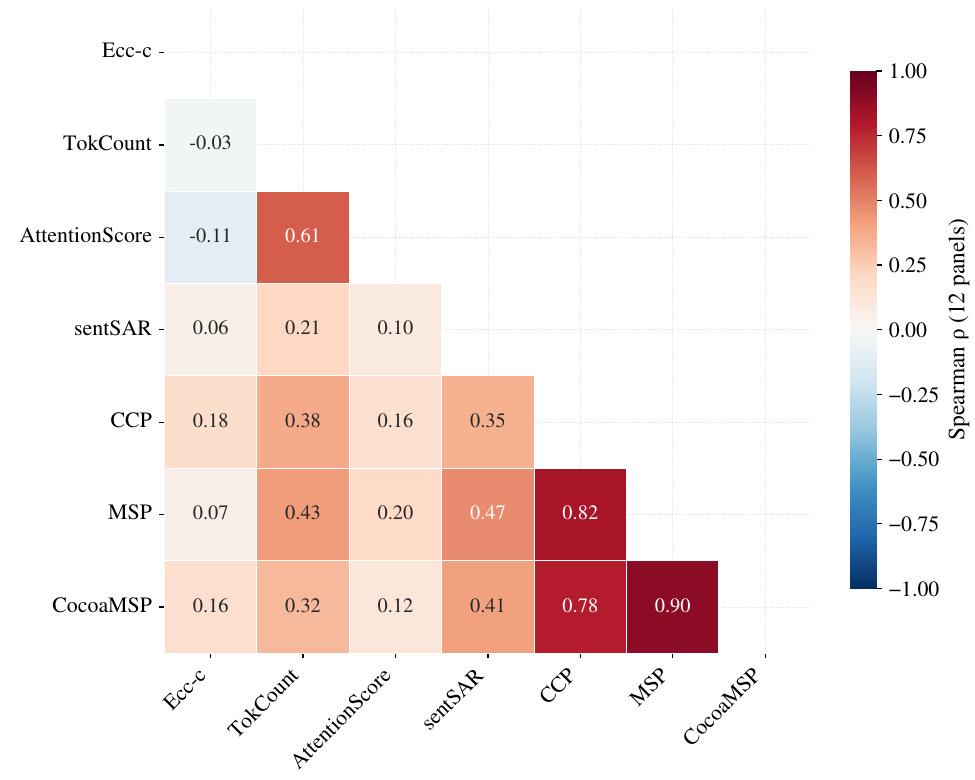}
    \caption[Spearman correlation between the top uncertainty estimators across panels]{Spearman correlation between the top uncertainty estimators across panels.}
    \label{fig:robust_top_redundancy}
\end{wrapfigure}

{\bf The recurrent top estimators form three correlated ranking clusters.} 
\Cref{fig:robust_top_redundancy} reports Spearman's $\rho$ correlations between the seven estimators with the highest mean AUROC, averaged across panels. This figure shows three correlated ranking clusters. The first cluster is logit- and sample-based, and relies on logits: SentenceSAR, Maximum Sequence Probability (MSP), Claim-Conditioned Probability (CCP), and CocoaMSP. The strongest correlations occur between MSP, CCP, and CocoaMSP ($\rho \geq 0.78$), while SentenceSAR is more weakly linked to the others.  This correlation is informative because these estimators differ in construction: MSP is a direct sequence-likelihood score; CCP filters local token alternatives through a natural-language inference model; CocoaMSP modulates an MSP-derived score using inter-sample semantic dissimilarity; and SentenceSAR combines token-level and sentence-level relevance signals across sampled responses. 
The second cluster consists of Token count and AttentionScore, which are correlated at $\rho=0.61$, while their correlations with the probability-based cluster are lower. AttentionScore uses attention weights rather than token probabilities or comparisons between sampled responses. Its correlation with token count may be due to the fact that the scores sum attention token contributions. Correlation alone does not establish that length explains its detection performance or that combining different scores would improve detection. The third is reduced to Eccentricity (Ecc-c), which is weakly correlated with the other displayed scores. It is a black-box graph-based estimator that measures dispersion through semantic contradiction between sampled responses.

\vspace{7pt}
{\bf Access constraints should shape estimator selection.} The ranking clusters differ not only in performance, but also in input requirements. The first one mostly contains white-box estimators that require token probabilities. CocoaMSP and SentenceSAR further require sampled generations and a sentence-similarity model, while AttentionScore requires access to attention weights. In contrast, the NLI-graph estimators are black-box with respect to the generator: they use sampled text responses and an auxiliary NLI model, but do not require generator logits, hidden states, or attention weights. When white-box access is available, CocoaMSP and CCP perform better across datasets, while MSP remains an attractive, computationally efficient baseline.

\section{Sensitivity and controls}

In this section, we investigate the sensitivity of our analysis with respect to several factors. 

{\bf %
On the correlation between uncertainty estimates and response length.} We perform a length-sensitivity analysis on RT, because this is the only dataset with strong heterogeneity of response length, unlike LW and NR. 
On RT, longer responses hallucinate more often: the hallucination rate rises from $0.27$ in the shortest response-length quintile to $0.85$ in the longest for L13B. Hence, the number of tokens is a relevant predictor of hallucination. Yet, within quintiles, its performance is close to chance (mean AUROC $0.54$ to $0.57$). %
We then test the impact of the response length on the estimators we study. 
The estimators behave differently with respect to this control. For instance, MSP and CCP remain informative across all five M7B quintiles ($0.64$ to $0.75$) but fall near or below chance on both Llama models. AttentionScore performance improves with length on L13B ($0.50$ to $0.77$ on the different quintiles) while M7B shows no such trend. The complete strata and paired comparisons can be found in \Cref{apx:additive_length}. We do not find a recurring dependence of uncertainty estimators with response length. 
Hence, we conclude that the aggregated results presented in \Cref{sec:results} may hide some length-dependent behavior, yet length by itself does not explain the performance of all estimators. 

{\bf %
Impact of the judge model.} We perform a sensitivity analysis of our study with respect to the LLM used as a judge to label hallucinations in the HalluLens datasets. Recall that this hallucination label is in $\{0,1\}$ for NER and PW, while for LW (because of the response length), this labeling is done at the claim-level (a response being divided in several claims), resulting in an averaged score per response in $[0,1]$. On NER and PW, we find a strong agreement between labeling of the two models, and consequently on the uncertainty estimators results. On LW, where responses are longer, we find a difference of labeling between the two models (with Qwen tagging fewer answers as hallucinated), yet the results of the uncertainty estimators are not affected. This is because the relative ranking of the answers in terms of quality is preserved. 
Thus, the comparison between uncertainty estimators is stable with respect to the choice of the LLM used as a judge. %
The protocols and detailed results can be found in \Cref{apx:judge_results}.

{\bf %
Impact of the hallucination type.} In our study, we generally only have the information that a response is hallucinated or not, except on RT, where the dataset additionally incorporates information on the type of hallucination. More specifically, it can be tagged as `baseless-information' and/or `evidence-conflict' with respect to the provided evidence (context). We therefore performed a sensitivity analysis on the RT dataset with respect to hallucination type.
We find that rankings of the uncertainty estimators' performance remain very similar between the two hallucination types (they are highly correlated), although the best method can differ. Interestingly, the best discrimination performances (the highest AUC) are achieved when considering cases where both subtypes co-occur as the hallucination target.
The complete results are reported in \Cref{apx:review_subtype_checks}.

%% file: sections/conclusion.tex
Our study clarifies when existing uncertainty estimation methods align with hallucination-related output quality in large language models. The main finding is that uncertainty is informative, but only conditionally. Task-specific evaluation targets are the main source of heterogeneity, which argues against treating any uncertainty estimator as a universal hallucination detector. 
This leaves the question of which uncertainty estimator should be favored in practice. This choice is actually not only guided by practical use based on the target task and validation performance, but also access to the generator internals and the computational budget.  

For the small open-weight models studied here, our results provide practical starting points based on the level of access to the generator. When only sampled outputs are available, contradiction-based graph estimators, including Eccentricity variants, are useful candidates for short-answer correctness and supported long-form quality, but perform less well on context faithfulness and refusal. When token probabilities are also available, MSP provides a strong baseline, while CCP is competitive for short-answer correctness and refusal. When internal attention information is accessible, AttentionScore is a promising candidate for context faithfulness, although its refusal performance is more model-dependent and should be compared against token count. 
Score correlations further help identify overlapping candidates before investing in more costly methods. These findings guide which estimators to validate, rather than establish a universally best method or a validated ensemble. %

\section*{Limitations and future work}
Our conclusions concern three open-weight models and four tasks. In our study, the only dataset with human-annotated hallucinations is RAGTruth. However, this dataset provides these labels only for a fixed set of model responses: MISTRAL7B, LLAMA7B, and LLAMA13B. Hence, because our motivation was to investigate the performance of the widest set of uncertainty estimators across several hallucination settings, we kept this set of models as the evaluated models across alternative hallucination tasks (hallucination datasets). Thus, to conduct a comparative study, we constrained the set of evaluated models. 
Then, the hallucination tasks we consider cover context faithfulness, short-answer correctness, supported claims in long-form generation, and failure to reject false premises. They do not cover every form of model error. %

Future work could extend our evaluation in two directions. First, hallucination targets chould be made more fine-grained. For instance, span-level annotations, such as those available in RAGTruth, can be used to test whether uncertainty scores indicate where hallucinations occur in the answer, rather than only whether a response is unreliable. Second, extension to other datasets and larger and more recent models may strengthen the analysis.

%% file: appendices/additive_navigation.tex
\section*{Appendix contents}
\noindent
\hyperref[sec:rel_work]{A. Related works}\hfill\pageref*{sec:rel_work}\\
\hyperref[apx:uncertainty_estimators]{B. Uncertainty estimators}\hfill\pageref*{apx:uncertainty_estimators}\\
\hyperref[apx:additive_methodology_details]{C. Additional methodology}\hfill\pageref*{apx:additive_methodology_details}\\
\hspace*{1em}\hyperref[apx:target_processing]{Datasets and response evaluation}\hfill\pageref*{apx:target_processing}\\
\hspace*{1em}\hyperref[apx:alignscore_comparator]{AlignScore reference texts}\hfill\pageref*{apx:alignscore_comparator}\\
\hspace*{1em}\hyperref[apx:evaluation_conventions]{Evaluation metrics}\hfill\pageref*{apx:evaluation_conventions}\\
\hyperref[apx:additive_existing_results]{D. Additional results}\hfill\pageref*{apx:additive_existing_results}\\
\hspace*{1em}\hyperref[apx:performance_tables]{Complete performance tables}\hfill\pageref*{apx:performance_tables}\\
\hspace*{1em}\hyperref[apx:ranking_transfer]{Agreement between estimator rankings}\hfill\pageref*{apx:ranking_transfer}\\
\hspace*{1em}\hyperref[apx:performance_summaries]{Performance by task and estimator family}\hfill\pageref*{apx:performance_summaries}\\
\hspace*{1em}\hyperref[apx:additive_length]{Response length}\hfill\pageref*{apx:additive_length}\\
\hspace*{1em}\hyperref[apx:judge_results]{Changing the judge}\hfill\pageref*{apx:judge_results}\\
\hspace*{1em}\hyperref[apx:review_subtype_checks]{RAGTruth: error types and proportions}\hfill\pageref*{apx:review_subtype_checks}\\
\hspace*{1em}\hyperref[apx:longwiki_components]{LongWiki: precision, recall and error thresholds}\hfill\pageref*{apx:longwiki_components}
\par\medskip

%% file: sections/related_works.tex
\subsection{Evaluating uncertainty estimation methods}
\label{sec:rel_work_ue_eval}

Work on evaluating uncertainty estimation (UE) methods for LLMs has addressed three concerns: building infrastructure that places heterogeneous methods on a common framework; assessing whether comparisons drawn under controlled conditions hold up in deployment; and questioning whether the evaluation protocols themselves are sound. Within the third concern, we examine whether the evaluation target reflects the particular response failure that uncertainty is intended to detect.

\paragraph{Method-coverage frameworks.}
The most direct response to method heterogeneity has been to consolidate estimators behind shared interfaces and shared intermediate quantities. \textsc{LM-Polygraph} \citep{vashurin-etal-2025-benchmarking} implements a comprehensive estimator suite behind a single pipeline that computes greedy responses, token probabilities, sampled responses, and semantic relation matrices once and reuses them across estimators, controlling for implementation differences. Adjacent toolkits -- \textsc{UQLM} \citep{bouchard_uqlm_2026}, \textsc{UncertaintyZoo} \citep{wu_uncertaintyzoo_2025}, and \textsc{UBench} \citep{wang_ubench_2025} -- provide complementary coverage or standardize estimator outputs to a common confidence range. Across these frameworks, the contribution is comparability rather than validity: each evaluation is only as meaningful as its target for correctness.

\paragraph{Deployment- and protocol-level critiques.}
A second strand asks whether rankings produced under controlled benchmarks transfer to deployment, and whether the protocols themselves can be trusted. \citet{bakman-etal-2025-reconsidering} evaluate nineteen UE methods under decision-threshold drift, prompt perturbation, long-form adaptation, and ensembling, and find that most methods are sensitive to threshold choice and vulnerable to adversarial prompts -- patterns invisible under standard evaluation conditions. \citet{santilli_revisiting_2025} show that the same response length can bias both UE scores and correctness functions so that a trivial length baseline can rival principled methods under length-biased correctness metrics. \citet{ielanskyi_addressing_2025} reinforce this concern by showing that the choice of approximate correctness function (ROUGE, BLEURT, LLM-as-judge variants) substantially reorders method rankings, and propose marginalising over multiple judges as a more robust alternative. Complementary critiques target the limited ecological validity of QA-style evaluations \citep{devic_calibration_2025}, the inability of realisable benchmarks to test calibration over inherently variable outcomes \citep{cruz_evaluating_2024,tomov_illusion_2026}, and the incomparable scales on which estimator outputs live \citep{huang_uncertainty_2024}. The shared message is that ``method $A$ outperforms method $B$ on benchmark $X$'' is routinely confounded by length, choice of correctness function, threshold calibration, or estimator output scale -- and that fixing the benchmark itself does not fix the comparison.

\paragraph{The evaluation gap we address.}
Across these strands, a common limitation is that the evaluation \emph{target} is rarely a hallucination-specific signal: correctness is operationalised through lexical or semantic overlap with a reference answer, or through a multiple-choice answer key, rather than through an explicit characterisation of when a response is unsupported by the evidence the model was supposed to use. Our contribution sits in this space: we retain the \textsc{LM-Polygraph} estimator catalogue and the methodological caution of \citet{santilli_revisiting_2025}, \citet{bakman-etal-2025-reconsidering}, and \citet{ielanskyi_addressing_2025}, but replace generic correctness targets with hallucination targets defined through operational distinctions established in the hallucination literature, which we develop next.

\subsection{Hallucination: definitions and benchmarks}
\label{sec:rel_work_halluc}

The definition of hallucination has shifted as LLMs have moved from specific tasks to largely open-ended generation, and the choice of definition directly shapes what a benchmark measures.

\paragraph{From task-specific taxonomies to LLM-specific definitions.}
The intrinsic/extrinsic distinction was originally formulated for specific natural language generation tasks by \citet{ji_survey_2023}: a response is \emph{intrinsically} hallucinated if it contradicts an input source, and \emph{extrinsically} hallucinated if it cannot be verified from that source -- a notion that presupposes a single explicit input. As LLMs increasingly generate free-form text without such a source, two reframings followed. \citet{huang_survey_2024} proposed \emph{factuality} versus \emph{faithfulness} as the LLM-era replacement axis (correctness against the world vs.\ adherence to context), while \citet{augenstein_factuality_2024} argued that \emph{factuality} and \emph{hallucination} should be treated as distinct concepts entirely: factuality requires an external oracle, whereas hallucination admits an internal definition -- whether the response matches the evidence the model was supposed to use. The separation is consequential because the two concepts overlap but are not identical: a model can be factually correct while unfaithful to its evidence, or factually wrong while faithful to a misleading source.

\paragraph{HalluLens and the new extrinsic/intrinsic taxonomy.}
\citet{bang-etal-2025-hallulens} consolidate this picture and introduce the \textsc{HalluLens} benchmark. They redefine \emph{intrinsic} hallucinations as responses unsupported by information \emph{explicitly provided at inference time}, and \emph{extrinsic} hallucinations as responses unsupported by the \emph{pretraining knowledge the model is expected to have encountered}. The reformulation is operational: each regime corresponds to a concrete source of evidence against which a response can be evaluated, with ground-truth labels defined without implicit appeal to world truth. \textsc{HalluLens} instantiates this with three extrinsic tasks anchored to Wikipedia, along with items generated on-the-fly to mitigate test-set leakage. The Wikipedia anchor rests on a documented approximation -- recent pretraining mixtures are assumed to contain Wikipedia, so inconsistency with a Wikipedia article is treated as a proxy for inconsistency with pretraining knowledge and remains imperfect for long-form generation \citep{fan_halluhard_2026}.

\paragraph{Benchmark heterogeneity and our positioning.}
With the operational split in hand, the existing landscape can be repartitioned by the definition each benchmark encodes. \emph{Explicitly intrinsic} benchmarks include \textsc{RAGTruth} \citep{niu-etal-2024-ragtruth}, which provides word-level hallucination annotations for retrieval-augmented generation. Closed-book QA benchmarks such as \textsc{TriviaQA} \citep{joshi_triviaqa_2017}, \textsc{Natural Questions} \citep{kwiatkowski_natural_2019}, and \textsc{SimpleQA} \citep{wei_measuring_2024} target some dimension of factuality but sit ambiguously with respect to the operational split: classifying them as extrinsic-regime benchmarks would require evidence that all their questions are answerable from the model's pretraining data, which is generally not established. \textsc{HalluLens} fills this gap by restricting questions to those generated from Wikipedia, making the pretraining-coverage assumption a documented approximation rather than an implicit one; \emph{explicitly extrinsic} benchmarks otherwise remain comparatively rare. As a result, comparing performance across these benchmarks partly compares definitions rather than methods. We therefore pair \textsc{RAGTruth} (intrinsic regime, with human span annotations) with the three \textsc{HalluLens} tasks -- \textsc{PreciseWikiQA}, \textsc{LongWiki}, and \textsc{NonExistentRefusal} -- to cover three extrinsic settings with explicit evaluation procedures. Together, these commit us to the hallucination/factuality distinction articulated by \citet{augenstein_factuality_2024} and operationalized by \citet{bang-etal-2025-hallulens}, and allow us to compare uncertainty methods while keeping the meaning of each target explicit. Cross-regime differences still combine target, task, and generation effects.

%% file: appendices/ue_methods_detailed.tex
This appendix specifies the uncertainty estimators evaluated in our benchmark, briefly described in \Cref{sec:background}. We pursue two goals jointly: a conceptual map organized by the signal each method exploits, and an implementation-faithful description of the exact variants executed in our pipeline. Every formula below was derived by auditing the corresponding implementation in our \textsc{LM-Polygraph}-based pipeline \citep{vashurin-etal-2025-benchmarking,fadeeva2023lm} rather than from the high-level descriptions in the original papers.

\paragraph{Pipeline.} We use \textsc{LM-Polygraph} as the uncertainty-score computation engine. The framework separates two stages: shared intermediate quantities (greedy responses, token log-probabilities, stochastic samples, semantic relation matrices, internal representations) are computed \emph{once} per input, and the estimators consume these quantities to produce the final scores. We do not modify any estimator implementation. The version of \textsc{LM-Polygraph} we use is \href{https://github.com/IINemo/lm-polygraph/commit/0e5bcdd35efe7de4e3ae8124427aa12575a1bd2c}{\texttt{main/0e5bcdd35efe7de4e3ae8124427aa12575a1bd2c}}. Downstream evaluation is performed in our analysis pipeline, externally to \textsc{LM-Polygraph}.

\subsection{Implementation details}
\label{apx:ue_impl_details_sec}

Table~\ref{tab:ue_impl_details} consolidates the numerical hyperparameters and external models used in the executed pipeline. Default \textsc{LM-Polygraph} settings are used for any value not listed.

    \subsection{Code availability}\label{apx:code_availability}
The code is available here: \href{https://anonymous.4open.science/r/uncertainty-benchmark-867C}{https://anonymous.4open.science/r/uncertainty-benchmark-867C}.
    
    \subsection{Compute resources}\label{sec:hardware}
All GPU-demanding computation was performed on a university HPC cluster using usage-constrained GPU allocations. The primary hardware was an NVIDIA RTX PRO 6000 Blackwell (96 GB GDDR7) for large-model runs, with NVIDIA RTX 6000 Ada/RTX A6000 (48 GB GDDR6) nodes used for smaller models. Nodes were equipped with 24–32 CPU cores and 192–512 GB RAM. The two main GPU-demanding phases totaled approximately 700 productive GPU-hours:
    
    \begin{itemize}
        \item Benchmark inference (3 tasks × 3 models, up to 2,000 samples each, ~50 uncertainty estimators): roughly 415 GPU-hours for HalluLens tasks and 255 GPU-hours for RAGTruth, ranging from ~5 GPU-hours per run (small datasets, 7B models) to ~87 GPU-hours (large datasets, 13B models). 48 GB of VRAM suffices for 7B models at a batch size of 1; 13B models require 96 GB.
        
        \item LongWiki evaluation (Llama-3.1-70B-AWQ judge via vLLM): ~31 productive GPU-hours across 2 model runs, using 2 GPUs simultaneously (judge server + evaluation client).
    \end{itemize}

\subsection{Notation and conventions}
\label{apx:ue_notation}

Let $x$ denote the input query and $\hat{y} = (\hat{y}_1, \dots, \hat{y}_T)$ the \emph{response to be scored} of length $T$. In the standard free-generation setting, $\hat{y}$ is the greedy response of the model; in the \ragtruth dataset, where responses have already been generated and annotated, $\hat{y}$ is the observed response supplied by the dataset. The autoregressive conditional probability of the token $\hat{y}_t$ is $p_{\vtheta}(\hat{y}_t \mid x, \hat{y}_{<t})$, and the unconditional language-model probability (input $x$ removed) is $p_{\vtheta}(\hat{y}_t \mid \hat{y}_{<t})$. The full predictive distribution at position $t$ lives on the simplex $\Delta(\mathcal{V})$ over the vocabulary of size $V = |\mathcal{V}|$, i.e.\ $p_{\vtheta}(\cdot \mid x, \hat{y}_{<t}) \in \Delta(\mathcal{V})$. The full sequence probability is $P_{\vtheta}(\hat{y} \mid x) = \prod_{t=1}^{T} p_{\vtheta}(\hat{y}_t \mid x, \hat{y}_{<t})$. Shannon entropy of any $q \in \Delta(\mathcal{V})$ is $\mathcal{H}(q) = -\sum_{v \in \mathcal{V}} q(v) \log q(v)$, and $\mathcal{U}(\mathcal{V})$ denotes the uniform distribution on $\mathcal{V}$.

When stochastic sampling is required, we draw $S = 10$ responses $y^{(1)}, \dots, y^{(S)} \sim p_{\vtheta}(\cdot \mid x)$ at sampling temperature $1$ with random seed $42$. We write $|y^{(s)}|$ for the length of sample $s$ and $P_{\vtheta}(y^{(s)} \mid x) = \prod_{t=1}^{|y^{(s)}|} p_{\vtheta}(y^{(s)}_t \mid x, y^{(s)}_{<t})$ for its sequence probability. The same sample pool is shared across all sample-based estimators for a given input, so any two such estimators are compared on identical samples. For semantic-consistency methods, NLI-based pairwise scores are
\[
E_{ij} = p_{\mathrm{NLI}}(\mathrm{entailment} \mid y^{(i)}, y^{(j)}),
\quad
C_{ij} = p_{\mathrm{NLI}}(\mathrm{contradiction} \mid y^{(i)}, y^{(j)}),
\]
produced by \texttt{microsoft/deberta-large-mnli} \citep{he2021deberta}.\footnote{\href{https://huggingface.co/microsoft/deberta-large-mnli}{https://huggingface.co/microsoft/deberta-large-mnli}}
Cross-encoder semantic similarity, when used, is computed with
\texttt{cross-encoder/stsb-roberta-large}.\footnote{\href{https://huggingface.co/cross-encoder/stsb-roberta-large}{https://huggingface.co/cross-encoder/stsb-roberta-large}}

\paragraph{Score orientation.} All estimators are reported as \emph{uncertainty} scores: larger values indicate greater uncertainty. Two estimators in \textsc{LM-Polygraph} are confidence-oriented (Rényi Divergence and Fisher--Rao Distance) and are negated before analysis. The formulas below are uncertainty-oriented.

\paragraph{Name versus executed formula.} Two estimator names refer to a probability, whereas the executed scores are log-transformations of that probability. We make this explicit to avoid confusion. Maximum Sequence Probability refers to the quantity $P_{\vtheta}(\hat{y} \mid x)$, but the implemented score in \textsc{LM-Polygraph} is its negative log -- the sequence negative log-likelihood (NLL). Similarly, Perplexity refers to $\exp\!\bigl(-\tfrac{1}{T}\sum_t \log p_{\vtheta}(\hat{y}_t \mid x, \hat{y}_{<t})\bigr)$, whereas the executed score is the \emph{logarithm} of this perplexity, i.e.\ the length-normalized sequence negative log-likelihood (LN-NLL).

\paragraph{Definitional convention.} We use $:=$ for the exact quantity returned by the implementation and $=$ for subsequent algebraic rewrites.

\paragraph{Configuration convention.} Unless stated otherwise, we use the default \textsc{LM-Polygraph} configuration. When a hyperparameter materially affects the score, we report its value, see Table~\ref{tab:ue_impl_details}.

\subsection{White-box estimators} \label{apx:ue_whitebox} %

White-box estimators require access to information beyond the generated text of the response: the model's logits, attention maps, hidden states, or training data. We organize them into five families, ordered from the most direct readouts of the predictive distribution to the most indirect: (i) information-based estimators, which aggregate token-level distributional quantities computed in a single forward pass; (ii) sample-dispersion estimators, which lift these quantities to multiple stochastic samples and combine them with measures of inter-sample dispersion; (iii) internal-state estimators, which use attention or hidden-state geometry; (iv) training-based estimators, which score representation-space distance from a fitted reference distribution; and (v) reflexive estimators, which prompt the model to judge its own answer.

\subsubsection{Information/Logit-based estimators} \label{apx:ue_information_based} %

These estimators consume only the token-level predictive distributions encountered along a single response $\hat{y}$ and aggregate a position-wise quantity across the sequence. Within this family we distinguish methods that look only at the probability of the realized token (Maximum Sequence Probability, Perplexity), methods that examine the full predictive distribution at each position (Mean Token Entropy, SelfCertainty, Rényi Divergence, Fisher--Rao Distance), methods that contrast the conditional predictive distribution with its unconditional counterpart (Pointwise Mutual Information and its conditional variant), and two methods that introduce auxiliary signals at the token level (TokenSAR, Claim-Conditioned Probability).

\paragraph{Maximum Sequence Probability.} Maximum Sequence Probability is the most basic logit-based estimator: it considers only the probability assigned to the realized response and ignores the rest of the predictive distribution. The executed score is the sequence negative log-likelihood,
\[
U_{\mathrm{MSP}}(x; \hat{y})
\;:=\;
-\sum_{t=1}^{T} \log p_{\vtheta}(\hat{y}_t \mid x, \hat{y}_{<t})
\;=\;
-\log P_{\vtheta}(\hat{y} \mid x).
\]
The score is zero when the model assigns probability $1$ to every realized token and grows without bound as confidence in the realized response drops. It is length-sensitive: longer responses accumulate more uncertainty at constant per-token confidence.

\paragraph{Perplexity.}
Perplexity removes the length sensitivity of Maximum Sequence Probability by averaging the per-token negative log-likelihood,
\[
U_{\mathrm{PPL}}(x; \hat{y})
\;:=\;
-\frac{1}{T} \sum_{t=1}^{T} \log p_{\vtheta}(\hat{y}_t \mid x, \hat{y}_{<t}),
\]
which equals the logarithm of the standard perplexity of $\hat{y}$ under the model. Like Maximum Sequence Probability, this score uses only the probability of the realized token, but it is comparable across responses of
different lengths \citep{fomicheva2020unsupervised}.

\paragraph{Mean Token Entropy.}
Mean Token Entropy is the simplest logit-based estimator that uses the \emph{full} predictive distribution at each position rather than only the probability of the realized token. The score averages the Shannon entropy of the predictive distribution across positions, which equals the chain-rule expansion of the sequence entropy
\citep{cover_elements_2001,malinin2020uncertainty}:
\[
U_{\mathrm{MTE}}(x)
\;:=\;
\frac{1}{T} \sum_{t=1}^{T} \mathcal{H}\!\bigl(p_{\vtheta}(\cdot \mid x, \hat{y}_{<t})\bigr).
\]
Unlike MSP and PPL, MTE is not determined by the realized token alone: even when $\hat{y}_t$ is itself among the most probable tokens, MTE can still be large if probability mass is spread broadly across alternatives; conversely, when $\hat{y}_t$ has only moderate probability but the distribution is peaked on a few competing tokens, MTE remains small because the distribution is concentrated.

\paragraph{SelfCertainty.}
SelfCertainty compares a uniform reference distribution to the model's predictive distribution, using the \emph{reverse} KL direction relative to the entropy-like divergence-from-uniform scores used by Rényi Divergence and Fisher--Rao Distance below. Let $\tilde{\mathcal{V}}_t \subseteq \mathcal{V}$ be the set of vocabulary entries with finite log-probability at position $t$. The implemented token-level quantity is
\[
\KL\!\bigl(\mathcal{U}(\tilde{\mathcal{V}}_t) \,\|\, p_{\vtheta}(\cdot \mid x, \hat{y}_{<t})\bigr)
=
-\log |\tilde{\mathcal{V}}_t|
-
\frac{1}{|\tilde{\mathcal{V}}_t|}
\sum_{v \in \tilde{\mathcal{V}}_t}
\log p_{\vtheta}(v \mid x, \hat{y}_{<t}),
\]
and the sequence-level uncertainty score reverses the polarity by taking the negative mean:
\[
U_{\mathrm{SC}}(x)
\;:=\;
-\frac{1}{T} \sum_{t=1}^{T}
\KL\!\bigl(\mathcal{U}(\tilde{\mathcal{V}}_t) \,\|\, p_{\vtheta}(\cdot \mid x, \hat{y}_{<t})\bigr).
\]
A flat predictive distribution yields a small divergence and $U_{\mathrm{SC}} \approx 0$ (high uncertainty); a sharply peaked distribution yields a large divergence and $U_{\mathrm{SC}} \ll 0$ (low uncertainty). Unlike entropy or Rényi-style scores, the reverse-KL direction is especially sensitive to tokens assigned very small probability: a single very small $p_{\vtheta}(v \mid \cdot)$ blows up one term of the sum.

\paragraph{Rényi Divergence.}
Rényi Divergence also compares the predictive distribution to a uniform reference, but in the \emph{opposite direction} from SelfCertainty -- and on a temperature-scaled version of the distribution. Let
\[
q_t(v \mid x, \hat{y}_{<t};\tau)
=
\softmax\!\left(\log p_{\vtheta}(\cdot \mid x, \hat{y}_{<t}) \,/\, \tau \right)\!(v)
\;\in\; \Delta(\mathcal{V}).
\]
The token-level Rényi divergence of order $\alpha$ from $q_t$ to the uniform
distribution is
\[
D_{\alpha}\!\bigl(q_t(\cdot \mid x, \hat{y}_{<t};\tau) \,\|\, \mathcal{U}(\mathcal{V})\bigr)
=
\frac{1}{\alpha - 1} \log \sum_{v \in \mathcal{V}} q_t(v \mid x, \hat{y}_{<t};\tau)^{\alpha}
- \log |\mathcal{V}|,
\]
and the sequence-level uncertainty score, after polarity reversal, is
\[
U_{\mathrm{Renyi}}(x; \tau)
\;:=\;
-\frac{1}{T} \sum_{t=1}^{T}
D_{\alpha}\!\bigl(q_t(\cdot \mid x, \hat{y}_{<t};\tau) \,\|\, \mathcal{U}(\mathcal{V})\bigr).
\]
Larger divergence corresponds to a sharper, more concentrated distribution, hence the negation. The parameter $\alpha$ controls sensitivity to the distribution: $\alpha < 1$ gives more weight to low-probability tokens, $\alpha > 1$ emphasizes the peak; the temperature $\tau$ smooths the distribution before the divergence is computed. SelfCertainty should \emph{not} be read as the $\alpha = 1$ case of this estimator: it uses the KL direction $\mathcal{U} \,\|\, p$, whereas this Rényi score uses $q \,\|\, \mathcal{U}$, which measures a different geometric quantity (peak concentration rather than coverage). We use $\alpha = 0.5$ and $\tau = 2$ \citep{darrin2023rainproof}.

\paragraph{Fisher--Rao Distance.}
Fisher--Rao Distance also measures how far the temperature-scaled predictive distribution $q_t(\cdot \mid x, \hat{y}_{<t};\tau)$ lies from the uniform distribution on $\Delta(\mathcal{V})$, but uses Fisher--Rao geometry rather than a divergence. The token-level geodesic distance is
\[
d_{\mathrm{FR}}\!\bigl(q_t(\cdot \mid x, \hat{y}_{<t};\tau), \mathcal{U}(\mathcal{V})\bigr)
=
\frac{2}{\pi} \arccos\!\left(\sum_{v \in \mathcal{V}} \sqrt{q_t(v \mid x, \hat{y}_{<t};\tau) \cdot \tfrac{1}{|\mathcal{V}|}}\right),
\]
and the sequence-level uncertainty score, after polarity reversal, is
\[
U_{\mathrm{FR}}(x; \tau)
\;:=\;
-\frac{1}{T} \sum_{t=1}^{T}
d_{\mathrm{FR}}\!\bigl(q_t(\cdot \mid x, \hat{y}_{<t};\tau), \mathcal{U}(\mathcal{V})\bigr).
\]
The geodesic distance is zero when $q_t$ is uniform and approaches $1$ as $q_t$ becomes more peaked; the negation orients the score so that a flatter (more uncertain) distribution receives a larger value. Unlike any divergence, Fisher-Rao is a true metric on the simplex (symmetric, satisfies the triangle inequality). We use $\tau = 2$
\citep{darrin2023rainproof}.

\paragraph{Mean Pointwise Mutual Information.}
The estimators above measure either how confident the model is in the realized response (MSP, PPL) or how broadly its predictive distribution spreads (MTE, SelfCertainty, Rényi, Fisher--Rao). They cannot detect responses that are likely under the model regardless of the input, that is, responses whose probability does not benefit from prompt grounding. Mean Pointwise Mutual Information addresses this by contrasting, for each realized token, the conditional log-probability $\log p_{\vtheta}(\hat{y}_t \mid x, \hat{y}_{<t})$ with the unconditional log-probability $\log p_{\vtheta}(\hat{y}_t \mid \hat{y}_{<t})$. The latter is obtained by an additional forward pass on the response alone, with the prompt $x$ removed. The executed score is
\[
U_{\mathrm{PMI}}(x; \hat{y})
\;:=\;
-\frac{1}{T} \sum_{t=1}^{T}
\bigl[\log p_{\vtheta}(\hat{y}_t \mid x, \hat{y}_{<t}) - \log p_{\vtheta}(\hat{y}_t \mid \hat{y}_{<t})\bigr].
\]
When the conditional probability exceeds the unconditional, the prompt is informative and $U_{\mathrm{PMI}}$ is negative (low uncertainty); when the two are comparable, the response could have been generated without the prompt and $U_{\mathrm{PMI}}$ approaches zero (high uncertainty), which is interpreted as a signal of weak grounding \citep{van_der_poel_mutual_2022}.

\paragraph{Mean Conditional Pointwise Mutual Information.}
Mean Conditional Pointwise Mutual Information refines PMI by gating the unconditional correction: the correction is applied only at positions where the predictive distribution is locally uncertain, and is ignored at sharp positions, where the unconditional log-probability would add noise rather than signal. ``Locally uncertain'' is operationalized through the same per-position entropy used by MTE: the correction is applied only when $\mathcal{H}(p_{\vtheta}(\cdot \mid x, \hat{y}_{<t})) \ge \tau$, with weight $\lambda$. The executed score is
\[
U_{\mathrm{CPMI}}(x; \hat{y})
\;:=\;
-\frac{1}{T} \sum_{t=1}^{T}
\Bigl[
\log p_{\vtheta}(\hat{y}_t \mid x, \hat{y}_{<t})
- \lambda \cdot \1\!\bigl[\mathcal{H}(p_{\vtheta}(\cdot \mid x, \hat{y}_{<t})) \ge \tau\bigr]
\cdot \log p_{\vtheta}(\hat{y}_t \mid \hat{y}_{<t})
\Bigr].
\]
At low-entropy positions, the score reduces to the realized-token negative log-likelihood (Perplexity-like behavior); at high-entropy positions, it behaves like PMI \citep{van_der_poel_mutual_2022}. We use the \textsc{LM-Polygraph} defaults $\tau = 0.0656$ and $\lambda = 3.599$.

\paragraph{TokenSAR.}
TokenSAR re-weights the per-token negative log-likelihoods by an external measure of token relevance, so that tokens whose removal changes the meaning of the response contribute more to the score than function words or filler.
Token relevance is computed by leave-one-out cross-encoder similarity: for each position $t$, let $\mathrm{sim}_t \in [0,1]$ be the semantic similarity between $\hat{y}$ and the response with token $\hat{y}_t$ removed. The
unnormalized relevance is $1 - \mathrm{sim}_t$, and the normalized relevance is $R_t = (1 - \mathrm{sim}_t) / \sum_{t'=1}^{T} (1 - \mathrm{sim}_{t'})$. The executed score is
\[
U_{\mathrm{TSAR}}(x; \hat{y})
\;:=\;
\sum_{t=1}^{T} R_t \cdot \bigl[-\log p_{\vtheta}(\hat{y}_t \mid x, \hat{y}_{<t})\bigr].
\]
Only high-relevance improbable tokens drive uncertainty; an improbable but semantically dispensable token has little effect \citep{duan_shifting_2024}.

\paragraph{Claim-Conditioned Probability.}
Claim-Conditioned Probability targets a different failure mode of standard token-probability estimators: at each position, several alternative tokens may be plausible substitutes, and only some of them would change the meaning of the response. The executed pipeline proceeds in three steps. At each position $t$, the top-$10$ token alternatives $\mathcal{A}_t = \{a_{t,0}, a_{t,1}, \dots, a_{t,9}\}$ are identified, where $a_{t,0}$ is the observed token if it belongs to the selected alternatives. The post-hoc calculator moves that token to the first position when present; it does not insert it when it lies outside the top ten. Each alternative is decoded to text and stripped of surrounding punctuation and whitespace. Bidirectional NLI compares these token strings with the first alternative, rather than comparing complete substituted responses. The two directional labels are combined into a single label $c_{t,i} \in \{\mathrm{entail}, \mathrm{contra}, \mathrm{neutral}\}$: if both directions agree, that label is used; if one direction is entail and the other contra, the label is neutral; otherwise the single non-neutral label is used. Alternatives with $c_{t,i} = \mathrm{entail}$ and the first alternative $a_{t,0}$ (included by construction) form the entailment set $\mathcal{E}_t$, alternatives with $c_{t,i} = \mathrm{contra}$ form the contradiction set $\mathcal{C}_t$, and neutral alternatives are discarded. At position $t$, the claim-conditioned probability is the share of probability mass on entailing alternatives within the entailing-or-contradicting subset,
\[
\mathrm{CCP}_t
=
\frac{\sum_{a \in \mathcal{E}_t} p_{\vtheta}(a \mid x, \hat{y}_{<t})}
     {\sum_{a \in \mathcal{E}_t \cup \mathcal{C}_t} p_{\vtheta}(a \mid x, \hat{y}_{<t})},
\]
and the sequence-level uncertainty score is the negated product across positions,
\[
U_{\mathrm{CCP}}(x) \;:=\; -\prod_{t=1}^{T} \mathrm{CCP}_t.
\]
Since neutral alternatives are excluded from the denominator, the ratio measures semantic support \emph{conditional on the alternative being either entailing or contradicting}, not support against the full predictive
distribution. For a supplied RAGTruth response, the first alternative need not be the observed token; the frequency and effect of this case have not been quantified. A single position at which probability leaks toward contradicting alternatives is enough to pull the product toward zero and the score toward zero in absolute value (high uncertainty); concentration of mass on entailing alternatives produces a near-$1$ product and a score near $-1$ (low uncertainty) \citep{fadeeva_fact-checking_2024}.

\subsubsection{Sample semantic dispersion estimators}
\label{apx:ue_sample_dispersion}

Sample-dispersion estimators draw $S$ stochastic samples $y^{(1)}, \dots, y^{(S)}$ from the predictive distribution and combine sample-level probabilities with a measure of dispersion across samples. Beyond the probabilities themselves, several estimators in this family use auxiliary signals (semantic relations from an NLI model, semantic similarity from a cross-encoder, or internal embeddings); we characterize them as probability-aware sample-based estimators rather than as pure consistency estimators. They mirror the structure of the logit-based family: probability-only methods (Monte Carlo Sequence Entropy and its length-normalized variant), methods that summarize agreement under a notion of meaning (Semantic Entropy, Semantic Density), and hybrid methods that combine probability with an external relevance or consistency signal (SentenceSAR, SAR, the Cocoa family).

\paragraph{Monte Carlo Sequence Entropy.}
Monte Carlo Sequence Entropy approximates the model's expected sequence negative log-likelihood under its own sampling distribution:
\[
U_{\mathrm{MCSE}}(x)
\;:=\;
-\frac{1}{S} \sum_{s=1}^{S} \log P_{\vtheta}(y^{(s)} \mid x)
\;=\;
-\frac{1}{S} \sum_{s=1}^{S} \sum_{t=1}^{|y^{(s)}|} \log p_{\vtheta}\!\bigl(y^{(s)}_t \mid x, y^{(s)}_{<t}\bigr).
\]
A model that consistently produces high-probability samples receives a low score; a model that scatters probability mass across many possible responses receives a high score \citep{malinin2020uncertainty}.

\paragraph{Monte Carlo Normalized Sequence Entropy.}
Monte Carlo Normalized Sequence Entropy is the length-normalized version of the previous estimator: each sample's negative log-likelihood is divided by its own length before averaging, so that the score is comparable across prompts that elicit responses of different lengths:
\[
U_{\mathrm{MCNSE}}(x)
\;:=\;
-\frac{1}{S} \sum_{s=1}^{S} \frac{1}{|y^{(s)}|} \log P_{\vtheta}(y^{(s)} \mid x).
\]

\paragraph{Semantic Entropy.}
Semantic Entropy uses the same sample-based negative log-likelihoods, but groups them by meaning before computing entropy, so that paraphrases of the same answer do not inflate the score. The pipeline first partitions the samples into semantic equivalence classes $C_1, \dots, C_M$ using bidirectional NLI entailment: $y^{(i)}$ and $y^{(j)}$ are placed in the same class when they entail one another. The probability of class $C_m$ is then obtained by summing sample probabilities and renormalizing:
\[
P(C_m \mid x)
=
\frac{\sum_{s \,:\, y^{(s)} \in C_m} P_{\vtheta}(y^{(s)} \mid x)}
     {\sum_{s=1}^{S} P_{\vtheta}(y^{(s)} \mid x)}.
\]
The uncertainty score is the entropy of this class distribution:
\[
U_{\mathrm{SE}}(x) \;:=\; -\sum_{m=1}^{M} P(C_m \mid x) \log P(C_m \mid x).
\]
A model that always produces the same meaning regardless of surface form yields a single class and zero entropy; a model that distributes mass over several meanings yields large entropy
\citep{kuhn2023semantic,farquhar_detecting_2024}.

\paragraph{Semantic Density.}
Semantic Density replaces the hard equivalence classes of Semantic Entropy with a soft NLI-based kernel and a probability-weighted aggregation around the greedy response $\hat{y}$. Let $\tilde{P}_{\vtheta}(y^{(s)} \mid x) = \exp\!\bigl(\tfrac{1}{|y^{(s)}|}\log P_{\vtheta}(y^{(s)} \mid x)\bigr)$ denote the length-normalized sample probability, and let $K_s$ denote the NLI-based kernel score that quantifies the soft semantic agreement between sample $s$ and the surrounding samples (including $\hat{y}$). Writing $\tilde{P}_{\vtheta}(\hat{y} \mid x)$ for the length-normalized probability of the greedy response, the executed score is the negated probability-weighted kernel density
\[
U_{\mathrm{SD}}(x; \hat{y})
\;:=\;
-\frac{\sum_{s=1}^{S} \tilde{P}_{\vtheta}(y^{(s)} \mid x) \, K_s + \tilde{P}_{\vtheta}(\hat{y} \mid x)}
      {\sum_{s=1}^{S} \tilde{P}_{\vtheta}(y^{(s)} \mid x) + \tilde{P}_{\vtheta}(\hat{y} \mid x)}.
\]
The score is close to $-1$ (low uncertainty) when the probable samples cluster in a dense semantic neighborhood of $\hat{y}$, and close to $0$ (high uncertainty) when the kernel scores are uniformly low \citep{qiu_semantic_2024}.

\paragraph{SentenceSAR.}
SentenceSAR extends the relevance-weighting idea of TokenSAR to the multi-sample setting. The contribution of each sample to the score is its own probability augmented by a relevance term that aggregates the probabilities of \emph{other} samples weighted by their semantic similarity to the current one. Writing $\mathrm{sim}_{sj}$ for the cross-encoder similarity between $y^{(s)}$ and $y^{(j)}$ and $\tau$ for a temperature parameter, the support term is
\[
R_S(s; \tau) = \frac{1}{\tau} \sum_{j \neq s} P_{\vtheta}(y^{(j)} \mid x) \, \mathrm{sim}_{sj},
\]
and the executed score is
\[
U_{\mathrm{SentSAR}}(x; \tau)
\;:=\;
\frac{1}{S} \sum_{s=1}^{S} -\log\!\bigl(P_{\vtheta}(y^{(s)} \mid x) + R_S(s; \tau)\bigr).
\]
A sample that is both probable and semantically supported by other samples contributes little to the score; a sample that is improbable and semantically isolated contributes a large negative log \citep{duan_shifting_2024}. We use $\tau = 1$.

\paragraph{SAR.}
SAR is the full pipeline of \citet{duan_shifting_2024}, combining TokenSAR at the token level and SentenceSAR at the sentence level. The raw sample probability used in SentenceSAR is replaced with the relevance-adjusted probability from TokenSAR. Concretely, with $\tilde{P}^{\mathrm{tsar}}_{\vtheta}(y^{(s)}) = \exp\!\bigl(-U_{\mathrm{TSAR}}(x, y^{(s)})\bigr)$ the relevance-weighted probability of sample $s$, the support term becomes $\tilde{R}_S(s; \tau) = (1/\tau) \sum_{j \neq s} \tilde{P}^{\mathrm{tsar}}_{\vtheta}(y^{(j)}) \, \mathrm{sim}_{sj}$, and the executed score is
\[
U_{\mathrm{SAR}}(x; \tau)
\;:=\;
\frac{1}{S} \sum_{s=1}^{S} -\log\!\bigl(\tilde{P}^{\mathrm{tsar}}_{\vtheta}(y^{(s)}) + \tilde{R}_S(s; \tau)\bigr).
\]
SAR composes two relevance signals -- token-level for each sample and
sentence-level across samples -- into a single uncertainty score. We use
$\tau = 1$.

\paragraph{Cocoa family.}
The Cocoa family follows a simple template: a base uncertainty score is multiplied by an inter-sample semantic dissimilarity factor, so that the final score is large only when the response is uncertain on its own \emph{and} semantically dissimilar to the alternative samples \citep{vashurin_cocoa_2025}. Let $u_{\mathrm{base}}(x; \hat{y})$ denote the base uncertainty and let $\mS^{\mathrm{sent}} \in \R^{K \times K}$ be the cross-encoder sentence-similarity matrix over the $K$ sentence units computed by \textsc{LM-Polygraph} for this input (greedy response and/or samples, depending on the module configuration). The dissimilarity factor is the mean of $1 - S^{\mathrm{sent}}_{ij}$ across the full matrix,
\[
D_{\mathrm{sent}}(x)
=
\frac{1}{K^2} \sum_{i=1}^{K} \sum_{j=1}^{K} \bigl(1 - S^{\mathrm{sent}}_{ij}\bigr),
\]
and the executed Cocoa score is the product
\[
U_{\mathrm{Cocoa}}(x; \hat{y})
\;:=\;
u_{\mathrm{base}}(x; \hat{y}) \cdot D_{\mathrm{sent}}(x).
\]
We instantiate three variants by choosing $u_{\mathrm{base}} \in \{U_{\mathrm{MSP}}, U_{\mathrm{PPL}}, U_{\mathrm{MTE}}\}$, denoted CocoaMSP, CocoaPPL, and CocoaMTE. An improbable response that is semantically consistent with the alternative samples is treated as a paraphrase and receives a low score; an improbable response that is also semantically isolated receives a high score. \emph{Implementation note:} although the \textsc{LM-Polygraph} source comment states that the diagonal is excluded, the executed code averages over the full $K \times K$ matrix. Since self-similarity is $1$, the diagonal entries contribute $0$ to $1 - S^{\mathrm{sent}}_{ij}$ and thus dilute $D_{\mathrm{sent}}$ by a factor of $(K-1)/K$.

\subsubsection{Internal-state estimators}
\label{apx:ue_internal}

Internal-state estimators use latent computations of the model -- attention weights or hidden representations -- rather than only the response distribution. Their common premise is that predictive unreliability may already be visible in the model's internal processing dynamics before it is fully reflected in the response token distribution.

\paragraph{AttentionScore.}
AttentionScore summarizes diagonal self-attention weights in the scored input--response sequence. Low diagonal weights increase the score; their association with response errors is evaluated empirically. Writing $\alpha_{\ell, h}(t, t) \in [0, 1]$ for the diagonal self-attention weight at layer $\ell$, head $h$, and position $t$, the executed score is
\[
U_{\mathrm{Attn}}(x; \hat{y})
\;:=\;
-\frac{1}{H} \sum_{h=1}^{H} \sum_{t\in\mathcal{I}(x,\hat{y})}
\log\!\bigl(\alpha_{\ell, h}(t, t) + \varepsilon\bigr),
\]
where $\mathcal{I}(x,\hat{y})$ denotes the token positions in the forward-pass attention matrix. The executed setting uses \texttt{gen\_only=False}, so input positions also contribute; this is not a response-only score. Here $\varepsilon = 10^{-12}$ is a numerical stabilizer and the layer $\ell$ is the model's middle layer ($\ell = L/2$), as selected by \textsc{LM-Polygraph} when no layer is explicitly specified. This choice matters: attention-based diagnostics vary substantially across layers, and the score should be understood as a layer-specific instantiation of the general idea \citep{sriramanan_llm-check_2024}.

\paragraph{RAUQ (Recurrent Attention Uncertainty Quantification).}
RAUQ propagates a confidence signal recursively along the sequence, combining the local token probability with attention from the current token to the previous one, and aggregates the per-layer scores only over the middle third of layers. Let $\mathcal{L} = \{\lfloor L/3 \rfloor, \dots, \lceil 2L/3 \rceil\}$ denote the indices of the middle third of the model's $L$ attention layers. For each $\ell \in \mathcal{L}$, the head attending most strongly (on average) to the previous token is
\[
h^*(\ell) = \arg\max_{h} \frac{1}{T - 1} \sum_{t=2}^{T} \alpha_{\ell, h}(t, t-1).
\]
The recurrent confidence at layer $\ell$ is initialized at
$C_1^{(\ell)} = p_{\vtheta}(\hat{y}_1 \mid x)$ and updated for $t > 1$ by
\[
C_t^{(\ell)}
=
\alpha \cdot p_{\vtheta}(\hat{y}_t \mid x, \hat{y}_{<t})
+
(1 - \alpha) \cdot \alpha_{\ell, h^*(\ell)}(t, t-1) \cdot C_{t-1}^{(\ell)},
\]
where $\alpha \in [0, 1]$ is a mixing weight. The per-layer score is $1 - \tfrac{1}{T} \sum_t \log C_t^{(\ell)}$, and the estimator returns the worst (largest) layer over the middle third:
\[
U_{\mathrm{RAUQ}}(x; \hat{y})
\;:=\;
\max_{\ell \in \mathcal{L}} \left[1 - \frac{1}{T} \sum_{t=1}^{T} \log C_t^{(\ell)}\right].
\]
Unlike AttentionScore, which evaluates each position independently, RAUQ captures how confidence accumulates or degrades along the sequence \citep{vazhentsev_uncertainty-aware_2025}. When $\alpha$ is not specified,
\textsc{LM-Polygraph} selects it through two binary flags: $\1_{\mathrm{ent}} = 1$ if the entropy-based variant is used (and $0$ for the probability variant), and $\1_{\mathrm{instr}} = 1$ if the underlying model is instruction-tuned (and $0$ otherwise). The default is
\[
\alpha \;=\; (1 - \1_{\mathrm{ent}})\,(0.2 + 0.3 \cdot \1_{\mathrm{instr}})
\;+\; \1_{\mathrm{ent}}\,(0.8 + 0.1 \cdot \1_{\mathrm{instr}}),
\]
which yields $\alpha \in \{0.2, 0.5, 0.8, 0.9\}$ depending on the flags. We use the probability variant ($\1_{\mathrm{ent}} = 0$) on instruction-tuned models ($\1_{\mathrm{instr}} = 1$), so the executed $\alpha = 0.5$.

\paragraph{Contextualized Sequence Likelihood (CSL).}
Contextualized Sequence Likelihood reweights token negative log-likelihoods by an attention-derived saliency, so that positions the model deems important when producing the final response contribute more to the score than filler positions. Let $\bar{\alpha}_t = \mathrm{mean}_{\ell, h} \, \alpha_{\ell, h}(\text{last}, t)$ be the mean attention paid to position $t$ from the final generation step, averaged over layers and heads, and let $w_t = \bar{\alpha}_t / \sum_{t'} \bar{\alpha}_{t'}$ be the corresponding saliency weight ($w_t \ge 0$, $\sum_t w_t = 1$). The executed score is
\[
U_{\mathrm{CSL}}(x; \hat{y})
\;:=\;
\sum_{t=1}^{T} w_t \cdot \bigl[-\log p_{\vtheta}(\hat{y}_t \mid x, \hat{y}_{<t})\bigr].
\]
The score has the same form as TokenSAR but uses an internal saliency
signal in place of an external cross-encoder relevance signal
\citep{lin_contextualized_2024}.

\paragraph{EigenScore.}
EigenScore measures the geometric spread of sampled internal embeddings in representation space. Let $\ve^{(s)} \in \R^d$ denote the decoder hidden state of the last generated token for sample $s$, and let $\mE = [\ve^{(1)}, \dots, \ve^{(S)}] \in \R^{d \times S}$ be the matrix whose columns are the $S$ sample embeddings. Writing $\mJ_d = \mI_d - \tfrac{1}{d} \vone_d \vone_d^\top \in \R^{d \times d}$ for the centering matrix and $\alpha$ for a regularization scalar, the regularized centered covariance matrix is
\[
\mC
=
\mE^\top \mJ_d \mE + \alpha \mI_S
\;\in\; \R^{S \times S},
\]
with eigenvalues $\lambda_1, \dots, \lambda_S$. The executed score is the
mean log-eigenvalue,
\[
U_{\mathrm{Eigen}}(x) \;:=\; \frac{1}{S} \sum_{k=1}^{S} \log \lambda_k.
\]
Sampled responses whose embeddings cluster in a low-dimensional subspace yield small eigenvalues and a low score (low uncertainty); sampled responses whose embeddings span many directions yield large eigenvalues and a high score (high uncertainty) \citep{chen_inside_2023}. We use $\alpha = 10^{-3}$.

\subsubsection{Training-based estimators}
\label{apx:ue_training}

Training-based estimators fit a reference distribution in representation space on a held-out training partition and score test inputs by their distance from that reference. They are best read as familiarity or out-of-distribution signals rather than as readouts of response confidence. All variants below operate on decoder hidden states averaged over the generated tokens of $\hat{y}$, and are all fitted to each task's own training split. The relative variants additionally subtract a background distance term fitted on a broad, general-purpose corpus that we describe below.

\paragraph{Mahalanobis Distance.}
Mahalanobis Distance is the basic representation-space estimator. Let $\ve(x, \hat{y}) \in \R^d$ denote the decoder hidden state averaged over all generated tokens of $\hat{y}$, and let $\vmu \in \R^d$ and $\mSigma \in \R^{d \times d}$ be the mean and covariance estimated from the task's training partition. The executed score is the Mahalanobis distance between $\ve(x, \hat{y})$ and the training centroid:
\[
U_{\mathrm{MD}}(x; \hat{y})
\;:=\;
\sqrt{\bigl(\ve(x, \hat{y}) - \vmu\bigr)^\top \mSigma^{-1} \bigl(\ve(x, \hat{y}) - \vmu\bigr)}.
\]
A test input whose representation lies near the training centroid receives a small distance (low uncertainty); an input far from the training distribution receives a large distance (high uncertainty)
\citep{lee2018simple}.

\paragraph{Relative Mahalanobis Distance.}
Relative Mahalanobis Distance subtracts a background Mahalanobis distance, fitted on a broad general-purpose corpus, in order to isolate task-specific novelty from generic rarity:
\[
U_{\mathrm{RMD}}(x; \hat{y})
\;:=\;
U_{\mathrm{MD}}(x; \hat{y}) - U_{\mathrm{MD}_0}(x; \hat{y}),
\]
where $U_{\mathrm{MD}_0}$ uses the background mean and covariance in place of the task mean and covariance. A test input far from both the task and the background receives a small score (general rarity); an input far from
the task but close to the background receives a large score, indicating domain shift specific to the task \citep{ren_simple_2021}. We fit the background distribution on \texttt{allenai/c4} (shard
\texttt{en/c4-train.00000-of-01024.json.gz}).\footnote{\href{https://huggingface.co/datasets/allenai/c4}{https://huggingface.co/datasets/allenai/c4}}

\paragraph{Robust Density Estimation (RDE).}
Robust Density Estimation replaces the Gaussian fit of Mahalanobis Distance with a robust covariance estimate computed over a Kernel-PCA-projected representation space, to better handle non-elliptical embedding distributions and reduce sensitivity to training-set outliers. Let $\tilde{\ve}(x; \hat{y}) \in \R^{d'}$ denote the Kernel-PCA projection of $\ve(x, \hat{y})$, and let $\hat{\vmu} \in \R^{d'}$ and $\hat{\mSigma} \in \R^{d' \times d'}$ be the mean and covariance estimated by the Minimum Covariance Determinant procedure in the projected space. The executed score is the Mahalanobis distance in this robust, dimensionality-reduced space:
\[
U_{\mathrm{RDE}}(x; \hat{y})
\;:=\;
\sqrt{\bigl(\tilde{\ve}(x, \hat{y}) - \hat{\vmu}\bigr)^\top \hat{\mSigma}^{-1} \bigl(\tilde{\ve}(x, \hat{y}) - \hat{\vmu}\bigr)}.
\]
\citep{yoo_detection_2022}.

\paragraph{HUQ-MD and HUQ-RMD.}
The HUQ variants combine a representation-space distance (epistemic-novelty signal) with a probability-based uncertainty (aleatoric signal) via \emph{empirical-CDF rank fusion} \citep{vazhentsev_hybrid_2023}. Because Perplexity and Mahalanobis Distance have different scales and distributions, one cannot combine them by raw averaging; rank fusion instead converts each score to a scale-free percentile relative to the training partition before averaging. Concretely, let $\train$ denote the task's training partition. The empirical CDF of $U_{\mathrm{PPL}}$ on that partition is
\[
F^{\mathrm{train}}_{\mathrm{PPL}}(u)
=
\frac{1}{|\train|} \sum_{x' \in \train} \1\!\bigl[U_{\mathrm{PPL}}(x', \hat{y}(x')) \le u\bigr]
\;\in\; [0, 1],
\]
and $F^{\mathrm{train}}_{\mathrm{MD}}$ is defined analogously for $U_{\mathrm{MD}}$. Evaluating these CDFs at a test input gives its percentile rank on each score (the fraction of training inputs with a
lower score), and the executed HUQ-MD score is the mean of the two percentile ranks:
\[
U_{\mathrm{HUQ\text{-}MD}}(x; \hat{y})
\;:=\;
\tfrac{1}{2}\!\Bigl[
F^{\mathrm{train}}_{\mathrm{PPL}}\!\bigl(U_{\mathrm{PPL}}(x; \hat{y})\bigr)
+
F^{\mathrm{train}}_{\mathrm{MD}}\!\bigl(U_{\mathrm{MD}}(x; \hat{y})\bigr)
\Bigr]
\;\in\; [0, 1].
\]
HUQ-RMD is defined analogously with $U_{\mathrm{RMD}}$ in place of $U_{\mathrm{MD}}$. Because both components are converted to $[0,1]$-valued percentile ranks before averaging, the combination is scale-free and distribution-free.

\subsubsection{Reflexive estimators (white-box)}
\label{apx:ue_reflexive_whitebox}

Reflexive estimators prompt the model to judge its own answer. The white-box variants additionally require direct access to the probability assigned to the ``True'' option; their black-box counterpart is described in Section~\ref{apx:ue_blackbox_reflexive}.

\paragraph{PTrue.}
Given an input $x$ and a proposed answer $\hat{y}$, PTrue constructs a self-evaluation prompt $\mathrm{prompt}_{\mathrm{PT}}(x; \hat{y})$ asking the model to judge whether $\hat{y}$ is correct and reads off the probability assigned to the ``True'' option. The executed score is
\[
U_{\mathrm{PTrue}}(x; \hat{y})
\;:=\;
-\log p_{\vtheta}\!\bigl(\text{``True''} \mid \mathrm{prompt}_{\mathrm{PT}}(x; \hat{y})\bigr).
\]
A high probability of ``True'' yields a small score (low uncertainty); a low probability yields a large score (high uncertainty) \citep{kadavath2022language}. The exact prompt template is:
\begin{quote}\small\ttfamily
Question: \{x\}.\\
Proposed Answer: \{$\hat{y}$\}.\\
Is the proposed answer True or False?
\end{quote}

\paragraph{PTrue Sampling.}
PTrue Sampling augments the self-evaluation prompt with $S$ sampled candidate answers $y^{(1)}, \dots, y^{(S)}$ before asking the truth judgment, in order to expose the model to its own response variability:
\[
U_{\mathrm{PTrue\text{-}S}}(x; \hat{y})
\;:=\;
-\log p_{\vtheta}\!\bigl(\text{``True''} \mid \mathrm{prompt}_{\mathrm{PT\text{-}S}}(x, \hat{y}, y^{(1)}, \dots, y^{(S)})\bigr).
\]
The exact prompt template is:
\begin{quote}\small\ttfamily
Question: \{x\}\\
Here are some ideas that were brainstormed: \{$y^{(1)}, \dots, y^{(S)}$\}\\
Possible answer: \{$\hat{y}$\}\\
Is the possible answer:\\
(A) True\\
(B) False\\
The possible answer is:
\end{quote}

\subsection{Black-box estimators}
\label{apx:ue_blackbox}

Black-box estimators operate only on sampled responses and external semantic/lexical-relation functions, without using token probabilities, attention, or hidden states. We organize them into four groups: the simplest estimators that summarize the semantic-class structure of the samples (NumSet, LabelProb); spectral and graph-density estimators on a pairwise similarity graph (Kernel Language Entropy, EigValLaplacian, Eccentricity, DegMat, LUQ); lexical baselines that replace semantic relations with pairwise surface overlap (ROUGE-L Similarity, BLEU Similarity); and a reflexive black-box estimator (PTrue Empirical).

\subsubsection{Semantic-class summaries}

\paragraph{NumSet.}
NumSet is the simplest semantic-consistency estimator: it counts the number of distinct semantic classes among the sampled responses. Samples are partitioned into bidirectional-entailment classes $C_1, \dots, C_M$, as in Semantic Entropy, and the executed score is the cluster count:
\[
U_{\mathrm{NumSet}}(x) \;:=\; M.
\]
The score is integer-valued, with a minimum of $1$ when all samples express the same meaning and a maximum of $S$ when every sample expresses a different meaning \citep{lin_generating_2024}. The granularity is coarse, but the score is robust to NLI-model calibration.

\paragraph{LabelProb.}
We retain the package name \textsc{LabelProb} for consistency with \textsc{LM-Polygraph}, but the executed implementation is more accurately described as the complement of the dominant semantic-class frequency:
\[
U_{\mathrm{LabelProb}}(x) \;:=\; 1 - \max_{1 \le m \le M} \frac{|C_m|}{S}.
\]
The score is zero when a single semantic class accounts for all samples, and approaches $1 - 1/S$ when every sample belongs to a different class. LabelProb is therefore best understood as a majority-class complement, not as a black-box approximation of maximum sequence probability as the name might suggest \citep{vashurin-etal-2025-benchmarking}.

\subsubsection{Spectral and graph-density estimators}

These estimators build a pairwise relation matrix $\mW = (W_{ij})_{i, j = 1}^{S} \in \R^{S \times S}$ over sampled responses and summarize its topology. Three of them rely on spectral analysis of the corresponding graph (Kernel Language Entropy uses kernel eigenvalues; EigValLaplacian uses Laplacian eigenvalues; Eccentricity uses Laplacian eigenvectors), and the remaining two summarize the graph through its degree structure (DegMat) or through average pairwise scores (LUQ). For EigValLaplacian, DegMat, and Eccentricity, we run three edge constructions: NLI-entailment ($W_{ij} = E_{ij}$), NLI-contradiction ($W_{ij} = 1 - C_{ij}$), and Jaccard token similarity.

\paragraph{Kernel Language Entropy (KLE).}
KLE quantifies semantic uncertainty as the von Neumann entropy of a heat-diffusion kernel on the semantic graph. Let $\mD = \mathrm{diag}\bigl(\sum_j W_{ij}\bigr)$ be the degree matrix, $\mL = \mD - \mW$ the unnormalized graph Laplacian, and $\mK = \exp(-t \mL)$ the heat kernel (subsequently row-normalized to unit trace). Writing $\lambda_k(\mK)$ for the eigenvalues of $\mK$:
\[
U_{\mathrm{KLE}}(x)
\;:=\;
-\Tr\bigl(\mK \log \mK\bigr)
\;=\;
-\sum_{k} \lambda_k(\mK) \log \lambda_k(\mK).
\]
A graph with many disconnected semantic clusters yields several large eigenvalues and a high entropy; a fully connected graph yields a single dominant eigenvalue and low entropy \citep{nikitin_kernel_2024}.

\paragraph{EigValLaplacian.}
EigValLaplacian also operates on the spectrum of the graph, but on the eigenvalues of the normalized Laplacian $\tilde{\mL} = \mI - \mD^{-1/2} \mW \mD^{-1/2}$ rather than on a heat kernel. Writing $\lambda_1, \dots, \lambda_S \in [0, 2]$ for its eigenvalues:
\[
U_{\mathrm{EVL}}(x) \;:=\; \sum_{k=1}^{S} \max(0, 1 - \lambda_k).
\]
Each term measures how close one mode of variation is to being fully disconnected; the sum is a continuous relaxation of the number of semantic sets \citep{lin_generating_2024}.

\paragraph{Eccentricity.}
Eccentricity (Ecc) uses the eigen\emph{vectors} of $\tilde{\mL}$ rather than its eigenvalues. Let $\vu_1, \dots, \vu_k$ be the eigenvectors of $\tilde{\mL}$ corresponding to its smallest $k$ eigenvalues, and stack their per-sample entries into spectral embeddings $\vv_i = \bigl(\vu_1[i], \dots, \vu_k[i]\bigr) \in \R^{k}$ for sample $i$. Writing $\bar{\vv} = \tfrac{1}{S} \sum_i \vv_i$ for the centroid:
\[
U_{\mathrm{Ecc}}(x)
\;:=\;
\left(\sum_{i=1}^{S} \|\vv_i - \bar{\vv}\|_2^2\right)^{1/2}.
\]
A graph in which all samples occupy a tight region of the spectral embedding space yields a small score; a graph with spectral outliers yields a large score \citep{lin_generating_2024}.

\paragraph{DegMat.}
Degree Matrix (DegMat) summarizes a graph's degree structure rather than its spectrum. The executed score is the mean missing pairwise similarity:
\[
U_{\mathrm{DegMat}}(x)
\;:=\;
\frac{S^2 - \sum_{i=1}^{S} \sum_{j=1}^{S} W_{ij}}{S^2},
\]
which equals zero when all pairs have $W_{ij} = 1$ (full mutual agreement) and approaches $1$ when no pairs agree. Like Eccentricity, DegMat depends on the choice of edge construction, and we report all three variants
\citep{lin_generating_2024}.

\paragraph{LUQ.}
Long-text Uncertainty Quantification (LUQ) summarizes the graph through average pairwise NLI confidence, using
the continuous softmax score from the NLI logits rather than discretized entail/contradict decisions. Writing
$\sigma_{ij} = \exp(z^E_{ij}) / \bigl(\exp(z^E_{ij}) + \exp(z^C_{ij})\bigr)$
for the soft entailment confidence between $y^{(i)}$ and $y^{(j)}$:
\[
U_{\mathrm{LUQ}}(x)
\;:=\;
1 - \frac{1}{S(S - 1)} \sum_{\substack{i, j = 1 \\ i \neq j}}^{S} \sigma_{ij}.
\]
A high average pairwise entailment confidence yields a small score (low uncertainty); weak average confidence yields a large score \citep{zhang_luq_2024}.

\subsubsection{Lexical similarity baselines}

\paragraph{ROUGE-L Similarity and BLEU Similarity.}
These baselines replace NLI-based semantic relations with pairwise surface-overlap measures. For
$\mathrm{metric}(\cdot, \cdot) \in \{\mathrm{ROUGE\text{-}L}, \mathrm{BLEU}\}$, the executed score is the negated mean pairwise similarity:
\[
U_{\mathrm{LexSim}}(x)
\;:=\;
-\frac{2}{S(S - 1)} \sum_{1 \le i < j \le S} \mathrm{metric}(y^{(i)}, y^{(j)}).
\]
Surface-form variation across samples yields a high score even when the samples convey the same meaning, so these baselines cannot distinguish valid paraphrasing from semantic disagreement \citep{lin-2004-rouge,papineni-etal-2002-bleu}. We report ROUGE-L and BLEU variants separately.

\subsubsection{Reflexive estimator (black-box)}
\label{apx:ue_blackbox_reflexive}

\paragraph{PTrue Empirical.}
PTrue Empirical is the black-box counterpart of PTrue: instead of reading off the probability of the ``True'' token, it draws $N$ stochastic self-evaluation responses from the model and measures the empirical frequency with which the model judges its own answer to be true:
\[
U_{\mathrm{PTrue\text{-}E}}(x; \hat{y})
\;:=\;
1 - \frac{1}{N} \sum_{n = 1}^{N}
\1\!\Bigl[\text{``True''} \in \mathrm{sample}_n\!\bigl(\mathrm{prompt}_{\mathrm{PT}}(x; \hat{y})\bigr)\Bigr],
\]
using the same self-evaluation prompt as PTrue. We use $N = 10$
\citep{tian_just_2023}.

\begin{table*}[t]
\caption{Numerical hyperparameters and external models for every estimator executed in our benchmark. ``Samples'' is the number of stochastic generations generated per input. Estimators that have no tunable hyperparameters and no external models (MSP, Perplexity, Mean Token Entropy, SelfCertainty, PMI, MC-SE, MC-NSE, CSL) are omitted from the table. Across all sample-based estimators, the sample pool is shared, the sampling temperature is $1$, and the random seed is $42$.}
\label{tab:ue_impl_details}
\centering
\small
\renewcommand{\arraystretch}{1.1}
\begin{tabular}{@{}lll@{}}
\toprule
\textbf{Estimator} & \textbf{Samples} & \textbf{Hyperparameters and external models} \\
\midrule
\multicolumn{3}{l}{\emph{logit-based (Sec.~\ref{apx:ue_information_based})}} \\
Rényi Divergence       & 1 & $\alpha = 0.5$, $\tau = 2$ \\
Fisher--Rao Distance   & 1 & $\tau = 2$ \\
CPMI                   & 1 & $\tau = 0.0656$, $\lambda = 3.599$ \\
TokenSAR               & 1 & cross-encoder \texttt{stsb-roberta-large} \\
CCP                    & 1 & NLI \texttt{deberta-large-mnli}; top-$k$ alternatives, $k = 10$ \\
\midrule
\multicolumn{3}{l}{\emph{Sample semantic dispersion (Sec.~\ref{apx:ue_sample_dispersion})}} \\
MC-SE, MC-NSE          & $S = 10$ & sampling temperature $1$ \\
Semantic Entropy       & $S = 10$ & NLI \texttt{deberta-large-mnli} \\
Semantic Density       & $S = 10$ & NLI \texttt{deberta-large-mnli} \\
SentenceSAR            & $S = 10$ & cross-encoder \texttt{stsb-roberta-large}; $\tau = 1$ \\
SAR                    & $S = 10$ & cross-encoder \texttt{stsb-roberta-large}; $\tau = 1$ \\
Cocoa\{MSP, PPL, MTE\} & $S = 10$ & cross-encoder \texttt{stsb-roberta-large}; \\
\midrule
\multicolumn{3}{l}{\emph{Internal state (Sec.~\ref{apx:ue_internal})}} \\
AttentionScore         & 1 & layer $\ell = L/2$; $\varepsilon = 10^{-12}$ \\
RAUQ                   & 1 & $\alpha = 0.5$; middle-third layers $\mathcal{L} = \{\lfloor L/3 \rfloor, \dots, \lceil 2L/3 \rceil\}$ \\
EigenScore             & $S = 10$ & decoder hidden state of last token; $\alpha = 10^{-3}$ \\
\midrule
\multicolumn{3}{l}{\emph{Training-based (Sec.~\ref{apx:ue_training})}} \\
Mahalanobis Distance   & 1 & decoder hidden states; fitted per task on training partition \\
Relative MD            & 1 & background: \texttt{allenai/c4}, shard \texttt{en/c4-train.00000} \\
RDE                    & 1 & Kernel-PCA projection + MCD covariance \\
HUQ-MD, HUQ-RMD        & 1 & CDF rank fusion of $U_{\mathrm{PPL}}$ and $U_{\mathrm{MD}}$ / $U_{\mathrm{RMD}}$ on training partition \\
\midrule
\multicolumn{3}{l}{\emph{Reflexive, white-box (Sec.~\ref{apx:ue_reflexive_whitebox})}} \\
PTrue                  & 1 & prompt: see \S\ref{apx:ue_reflexive_whitebox} \\
PTrue Sampling         & $S = 10$ & prompt: see \S\ref{apx:ue_reflexive_whitebox} \\
\midrule
\multicolumn{3}{l}{\emph{Black-box (Sec.~\ref{apx:ue_blackbox})}} \\
NumSet, LabelProb              & $S = 10$ & NLI \texttt{deberta-large-mnli} \\
KLE                            & $S = 10$ & NLI \texttt{deberta-large-mnli}; heat-kernel time $t = 0.3$ \\
EigValLaplacian, Ecc, DegMat   & $S = 10$ & NLI \texttt{deberta-large-mnli}; \\
LUQ                            & $S = 10$ & NLI \texttt{deberta-large-mnli} \\
Lexical Similarity             & $S = 10$ & metrics: ROUGE-L, BLEU \\
PTrue Empirical                & $N = 10$ self-eval. & prompt: see \S\ref{apx:ue_blackbox_reflexive} \\
\bottomrule
\end{tabular}
\end{table*}

%% file: appendices/methodology_detailed.tex
These details complete the benchmark setup and evaluation methodology in Sections~3 and~4. Procedures for the sensitivity analyses appear beside their results in \Cref{apx:additive_existing_results}.

\subsection{Datasets and response evaluation}
\label{apx:target_processing}\label{apx:additive_comparisons}
This subsection completes the dataset descriptions in Section~3.1 with the quality formula, label processing and sample counts.

For LongWiki, let $m$ be the number of extracted response claims and $s$ the number judged supported. Precision is $P=s/m$, Recall@$K$ is $R_K=\min(s/K,1)$, and their harmonic mean is $F_K=2PR_K/(P+R_K)$, with $K=32$. We assign $F_K=0$ when both components are zero. The recall component measures supported claims relative to a fixed quota, not coverage of an independently enumerated reference answer.

We use the \texttt{Meta-Llama-3.1-70B-Instruct-AWQ-INT4} checkpoint for the primary HalluLens evaluation. PreciseWikiQA's correctness prompt requests correct, incorrect, or unverifiable judgments, but also contains a Yes/No instruction. The parser accepts \texttt{CORRECT}/\texttt{YES} as correct and \texttt{INCORRECT}/\texttt{NO} as incorrect. An abstention flag can overlap a correctness label. The evaluation retains verifiable judgments without an abstention flag; The tables below report the retained responses and the original judgment categories.

LongWiki responses without a quality target after the knowledge gate are excluded: 4, 1, and 3 of the 2,000 responses for L7B, L13B, and M7B, respectively. \Cref{tab:longwiki_target_components} reports response quality and claim counts. 
\input{tables/additive/population_counts}
\input{tables/target_sensitivity_statistics}
\clearpage

\subsection{AlignScore reference texts}
\label{apx:alignscore_comparator}\label{apx:selection_details}
The reference-based comparator in Section~3.2 requires a reference text and a scoring direction, specified here.

AlignScore \citep{zha2023alignscore} estimates whether one text is supported by another. For PreciseWikiQA and LongWiki, $\mathrm{AlignScore}(y,\mathrm{ref})$  checks the response against a gold answer (AS-Precision-gold) or source Wikipedia page (AS-Precision-Wikipedia), measuring a precision score with respect to the reference. Reversing the arguments, $\mathrm{AlignScore}(\mathrm{ref},y)$ (AS-Recall), measures how much of the reference the response recovers, giving a recall-like score. For NonExistentRefusal, no gold answer is available, so the reference is the abstention template below. The complete scores appear in \Cref{apx:performance_tables}.

\paragraph{Abstention reference.}\label{apx:abstention_template}
This text is used as an AlignScore reference for refusal responses. It is not the prompt used by the judge to assign abstention labels.
\begin{quote}
\textbf{Abstention template.}
“No information is available about this, and I am not aware of any such entity being recognized or documented.
This name does not appear to correspond to anything known, and I do not believe this exists as a real or established thing.
There may be some confusion, or this could be a misspelling or an alternative rendering of a different name.
I am not certain what this refers to, and this term is not widely recognized in any domain I can identify.
If you could provide more context or clarify your intent, that would help determine whether a related concept exists.”
\end{quote}

\subsection{Evaluation metrics}
\label{apx:evaluation_conventions}
The following conventions specify the metrics introduced in Section~4.

We compute AUPRC as average precision. Trapezoidal integration can overestimate the precision--recall area \citep{davis2006relationship}; for a constant score, it gives half of one plus the error rate, rather than the error rate itself. RCE uses the plug-in $\ell_1$ estimator with ten equal-mass bins and the native quality target, including continuous F1@32 for LongWiki.

For PRR, responses are ordered from lowest to highest uncertainty, preserving input order within score ties. At rejection rate $r$, the retained count is $n-\lfloor rn\rfloor$, and retained quality is the mean quality of those responses. We integrate 101 points over rejection rates from 0 to 1. The empty-set endpoint is assigned quality one. The oracle and random curves use the same grid; the random baseline averages 1,000 permutations with seed 42. Negative PRR values are retained.

For metrics with bootstrap estimates, we use 1,000 resamples and report their standard deviation. Sensitivity results without bootstrap estimates are descriptive point estimates.

%% file: tables/additive/population_counts.tex
\begin{table}[htbp]
\centering\small
\setlength{\tabcolsep}{3pt}
\caption{Response counts used in the main evaluation. Excluded responses have no usable target or fail the task-specific inclusion rule. RAGTruth uses human annotation; the other tasks use the Llama judge.}
\label{tab:additive_population_counts}
\begin{tabular}{lllrrr}
\toprule
Task & Model & Labels & Evaluated & Errors & Excluded \\
\midrule
RAGTruth & Llama 2 7B & Human & 1,959 & 1,207 & 0 \\
RAGTruth & Llama 2 13B & Human & 1,959 & 1,110 & 0 \\
RAGTruth & Mistral 7B & Human & 1,959 & 1,308 & 0 \\
PreciseWikiQA & Llama 2 7B & Llama judge & 1,767 & 1,637 & 233 \\
PreciseWikiQA & Llama 2 13B & Llama judge & 1,923 & 1,672 & 77 \\
PreciseWikiQA & Mistral 7B & Llama judge & 1,955 & 1,686 & 45 \\
LongWiki & Llama 2 7B & Llama judge & 1,996 & 1,480 & 4 \\
LongWiki & Llama 2 13B & Llama judge & 1,999 & 1,729 & 1 \\
LongWiki & Mistral 7B & Llama judge & 1,997 & 1,785 & 3 \\
NonExistentRefusal & Llama 2 7B & Llama judge & 2,000 & 1,928 & 0 \\
NonExistentRefusal & Llama 2 13B & Llama judge & 2,000 & 1,665 & 0 \\
NonExistentRefusal & Mistral 7B & Llama judge & 2,000 & 1,875 & 0 \\
\bottomrule
\end{tabular}
\end{table}

%% file: tables/target_sensitivity_statistics.tex
\begin{table}[htbp]
\caption{LongWiki quality and claim counts among responses with a valid target. Precision is the fraction of supported response claims; Recall@32 is the number of supported claims divided by 32, capped at one. F1@32 is their harmonic mean. Retrieved passages are counted once per response; these are passage counts, not counts of reference claims.}
\label{tab:longwiki_target_components}
\centering
\small
\begin{tabular}{lrrr}
\toprule
Quantity & Llama 2 7B & Llama 2 13B & Mistral 7B \\
\midrule
Responses with F1 target & 1,996 & 1,999 & 1,997 \\
Mean precision & 0.3671 & 0.4176 & 0.2679 \\
Mean Recall@32 & 0.4056 & 0.2519 & 0.3258 \\
Mean F1@32 & 0.3796 & 0.3104 & 0.2906 \\
\midrule
Response claims: mean & 35.48 & 19.24 & 38.93 \\
Response claims: median & 36 & 19 & 39 \\
Response claims: min--max & 2--64 & 1--37 & 8--67 \\
Retrieved passages: mean & 17.32 & 14.65 & 19.32 \\
Retrieved passages: median & 16 & 14 & 18 \\
Retrieved passages: min--max & 3--69 & 3--49 & 3--81 \\
\bottomrule
\end{tabular}
\end{table}

\begin{table}[htbp]
\caption{PreciseWikiQA judgment counts and percentages among all 2,000 responses per model. Abstention overlaps correctness: one Llama 2 7B response is both correct and abstaining. The parser treats \texttt{No} as incorrect. Evaluated response counts, after excluding abstentions and unverifiable judgments, are in \Cref{tab:additive_population_counts}.}
\label{tab:precisewiki_target_counts}
\centering
\small
\begin{tabular}{lrrr}
\toprule
Judgment & Llama 2 7B & Llama 2 13B & Mistral 7B \\
\midrule
Correct & 131 (6.55\%) & 251 (12.55\%) & 269 (13.45\%) \\
Incorrect & 1,868 (93.40\%) & 1,747 (87.35\%) & 1,730 (86.50\%) \\
Unverifiable & 0 (0.00\%) & 1 (0.05\%) & 1 (0.05\%) \\
Incorrect alias: \texttt{No} & 1 (0.05\%) & 1 (0.05\%) & 0 (0.00\%) \\
Abstaining & 233 (11.65\%) & 76 (3.80\%) & 44 (2.20\%) \\
\bottomrule
\end{tabular}
\end{table}

%% file: appendices/results_detailed.tex
This appendix gives the complete results behind Section~5, followed by the target and control checks.

\subsection{Complete performance tables}
\label{apx:performance_tables}
The tables report each task and model separately. AlignScore uses reference information and is grouped separately from uncertainty estimators. P(True) Sampling was not run on LongWiki or NonExistentRefusal because of resource constraints; the other 45 estimators have complete coverage.
\input{tables/aggregate_metrics_all_models}
\input{tables/aggregate_metrics_mistral-7B-instruct}
\input{tables/aggregate_metrics_llama-2-7b-chat}
\input{tables/aggregate_metrics_llama-2-13b-chat}
\clearpage

\subsection{Agreement between estimator rankings}
\label{apx:ranking_transfer}
The first plots compare estimator rankings across models within each task; the second compare rankings across tasks for each model.
\begin{figure}[htbp]
    \centering
    \includegraphics[width=0.45\linewidth]{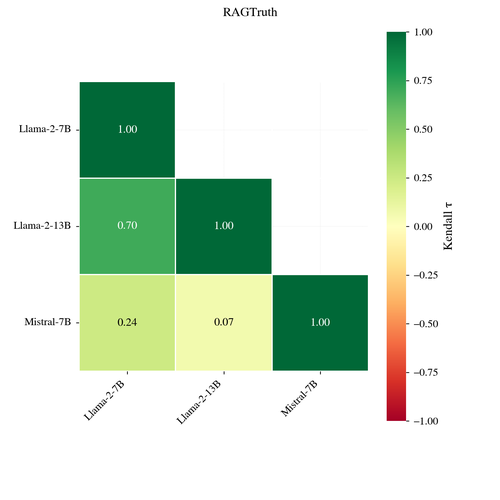}
    \includegraphics[width=0.45\linewidth]{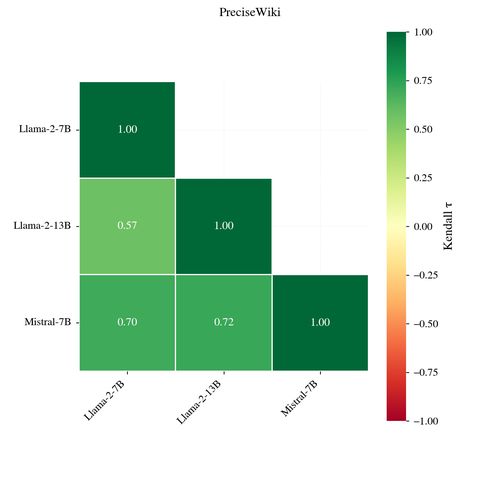}
    \includegraphics[width=0.45\linewidth]{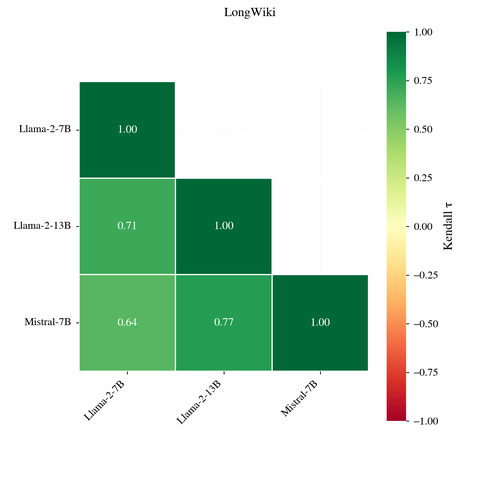}
    \includegraphics[width=0.45\linewidth]{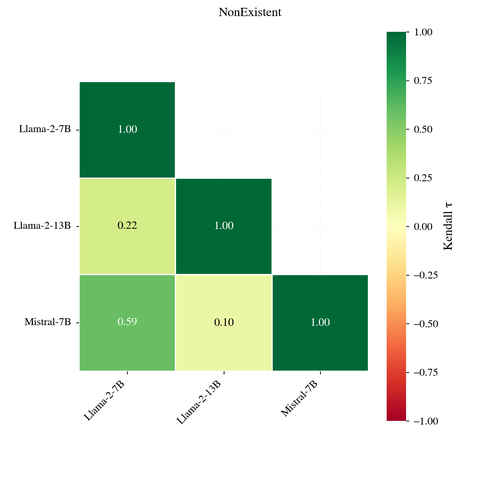}
    \caption{Pairwise Kendall's $\tau$ agreement between estimator rankings induced by ROC-AUC across models, shown separately for each task. Higher values indicate that the same estimators tend to rank similarly across models for a given hallucination type.}
    \label{fig:roc_rank_agreement_across_models}
\end{figure}
\clearpage
\begin{figure}[htbp]
    \centering
    \includegraphics[width=0.45\linewidth]{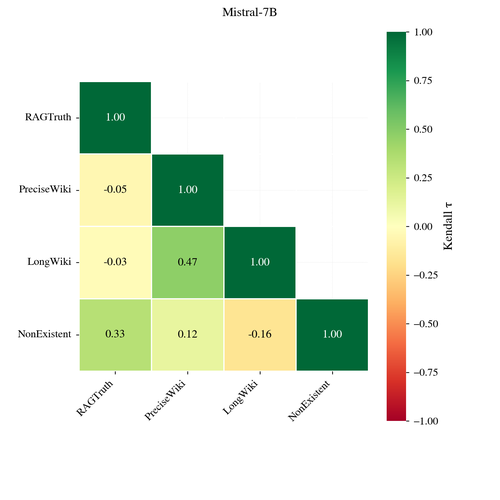}
    \includegraphics[width=0.45\linewidth]{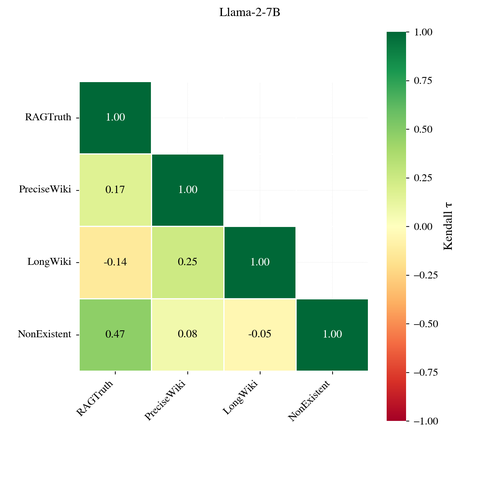}
    \includegraphics[width=0.45\linewidth]{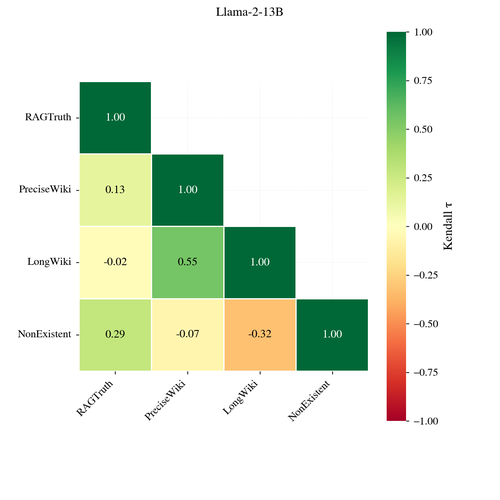}
    \caption{Pairwise Kendall's $\tau$ agreement between estimator rankings induced by ROC-AUC across tasks, shown separately for each model. Higher values indicate that the same estimators tend to rank similarly across hallucination types for a given model.}
    \label{fig:roc_rank_agreement_across_tasks}
\end{figure}
\clearpage
\paragraph{Agreement between evaluation metrics.}
This table holds the task and model fixed and compares rankings under different evaluation metrics.
\input{tables/evaluation_metric_agreement}
\clearpage
\subsection{Performance by task and estimator family}
\label{apx:performance_summaries}
The rank table and per-task plots expand the main performance summary. The family curves complete the main-text comparison with LongWiki and NonExistentRefusal.
\input{tables/estimator_ranks/rank_overall}
\clearpage
\begin{figure}[htbp]
    \centering
    \includegraphics[width=0.49\linewidth]{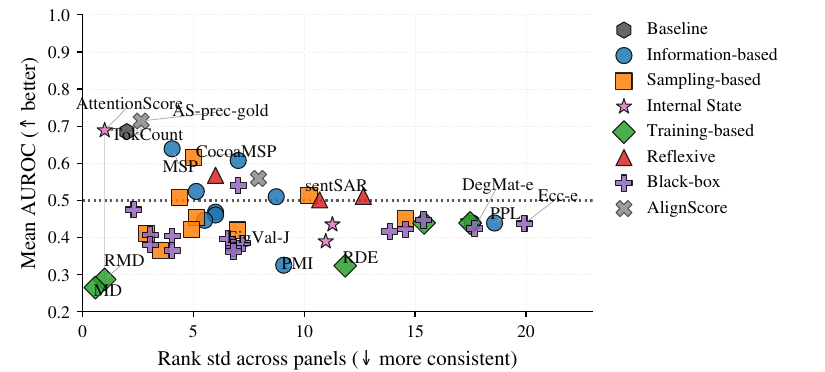}
    \includegraphics[width=0.49\linewidth]{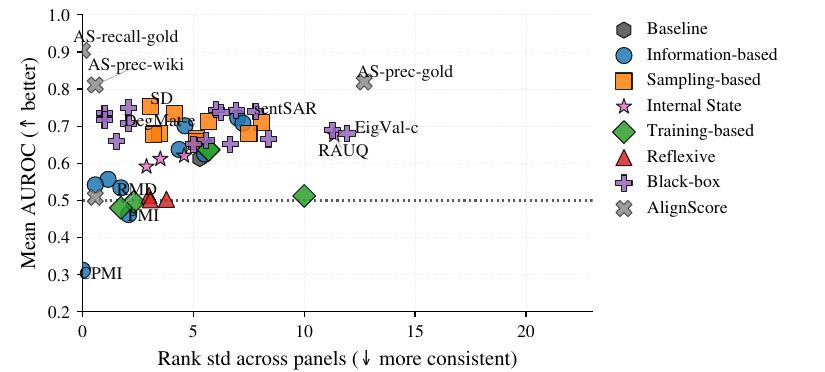}
    \includegraphics[width=0.49\linewidth]{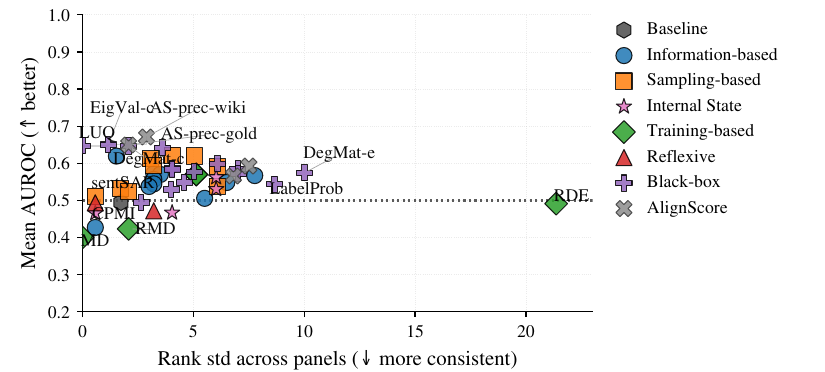}
    \includegraphics[width=0.49\linewidth]{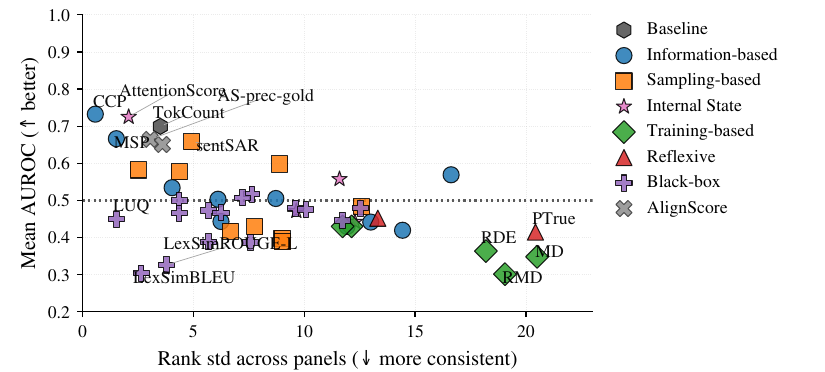}
    \caption{
    Performance--stability profiles for uncertainty estimators within
    each task, aggregated across the three models. Each point
    represents one estimator or AlignScore variant. The vertical axis
    reports mean ROC-AUC across models for the corresponding task, while
    the horizontal axis reports the standard deviation of the estimator's
    within-model rank; lower values indicate more stable rankings across
    models. The panels show, from left to right and top to bottom:
    \ragtruth, \precisewiki, \longwiki, and \nonexistent. AlignScore
    variants are reference-based comparators rather than uncertainty
    estimators.
    }
    \label{fig:stability_auroc_across_models_by_task}
\end{figure}
\clearpage
\begin{figure}
    \centering
    \includegraphics[]{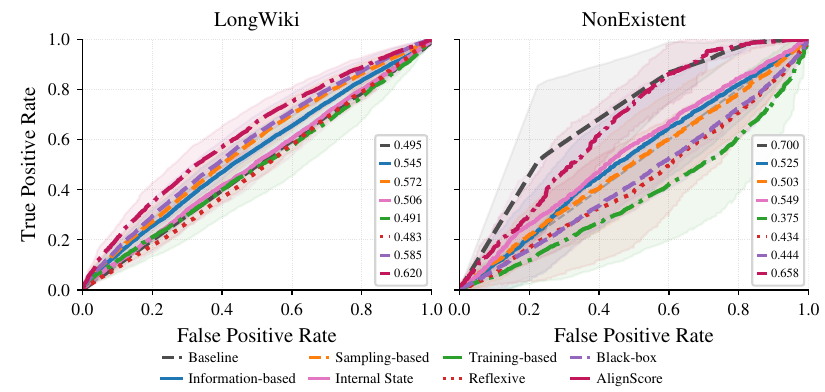}
    \caption{Family-level ROC aggregates per task, averaged across the three models. For each task--model panel, estimator ROC curves are first averaged within each family by taking the mean true positive rate at each false positive rate. For each task, these model-level family means are then averaged across models. Shaded bands show one standard deviation of these model-level family-mean true positive rates at each false positive rate; they reflect cross-model variability.} 
    \label{fig:family_roc_agg_per_task_lw_nr}
\end{figure}
\clearpage

\input{appendices/additive_comparisons}

%% file: tables/aggregate_metrics_all_models.tex
\begin{table*}[p]
\centering
\resizebox{\linewidth}{!}{\rotatebox{90}{\begin{minipage}{.95\textheight}
    \providecommand{\best}[1]{\mathbf{#1}}
    \caption{AUROC $\uparrow$, PRR $\uparrow$, RCE $\downarrow$ for \textbf{All Models} across tasks (higher AUROC/PRR and lower RCE are better; color shows numerical value). Uncertainty estimators and AlignScore variants are ranked separately: $\best{\mathrm{bold}}$ = best per column (ties marked); $\underline{\mathrm{underline}}$ = second-best (ties marked) within the uncertainty group; AlignScore group shows only the best.}
    \label{tab:metrics_all_models}
    \vskip 0.15in
    \centering
    \footnotesize
    \setlength{\tabcolsep}{4pt}
    \renewcommand{\arraystretch}{1.08}
    \resizebox{\linewidth}{!}{%
\begin{tabular}{cl|ccc|ccc|ccc|ccc}
    \toprule
     & \textbf{Estimator} & \multicolumn{3}{c|}{\textbf{RAGTruth}} & \multicolumn{3}{c|}{\textbf{PreciseWikiQA}} & \multicolumn{3}{c|}{\textbf{LongWiki}} & \multicolumn{3}{c}{\textbf{NonExistentRefusal}} \\
     &  & AUROC $\uparrow$ & PRR $\uparrow$ & RCE $\downarrow$ & AUROC $\uparrow$ & PRR $\uparrow$ & RCE $\downarrow$ & AUROC $\uparrow$ & PRR $\uparrow$ & RCE $\downarrow$ & AUROC $\uparrow$ & PRR $\uparrow$ & RCE $\downarrow$ \\
\bottomrule
    \end{tabular}%
    }
    \vskip -0.10in
\end{minipage}}}
\end{table*}

%% file: tables/aggregate_metrics_mistral-7B-instruct.tex
\begin{table*}[p]
\centering
\resizebox{\linewidth}{!}{\rotatebox{90}{\begin{minipage}{.95\textheight}
    \providecommand{\best}[1]{\mathbf{#1}}
    \caption{AUROC $\uparrow$, PRR $\uparrow$, RCE $\downarrow$ for \textbf{Mistral-7B-Instruct} across tasks (higher AUROC/PRR and lower RCE are better; color shows numerical value). Uncertainty estimators and AlignScore variants are ranked separately: $\best{\mathrm{bold}}$ = best per column (ties marked); $\underline{\mathrm{underline}}$ = second-best (ties marked) within the uncertainty group; AlignScore group shows only the best.}
    \label{tab:metrics_mistral_7b_instruct}
    \vskip 0.15in
    \centering
    \footnotesize
    \setlength{\tabcolsep}{4pt}
    \renewcommand{\arraystretch}{1.08}
    \resizebox{\linewidth}{!}{%
\begin{tabular}{cl|ccc|ccc|ccc|ccc}
    \toprule
     & \textbf{Estimator} & \multicolumn{3}{c|}{\textbf{RAGTruth}} & \multicolumn{3}{c|}{\textbf{PreciseWikiQA}} & \multicolumn{3}{c|}{\textbf{LongWiki}} & \multicolumn{3}{c}{\textbf{NonExistentRefusal}} \\
     &  & AUROC $\uparrow$ & PRR $\uparrow$ & RCE $\downarrow$ & AUROC $\uparrow$ & PRR $\uparrow$ & RCE $\downarrow$ & AUROC $\uparrow$ & PRR $\uparrow$ & RCE $\downarrow$ & AUROC $\uparrow$ & PRR $\uparrow$ & RCE $\downarrow$ \\
\bottomrule
    \end{tabular}%
    }
    \vskip -0.10in
\end{minipage}}}
\end{table*}

%% file: tables/aggregate_metrics_llama-2-7b-chat.tex
\begin{table*}[p]
\centering
\resizebox{\linewidth}{!}{\rotatebox{90}{\begin{minipage}{.95\textheight}
    \providecommand{\best}[1]{\mathbf{#1}}
    \caption{AUROC $\uparrow$, PRR $\uparrow$, RCE $\downarrow$ for \textbf{Llama-2-7B-Chat} across tasks (higher AUROC/PRR and lower RCE are better; color shows numerical value). Uncertainty estimators and AlignScore variants are ranked separately: $\best{\mathrm{bold}}$ = best per column (ties marked); $\underline{\mathrm{underline}}$ = second-best (ties marked) within the uncertainty group; AlignScore group shows only the best.}
    \label{tab:metrics_llama_2_7b_chat}
    \vskip 0.15in
    \centering
    \footnotesize
    \setlength{\tabcolsep}{4pt}
    \renewcommand{\arraystretch}{1.08}
    \resizebox{\linewidth}{!}{%
\begin{tabular}{cl|ccc|ccc|ccc|ccc}
    \toprule
     & \textbf{Estimator} & \multicolumn{3}{c|}{\textbf{RAGTruth}} & \multicolumn{3}{c|}{\textbf{PreciseWikiQA}} & \multicolumn{3}{c|}{\textbf{LongWiki}} & \multicolumn{3}{c}{\textbf{NonExistentRefusal}} \\
     &  & AUROC $\uparrow$ & PRR $\uparrow$ & RCE $\downarrow$ & AUROC $\uparrow$ & PRR $\uparrow$ & RCE $\downarrow$ & AUROC $\uparrow$ & PRR $\uparrow$ & RCE $\downarrow$ & AUROC $\uparrow$ & PRR $\uparrow$ & RCE $\downarrow$ \\
\bottomrule
    \end{tabular}%
    }
    \vskip -0.10in
\end{minipage}}}
\end{table*}

%% file: tables/aggregate_metrics_llama-2-13b-chat.tex
\begin{table*}[p]
\centering
\resizebox{\linewidth}{!}{\rotatebox{90}{\begin{minipage}{.95\textheight}
    \providecommand{\best}[1]{\mathbf{#1}}
    \caption{AUROC $\uparrow$, PRR $\uparrow$, RCE $\downarrow$ for \textbf{Llama-2-13B-Chat} across tasks (higher AUROC/PRR and lower RCE are better; color shows numerical value). Uncertainty estimators and AlignScore variants are ranked separately: $\best{\mathrm{bold}}$ = best per column (ties marked); $\underline{\mathrm{underline}}$ = second-best (ties marked) within the uncertainty group; AlignScore group shows only the best.}
    \label{tab:metrics_llama_2_13b_chat}
    \vskip 0.15in
    \centering
    \footnotesize
    \setlength{\tabcolsep}{4pt}
    \renewcommand{\arraystretch}{1.08}
    \resizebox{\linewidth}{!}{%
\begin{tabular}{cl|ccc|ccc|ccc|ccc}
    \toprule
     & \textbf{Estimator} & \multicolumn{3}{c|}{\textbf{RAGTruth}} & \multicolumn{3}{c|}{\textbf{PreciseWikiQA}} & \multicolumn{3}{c|}{\textbf{LongWiki}} & \multicolumn{3}{c}{\textbf{NonExistentRefusal}} \\
     &  & AUROC $\uparrow$ & PRR $\uparrow$ & RCE $\downarrow$ & AUROC $\uparrow$ & PRR $\uparrow$ & RCE $\downarrow$ & AUROC $\uparrow$ & PRR $\uparrow$ & RCE $\downarrow$ & AUROC $\uparrow$ & PRR $\uparrow$ & RCE $\downarrow$ \\
\bottomrule
    \end{tabular}%
    }
    \vskip -0.10in
\end{minipage}}}
\end{table*}

%% file: tables/evaluation_metric_agreement.tex
\begin{table}[t]
    \caption{Spearman correlation between estimator rankings induced by each pair of metrics, computed over uncertainty estimators. AlignScore variants are excluded. RCE is lower-is-better, so negative values indicate agreement. PRR is computed over the full rejection domain.}
    \label{tab:metric_agreement}
    \centering
    \small
    \resizebox{\linewidth}{!}{%
    \begin{tabular}{llccccc}
    \toprule
    Task & Model & AUROC--PRR & AUROC--AUPRC & AUROC--RCE & PRR--RCE & AUPRC--RCE \\
    \midrule
    RAGTruth & Llama 2 13B & .95 & .99 & -.96 & -.91 & -.95 \\
    -- & Llama 2 7B & .91 & .96 & -.94 & -.88 & -.91 \\
    -- & Mistral 7B & .95 & .94 & -.93 & -.90 & -.86 \\
    PreciseWikiQA & Llama 2 13B & .98 & .98 & -.92 & -.91 & -.94 \\
    -- & Llama 2 7B & .96 & .94 & -.72 & -.60 & -.79 \\
    -- & Mistral 7B & .97 & .98 & -.74 & -.72 & -.78 \\
    LongWiki & Llama 2 13B & .97 & .99 & -.92 & -.94 & -.92 \\
    -- & Llama 2 7B & .95 & .98 & -.90 & -.89 & -.87 \\
    -- & Mistral 7B & .94 & .96 & -.89 & -.92 & -.90 \\
    NonExistentRefusal & Llama 2 13B & .98 & .94 & -.92 & -.91 & -.83 \\
    -- & Llama 2 7B & .94 & .86 & -.93 & -.83 & -.80 \\
    -- & Mistral 7B & .98 & .99 & -.93 & -.90 & -.91 \\
    \bottomrule
    \end{tabular}%
    }
\end{table}

%% file: tables/estimator_ranks/rank_overall.tex
\begin{table}[t]
\centering
\caption{Within each benchmark panel (one task and one model), estimators are ranked by the metric value (rank 1 is best for that metric). Each cell gives the mean of that rank across panels in the slice, with a subscript giving +/- one standard deviation of the rank across those panels. Lower mean rank is better. Bold marks the lowest mean rank in the column separately within uncertainty estimators and within AlignScore variants (same convention as the aggregate metrics table). Underline marks second-lowest mean rank among uncertainty estimators only; AlignScore rows are not underlined for second place. The integer in parentheses counts how many panels in the slice had rank at most K (here K=3), i.e. were in the top-K for that metric within that panel. Slice: all tasks and models pooled.}
\begin{tabular}{@{}lccc@{}}
\toprule
\textbf{Estimator} & AUROC & PRR & RCE \\
\midrule
Output Token Count & 21.2$_{\pm 17.3}$ (3) & 23.2$_{\pm 20.1}$ (3) & 23.6$_{\pm 17.7}$ (3) \\
Maximum Seq. Probability (MSP) & 13.5$_{\pm 10.2}$ (1) & 14.9$_{\pm 12.3}$ (1) & 14.8$_{\pm 10.2}$ (2) \\
Perplexity (PPL) & 28.2$_{\pm 10.1}$ (0) & 29.7$_{\pm 8.8}$ (0) & 26.1$_{\pm 12.9}$ (0) \\
Mean Token Entropy (MTE) & 24.4$_{\pm 9.9}$ (0) & 25.7$_{\pm 10.5}$ (0) & 23.0$_{\pm 10.0}$ (0) \\
Pointwise Mutual Info. (PMI) & 40.8$_{\pm 7.8}$ (0) & 39.4$_{\pm 6.4}$ (0) & 37.2$_{\pm 10.2}$ (0) \\
Conditional PMI & 33.6$_{\pm 18.1}$ (0) & 32.7$_{\pm 19.2}$ (0) & 33.9$_{\pm 18.3}$ (1) \\
Self Certainty & 29.1$_{\pm 9.1}$ (0) & 27.3$_{\pm 11.0}$ (0) & 25.6$_{\pm 11.5}$ (0) \\
R\'enyi Divergence & 31.2$_{\pm 9.0}$ (0) & 29.1$_{\pm 10.6}$ (0) & 27.8$_{\pm 12.2}$ (0) \\
Fisher--Rao Distance & 27.8$_{\pm 11.3}$ (0) & 26.4$_{\pm 12.6}$ (0) & 25.8$_{\pm 10.2}$ (0) \\
Token SAR & 25.0$_{\pm 11.4}$ (0) & 25.8$_{\pm 10.1}$ (0) & 24.6$_{\pm 12.1}$ (0) \\
Claim-Conditioned Probability (CCP) & \textbf{8.7$_{\pm 5.7}$ (4)} & \underline{10.9$_{\pm 7.4}$ (3)} & \textbf{10.2$_{\pm 7.6}$ (3)} \\
Monte-Carlo Seq. Entropy & 25.3$_{\pm 10.0}$ (0) & 24.0$_{\pm 10.1}$ (0) & 22.0$_{\pm 11.7}$ (1) \\
Monte-Carlo Norm. Seq. Entropy & 32.0$_{\pm 7.8}$ (0) & 30.0$_{\pm 6.3}$ (0) & 29.4$_{\pm 12.3}$ (0) \\
Semantic Entropy & 23.6$_{\pm 9.4}$ (0) & 21.9$_{\pm 8.3}$ (0) & 23.0$_{\pm 13.1}$ (2) \\
Semantic Density & 20.0$_{\pm 11.5}$ (0) & 19.4$_{\pm 11.5}$ (0) & 20.6$_{\pm 11.8}$ (1) \\
Sentence SAR & 17.4$_{\pm 14.5}$ (2) & 17.8$_{\pm 14.3}$ (1) & 19.7$_{\pm 14.3}$ (1) \\
SAR & 22.8$_{\pm 9.9}$ (0) & 22.9$_{\pm 10.3}$ (0) & 25.7$_{\pm 11.7}$ (0) \\
Cocoa MSP & \underline{10.3$_{\pm 5.2}$ (1)} & \textbf{10.8$_{\pm 4.5}$ (1)} & \underline{14.2$_{\pm 9.4}$ (0)} \\
Cocoa PPL & 24.3$_{\pm 13.3}$ (0) & 24.1$_{\pm 12.2}$ (0) & 25.3$_{\pm 13.3}$ (1) \\
Cocoa MTE & 21.2$_{\pm 11.5}$ (1) & 21.2$_{\pm 11.8}$ (0) & 21.2$_{\pm 12.5}$ (1) \\
Attention Score & 22.2$_{\pm 19.4}$ (3) & 22.3$_{\pm 19.8}$ (5) & 23.0$_{\pm 18.6}$ (2) \\
RAUQ & 25.8$_{\pm 10.4}$ (0) & 25.6$_{\pm 9.7}$ (0) & 25.9$_{\pm 13.4}$ (1) \\
CSL & 33.5$_{\pm 7.5}$ (0) & 32.2$_{\pm 8.4}$ (0) & 32.5$_{\pm 7.5}$ (0) \\
EigenScore & 29.7$_{\pm 14.1}$ (0) & 27.3$_{\pm 13.1}$ (0) & 31.5$_{\pm 11.8}$ (0) \\
Mahalanobis Distance (MD) & 45.0$_{\pm 10.4}$ (0) & 44.4$_{\pm 10.8}$ (0) & 40.9$_{\pm 10.2}$ (0) \\
Relative MD (RMD) & 44.8$_{\pm 9.1}$ (0) & 43.5$_{\pm 9.3}$ (0) & 42.6$_{\pm 10.0}$ (0) \\
Robust Density Estimation (RDE) & 37.1$_{\pm 13.4}$ (0) & 38.7$_{\pm 11.9}$ (0) & 35.1$_{\pm 11.9}$ (0) \\
HUQ-MD & 28.8$_{\pm 10.0}$ (0) & 30.0$_{\pm 9.7}$ (0) & 25.9$_{\pm 13.8}$ (1) \\
HUQ-RMD & 29.0$_{\pm 9.4}$ (0) & 30.1$_{\pm 8.7}$ (0) & 26.2$_{\pm 13.5}$ (1) \\
P(True) & 35.5$_{\pm 15.0}$ (0) & 35.7$_{\pm 13.3}$ (0) & 33.7$_{\pm 13.5}$ (0) \\
P(True) Sampling & 31.2$_{\pm 16.2}$ (0) & 31.3$_{\pm 14.3}$ (0) & 30.7$_{\pm 16.9}$ (0) \\
P(True) Empirical & 31.2$_{\pm 15.2}$ (0) & 31.8$_{\pm 14.8}$ (0) & 31.3$_{\pm 15.3}$ (1) \\
NumSet & 23.9$_{\pm 10.5}$ (0) & 28.8$_{\pm 11.5}$ (0) & 27.2$_{\pm 11.2}$ (0) \\
LabelProb & 22.8$_{\pm 9.4}$ (0) & 22.6$_{\pm 9.1}$ (0) & 27.6$_{\pm 10.3}$ (0) \\
Kernel Language Entropy (KLE) & 20.5$_{\pm 11.0}$ (0) & 19.9$_{\pm 12.6}$ (0) & 22.5$_{\pm 12.6}$ (1) \\
EigValLap NLI (entail) & 21.1$_{\pm 11.2}$ (1) & 21.4$_{\pm 12.5}$ (0) & 21.5$_{\pm 9.7}$ (0) \\
EigValLap NLI (contra) & 19.3$_{\pm 12.1}$ (2) & 22.3$_{\pm 14.0}$ (2) & 21.0$_{\pm 14.0}$ (0) \\
EigValLap Jaccard & 31.3$_{\pm 10.1}$ (0) & 29.5$_{\pm 10.6}$ (0) & 28.2$_{\pm 14.9}$ (0) \\
Eccentricity NLI (entail) & 19.8$_{\pm 9.5}$ (0) & 20.6$_{\pm 10.1}$ (0) & 21.2$_{\pm 8.7}$ (0) \\
Eccentricity NLI (contra) & 17.6$_{\pm 11.9}$ (1) & 20.8$_{\pm 14.5}$ (1) & 18.5$_{\pm 13.4}$ (0) \\
Eccentricity Jaccard & 24.7$_{\pm 12.1}$ (0) & 25.5$_{\pm 13.6}$ (0) & 26.7$_{\pm 13.4}$ (0) \\
DegMat NLI (entail) & 21.5$_{\pm 12.7}$ (0) & 21.6$_{\pm 12.7}$ (0) & 18.2$_{\pm 12.6}$ (1) \\
DegMat NLI (contra) & 21.3$_{\pm 13.0}$ (1) & 23.9$_{\pm 14.9}$ (1) & 22.2$_{\pm 15.8}$ (2) \\
DegMat Jaccard & 29.3$_{\pm 9.6}$ (0) & 26.6$_{\pm 9.5}$ (0) & 26.5$_{\pm 13.0}$ (0) \\
LUQ & 21.2$_{\pm 14.9}$ (0) & 22.1$_{\pm 16.5}$ (1) & 23.6$_{\pm 13.0}$ (1) \\
Lexical Sim. (ROUGE-L) & 33.7$_{\pm 8.2}$ (0) & 31.5$_{\pm 8.0}$ (0) & 28.5$_{\pm 12.9}$ (0) \\
Lexical Sim. (BLEU) & 31.6$_{\pm 9.0}$ (0) & 29.3$_{\pm 9.5}$ (0) & 25.8$_{\pm 13.2}$ (1) \\
\midrule
AS-Precision(gold\_answer) & 5.7$_{\pm 6.1}$ (6) & \textbf{3.5$_{\pm 2.6}$ (8)} & 12.2$_{\pm 13.2}$ (3) \\
AS-Precision(wikipage) & \textbf{2.7$_{\pm 1.7}$ (5)} & 4.8$_{\pm 5.3}$ (4) & \textbf{8.4$_{\pm 8.7}$ (3)} \\
AS-Recall(gold\_answer) & 7.7$_{\pm 7.3}$ (5) & 7.0$_{\pm 6.2}$ (5) & 20.9$_{\pm 11.5}$ (0) \\
AS-Recall(wikipage) & 35.0$_{\pm 11.1}$ (0) & 33.5$_{\pm 9.5}$ (0) & 32.8$_{\pm 11.7}$ (0) \\
\bottomrule
\end{tabular}
\end{table}

%% file: appendices/additive_comparisons.tex
\subsection{Response length}
\label{apx:additive_length}\label{apx:length_protocol}
For the length comparison in Section~5, we compare estimator AUROC with token count on the same responses, then evaluate performance within five length groups.

For paired AUROC differences, both scores use the same 1,000 resampled sets of responses. Intervals are the 2.5th and 97.5th percentiles, with seed 42. The grouping identifier is exact input text because a source-document identifier is unavailable in the aligned inputs. All retained inputs are unique within each panel, so this reduces to response-level resampling. Dependence between different prompts from one source is not accounted for. All resamples in the reported RAGTruth comparisons contained both classes. The intervals describe each comparison separately; they do not establish simultaneous differences across all methods.

RAGTruth uses five approximately equal length groups without splitting tied token counts; both classes occur in every group. LongWiki and NonExistentRefusal responses concentrate at generation limits, preventing a comparable check. For example, 99.45\% of valid Llama 2 13B LongWiki responses reach the 256-token cap.

These groups retain some length variation and do not isolate a causal effect of length.
\begin{figure}[htbp]
\centering
\includegraphics[width=\linewidth]{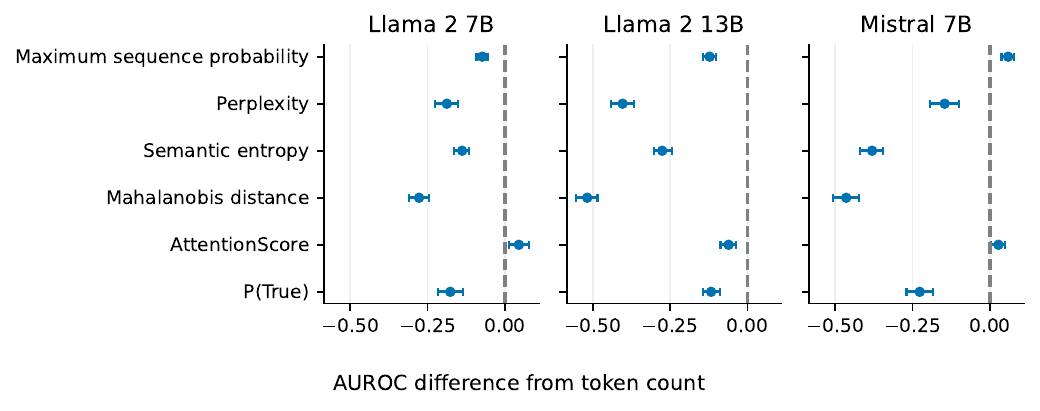}
\caption{RAGTruth AUROC difference from token count. Points to the right of zero favor the estimator. Bars are paired 95\% bootstrap intervals (1,000 resamples), calculated separately for each comparison.}
\label{fig:additive_paired_length}
\end{figure}
\begin{figure}[htbp]
\centering
\includegraphics[width=\linewidth]{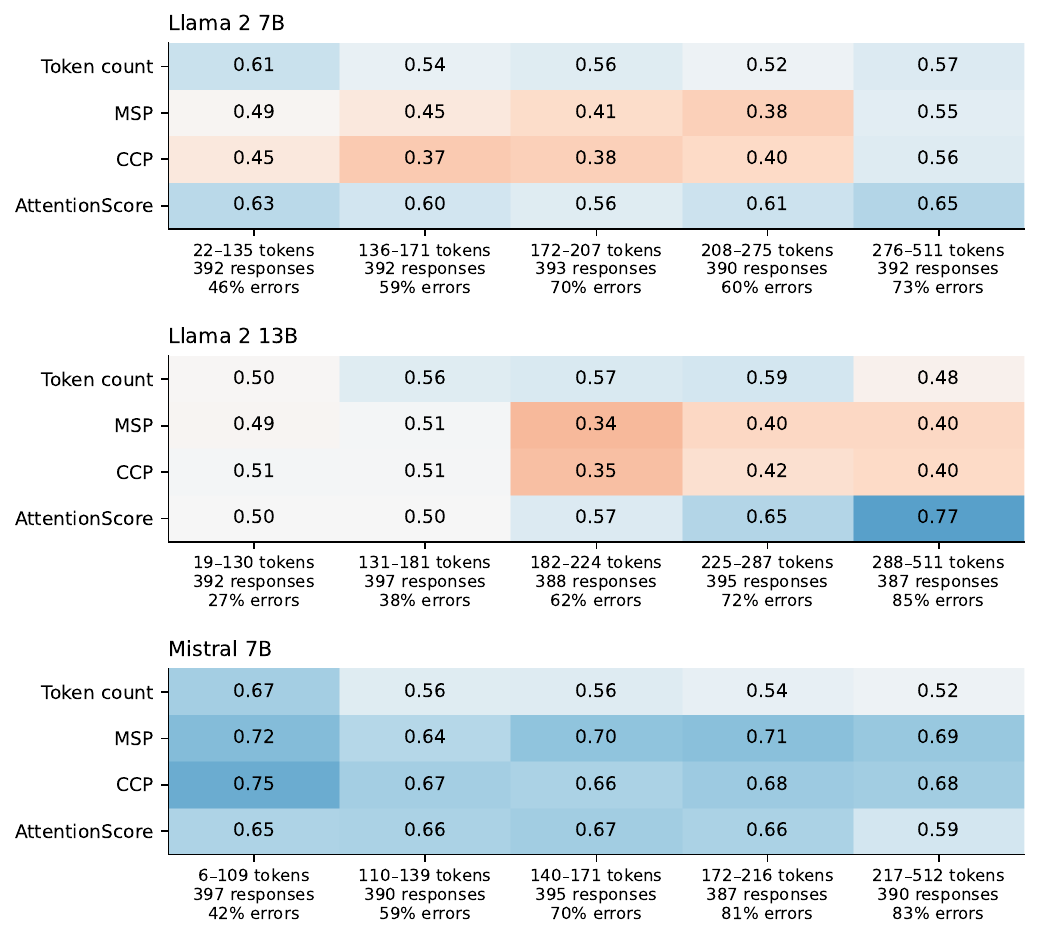}
\caption{RAGTruth AUROC within response-length groups, for the signals discussed in Section~5. Columns run from shortest to longest responses and give the token range, sample size and percentage of errors. AUROC 0.5 is chance; larger values are better. MSP: maximum sequence probability; CCP: claim-conditioned probability. Tied lengths stay together; both classes occur in every group.}
\label{fig:additive_length_ragtruth}
\end{figure}
\clearpage
\subsection{Changing the judge}
\label{apx:judge_results}\label{apx:control_protocols}
The judge comparison in Section~5 checks whether relative estimator performance depends on the primary HalluLens evaluator. We replace the Llama judge with Qwen2.5-72B-Instruct-AWQ while keeping generated responses and uncertainty scores fixed. On PreciseWikiQA, judge-specific exclusions can change the evaluated subset, so differences reflect both judgments and response selection. On LongWiki, extracted claims and retrieved evidence also remain fixed; only claim verification changes.

Judge replacement measures sensitivity, not label correctness. The quantized judge used here has not been independently validated against human judgments.

The nine task--model comparisons summarized in Section~5 have AUROC rank correlations from 0.906 to 0.995 under judge replacement. For LongWiki, agreement on continuous quality scores is weaker: Spearman correlations are 0.936, 0.930 and 0.431 for Llama 2 7B, Llama 2 13B and Mistral 7B. Mistral's binary labels agree on 71.86\% of responses. Thus, similar estimator rankings need not mean similar judgments of individual responses.
\input{tables/additive/judge_population_counts}
\clearpage

\input{appendices/review_subtype_checks}

\subsection{LongWiki: precision, recall and error thresholds}
\label{apx:longwiki_components}\label{apx:additive_thresholds}
F1@32 combines claim precision with a fixed quota of supported claims (\Cref{apx:target_processing}). For example, ten supported claims and no unsupported claims give precision 1 but F1@32 of 0.476. We keep responses and uncertainty scores fixed and compare the labels and estimator rankings obtained from precision and Recall@32 separately. Cohen's $\kappa$ measures label agreement after accounting for class frequencies; Spearman's $\rho$ measures agreement between AUROC rankings.
\input{tables/additive/longwiki_component_agreement}
\paragraph{Changing the F1@32 threshold.}
The tables below vary the cutoff used to label a response as an error for AUROC. Counts are reported at every audited threshold; performance is shown only where computed. PRR and RCE retain the continuous F1@32 target. This checks threshold sensitivity without selecting a preferred cutoff.
\input{tables/additive/longwiki_threshold_counts}
\input{tables/additive/longwiki_threshold_performance}

%% file: tables/additive/judge_population_counts.tex
\begin{table}[htbp]
\centering\small
\caption{PreciseWikiQA response selection under the two judges. Both evaluate the same 2,000 generated responses per model; each excludes its own abstentions and unverifiable judgments. Errors are incorrect answers among the retained responses.}
\label{tab:judge_population_counts}
\begin{tabular}{llrrr}
\toprule
Model & Judge & Evaluated & Errors & Excluded \\
\midrule
Llama 2 7B & Llama & 1,767 & 1,637 & 233 \\
Llama 2 7B & Qwen & 1,489 & 1,334 & 511 \\
Llama 2 13B & Llama & 1,923 & 1,672 & 77 \\
Llama 2 13B & Qwen & 1,718 & 1,443 & 282 \\
Mistral 7B & Llama & 1,955 & 1,686 & 45 \\
Mistral 7B & Qwen & 1,870 & 1,596 & 130 \\
\bottomrule
\end{tabular}
\end{table}

%% file: appendices/review_subtype_checks.tex
\subsection{RAGTruth: error types and proportions}
\label{apx:review_subtype_checks}\label{apx:target_protocol}\label{sec:results_sensitivity}
The comparisons in Section~5 use the same 1,959 responses and 45 estimators per model, changing only the definition of an error.

For RAGTruth, the four binary targets indicate any annotated span, conflict with the context, baseless information unsupported by it, and the presence of both subtypes. Conflict includes the evident and subtle annotations grouped by the processed dataset. Subtypes are inclusive: a response with both is positive for each subtype. We also measure prevalence within each response as the proportion of tokens covered by the union of annotated spans. Overlapping spans are counted once. This is distinct from the fraction of responses labeled as errors. The continuous quality target is one minus this token proportion.

The distribution plot separates responses with one subtype from those with both; the evaluation targets include co-occurring subtypes as described above.
\begin{figure}[htbp]
\centering
\includegraphics[width=\linewidth]{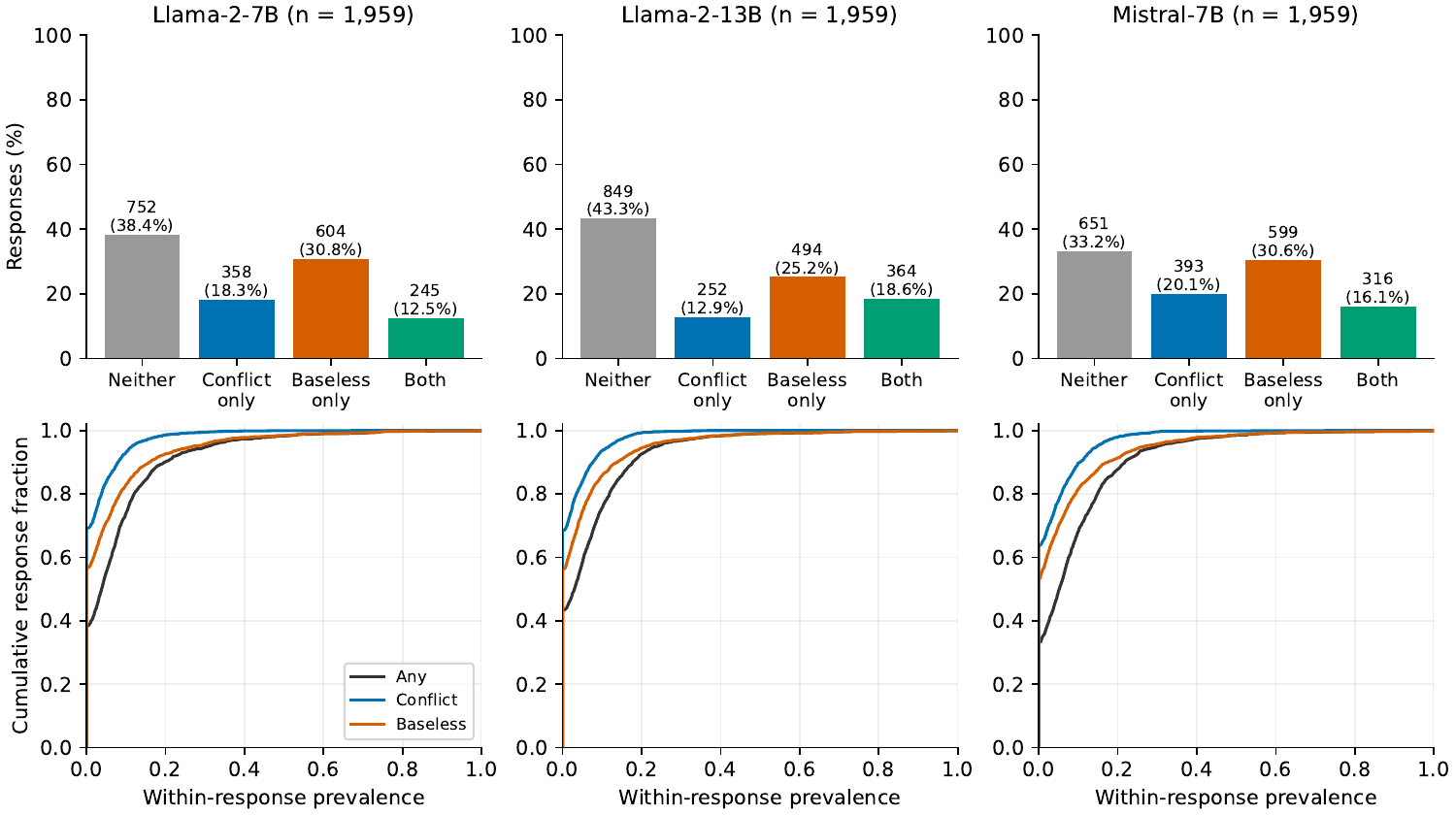}
\caption{RAGTruth subtype distributions for each model. Top: mutually exclusive occurrence categories with counts and percentages of all 1,959 responses. Bottom: empirical cumulative distributions of within-response prevalence, defined as the fraction of response tokens covered by annotated spans. All responses are included; the jump at zero represents responses without that annotation. Any combines both subtypes; overlapping spans are counted once. These distributions describe the targets rather than estimator performance.}
\label{fig:ragtruth_target_distribution}
\end{figure}
\begin{table}[htbp]
\centering\small
\setlength{\tabcolsep}{3pt}
\caption{Best observed metric values across the 45 estimators for each occurrence target. Each row contains 1,959 responses; Errors counts positive labels and Error rate gives their fraction (the constant-score AUPRC baseline). Maxima are reported for AUROC, AUPRC and PRR; minima for RCE. The best value in each column can come from a different estimator and is not a validated selection rule.}
\label{tab:review_subtype_metrics}
\begin{tabular}{llrrrrrr}
\toprule
Model & Target & Errors & Error rate & AUROC & AUPRC & PRR & RCE \\
\midrule
Llama 2 7B & Any & 1207 & 0.616 & 0.647 & 0.701 & 0.311 & 0.148 \\
Llama 2 7B & Conflict & 603 & 0.308 & 0.678 & 0.426 & 0.456 & 0.093 \\
Llama 2 7B & Baseless & 849 & 0.433 & 0.584 & 0.516 & 0.224 & 0.132 \\
Llama 2 7B & Both & 245 & 0.125 & 0.689 & 0.192 & 0.538 & 0.093 \\
Llama 2 13B & Any & 1110 & 0.567 & 0.690 & 0.699 & 0.388 & 0.070 \\
Llama 2 13B & Conflict & 616 & 0.314 & 0.751 & 0.515 & 0.602 & 0.070 \\
Llama 2 13B & Baseless & 858 & 0.438 & 0.673 & 0.558 & 0.449 & 0.040 \\
Llama 2 13B & Both & 364 & 0.186 & 0.771 & 0.353 & 0.692 & 0.055 \\
Mistral 7B & Any & 1308 & 0.668 & 0.761 & 0.842 & 0.560 & 0.033 \\
Mistral 7B & Conflict & 709 & 0.362 & 0.724 & 0.544 & 0.570 & 0.025 \\
Mistral 7B & Baseless & 915 & 0.467 & 0.684 & 0.629 & 0.464 & 0.025 \\
Mistral 7B & Both & 316 & 0.161 & 0.767 & 0.344 & 0.665 & 0.040 \\
\bottomrule
\end{tabular}
\end{table}
\input{tables/additive/target_metric_agreement}
PRR@50 restricts rejection to half the responses, as a sensitivity check on the integration range. Main results use the full range. Ranking estimators by direct score--target correlation gives positive agreement between subtypes, so the reversal does not apply to every evaluation criterion.

\paragraph{Changing the minimum annotated proportion.}
The following check requires an annotated span of any length, or at least 10\% or 25\% of response tokens to be annotated. This changes the binary error labels used for AUROC; the continuous targets used for PRR and RCE remain unchanged. The thresholds are sensitivity checks, not selected operating points.
\input{tables/additive/ragtruth_threshold_counts}
\begin{figure}[htbp]
\centering
\includegraphics[width=\linewidth]{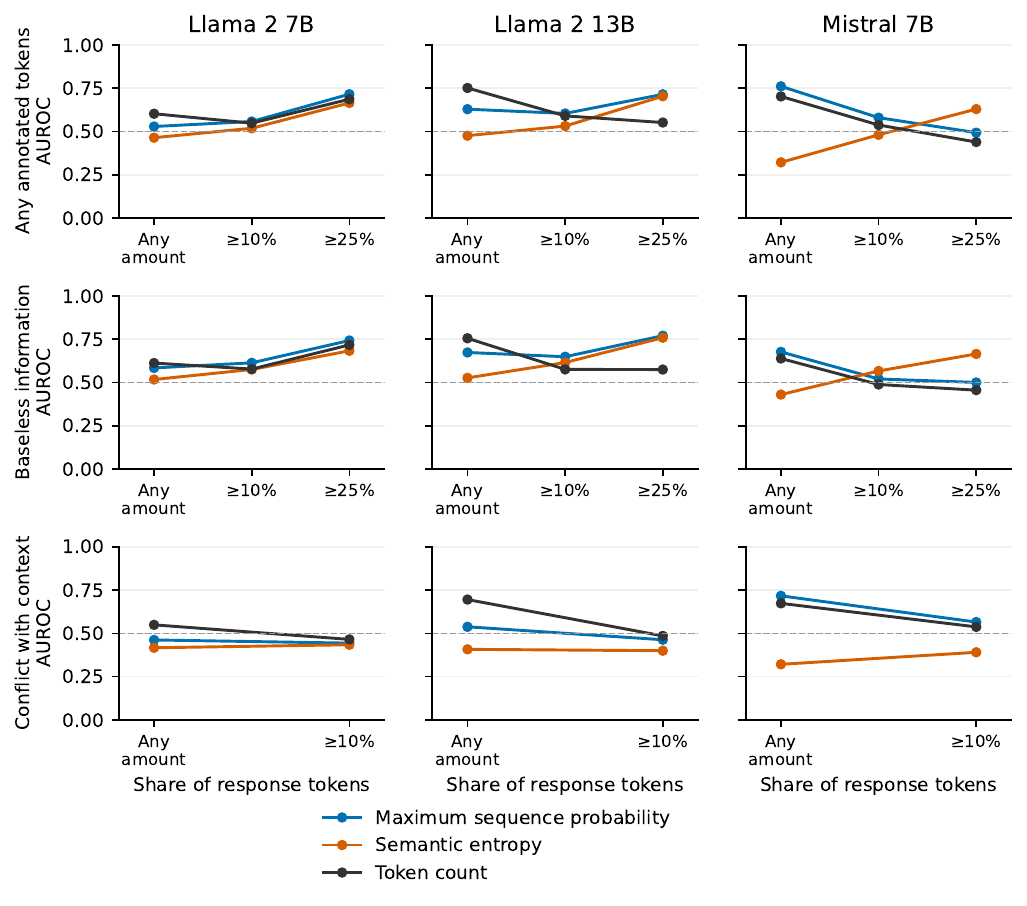}
\caption{RAGTruth AUROC for three fixed examples as the error definition changes. ``Any amount'' means at least one annotated span of the indicated type; 10\% and 25\% require that share of response tokens to be annotated. Rows distinguish all annotations, baseless information, and conflict. The dashed line marks chance (0.5). Lines connect computed settings only; conflict was not evaluated at 25\%. Class counts are in \Cref{tab:ragtruth_threshold_counts}.}
\label{fig:additive_thresholds_ragtruth}
\end{figure}
\clearpage

%% file: tables/additive/target_metric_agreement.tex
\begin{table}[htbp]
\caption{Spearman agreement between conflict and baseless estimator rankings, using the same 1,959 responses and 45 estimators per model. PRR integrates over the full rejection range; PRR@50 permits rejection of at most half the responses. RCE uses native quality. The last row in each block ranks estimators by their Spearman association with the binary error or token-share target; it is not a direct correlation between the two targets. For prevalence, the top-five PRR sets have no overlap for either rejection range.}
\label{tab:positioning_target_metrics}
\centering\small
\begin{tabular}{llrrr}
\toprule
Target & Evaluation & Llama 2 7B & Llama 2 13B & Mistral 7B \\
\midrule
Occurrence & AUROC & 0.613 & 0.846 & 0.933 \\
Occurrence & AUPRC & 0.519 & 0.857 & 0.888 \\
Occurrence & PRR & 0.679 & 0.821 & 0.921 \\
Occurrence & PRR@50 & 0.546 & 0.849 & 0.901 \\
Occurrence & RCE & 0.511 & 0.878 & 0.912 \\
Occurrence & Score--target $\rho$ & 0.591 & 0.843 & 0.933 \\
Prevalence & PRR & -0.251 & -0.123 & -0.561 \\
Prevalence & PRR@50 & -0.425 & -0.544 & -0.604 \\
Prevalence & RCE & -0.514 & -0.222 & -0.528 \\
Prevalence & Score--target $\rho$ & 0.179 & 0.591 & 0.716 \\
\bottomrule
\end{tabular}
\end{table}

%% file: tables/additive/ragtruth_threshold_counts.tex
\begin{table}[htbp]
\centering\small
\caption{RAGTruth error counts out of 1,959 responses per model for the thresholds in \Cref{fig:additive_thresholds_ragtruth}. Remaining responses are non-errors. Percentages in the threshold column refer to the share of annotated response tokens, not the fraction of erroneous responses.}
\label{tab:ragtruth_threshold_counts}
\begin{tabular}{llrrr}
\toprule
Annotation & Token share & Llama 2 7B & Llama 2 13B & Mistral 7B \\
\midrule
Any annotated tokens & Any amount & 1207 & 1110 & 1308 \\
Any annotated tokens & $\geq10\%$ & 516 & 479 & 624 \\
Any annotated tokens & $\geq25\%$ & 134 & 90 & 150 \\
Baseless information & Any amount & 849 & 858 & 915 \\
Baseless information & $\geq10\%$ & 338 & 279 & 364 \\
Baseless information & $\geq25\%$ & 110 & 74 & 114 \\
Conflict with context & Any amount & 603 & 616 & 709 \\
Conflict with context & $\geq10\%$ & 137 & 126 & 201 \\
\bottomrule
\end{tabular}
\end{table}

%% file: tables/additive/longwiki_component_agreement.tex
\begin{table}[htbp]
\centering\small
\caption{LongWiki precision versus Recall@32 at threshold 0.5. Errors have quality below 0.5. Label agreement is Cohen's $\kappa$; rank agreement is Spearman's $\rho$ between AUROC rankings of 45 estimators on the same responses. The two agreement columns concern different objects: response labels and estimator rankings.}
\label{tab:longwiki_component_agreement}
\begin{tabular}{lrrrrr}
\toprule
Model & Responses & \multicolumn{2}{c}{Errors} & Label & Rank \\
 & & Precision & Recall@32 & agreement & agreement \\
\midrule
Llama 2 7B & 1996 & 1506 & 1336 & 0.588 & 0.903 \\
Llama 2 13B & 1999 & 1262 & 1879 & 0.192 & 0.961 \\
Mistral 7B & 1997 & 1837 & 1617 & 0.457 & 0.844 \\
\bottomrule
\end{tabular}
\end{table}

%% file: tables/additive/longwiki_threshold_counts.tex
\begin{table}[htbp]
\caption{LongWiki error / non-error counts under the strict rule F1@32 $<t$. Responses without an F1 target are excluded. These are target counts, not estimator performance estimates. At $t=0.9$, AUROC is undefined for Llama 2 13B and Mistral 7B because only one class remains.}
\label{tab:longwiki_threshold_counts}
\centering
\small
\begin{tabular}{lrrr}
\toprule
Threshold & Llama 2 7B & Llama 2 13B & Mistral 7B \\
\midrule
0.1 & 124 / 1,872 & 202 / 1,797 & 216 / 1,781 \\
0.3 & 747 / 1,249 & 993 / 1,006 & 1,128 / 869 \\
0.5 & 1,480 / 516 & 1,729 / 270 & 1,785 / 212 \\
0.7 & 1,869 / 127 & 1,973 / 26 & 1,967 / 30 \\
0.9 & 1,992 / 4 & 1,999 / 0 & 1,997 / 0 \\
\bottomrule
\end{tabular}
\end{table}

%% file: tables/additive/longwiki_threshold_performance.tex
\begin{table}[htbp]
\centering\small
\caption{LongWiki AUROC as the F1@32 error threshold changes. A response is an error when F1@32 is strictly below the threshold. These three fixed examples illustrate threshold sensitivity; they are not selected winners. Class counts are in \Cref{tab:longwiki_threshold_counts}.}
\label{tab:longwiki_threshold_performance}
\begin{tabular}{llrrr}
\toprule
Model & Threshold & MSP & Semantic entropy & Token count \\
\midrule
Llama 2 7B & 0.3 & 0.540 & 0.506 & 0.469 \\
Llama 2 7B & 0.5 & 0.559 & 0.509 & 0.483 \\
Llama 2 7B & 0.7 & 0.660 & 0.582 & 0.478 \\
Llama 2 13B & 0.3 & 0.562 & 0.558 & 0.499 \\
Llama 2 13B & 0.5 & 0.589 & 0.577 & 0.499 \\
Llama 2 13B & 0.7 & 0.566 & 0.678 & 0.497 \\
Mistral 7B & 0.3 & 0.513 & 0.510 & 0.483 \\
Mistral 7B & 0.5 & 0.522 & 0.530 & 0.502 \\
Mistral 7B & 0.7 & 0.466 & 0.421 & 0.520 \\
\bottomrule
\end{tabular}
\par\smallskip\footnotesize MSP: maximum sequence probability.
\end{table}

%% file: references.bib
@article{vashurin-etal-2025-benchmarking,
    title = "Benchmarking Uncertainty Quantification Methods for Large Language Models with {LM}-Polygraph",
    author = "Vashurin, Roman  and
      Fadeeva, Ekaterina  and
      Vazhentsev, Artem  and
      Rvanova, Lyudmila  and
      Vasilev, Daniil  and
      Tsvigun, Akim  and
      Petrakov, Sergey  and
      Xing, Rui  and
      Sadallah, Abdelrahman  and
      Grishchenkov, Kirill  and
      Panchenko, Alexander  and
      Baldwin, Timothy  and
      Nakov, Preslav  and
      Panov, Maxim  and
      Shelmanov, Artem",
    journal = "Transactions of the Association for Computational Linguistics",
    volume = "13",
    year = "2025",
    naddress = "Cambridge, MA",
    npublisher = "MIT Press",
    url = "https://aclanthology.org/2025.tacl-1.11/",
    doi = "10.1162/tacl_a_00737",
    pages = "220--248"
}

@inproceedings{bakman-etal-2025-reconsidering,
    title = "Reconsidering {LLM} Uncertainty Estimation Methods in the Wild",
    author = "Bakman, Yavuz Faruk  and
      Yaldiz, Duygu Nur  and
      Kang, Sungmin  and
      Zhang, Tuo  and
      Buyukates, Baturalp  and
      Avestimehr, Salman  and
      Karimireddy, Sai Praneeth",
    neditor = "Che, Wanxiang  and
      Nabende, Joyce  and
      Shutova, Ekaterina  and
      Pilehvar, Mohammad Taher",
    booktitle = "Proceedings of the 63rd Annual Meeting of the Association for Computational Linguistics (Volume 1: Long Papers)",
    month = jul,
    year = "2025",
    naddress = "Vienna, Austria",
    npublisher = "Association for Computational Linguistics",
    url = "https://aclanthology.org/2025.acl-long.1429/",
    doi = "10.18653/v1/2025.acl-long.1429",
    pages = "29531--29556",
    ISBN = "979-8-89176-251-0"
}

@inproceedings{bang-etal-2025-hallulens,
    title = "{H}allu{L}ens: {LLM} Hallucination Benchmark",
    author = "Bang, Yejin  and
      Ji, Ziwei  and
      Schelten, Alan  and
      Hartshorn, Anthony  and
      Fowler, Tara  and
      Zhang, Cheng  and
      Cancedda, Nicola  and
      Fung, Pascale",
    neditor = "Che, Wanxiang  and
      Nabende, Joyce  and
      Shutova, Ekaterina  and
      Pilehvar, Mohammad Taher",
    booktitle = "Proceedings of the 63rd Annual Meeting of the Association for Computational Linguistics (Volume 1: Long Papers)",
    month = jul,
    year = "2025",
    naddress = "Vienna, Austria",
    npublisher = "Association for Computational Linguistics",
    url = "https://aclanthology.org/2025.acl-long.1176/",
    doi = "10.18653/v1/2025.acl-long.1176",
    pages = "24128--24156",
    ISBN = "979-8-89176-251-0"
}

@inproceedings{niu-etal-2024-ragtruth,
    title = "{RAGT}ruth: A Hallucination Corpus for Developing Trustworthy Retrieval-Augmented Language Models",
    author = "Niu, Cheng  and
      Wu, Yuanhao  and
      Zhu, Juno  and
      Xu, Siliang  and
      Shum, KaShun  and
      Zhong, Randy  and
      Song, Juntong  and
      Zhang, Tong",
    neditor = "Ku, Lun-Wei  and
      Martins, Andre  and
      Srikumar, Vivek",
    booktitle = "Proceedings of the 62nd Annual Meeting of the Association for Computational Linguistics (Volume 1: Long Papers)",
    month = aug,
    year = "2024",
    naddress = "Bangkok, Thailand",
    npublisher = "Association for Computational Linguistics",
    url = "https://aclanthology.org/2024.acl-long.585/",
    doi = "10.18653/v1/2024.acl-long.585",
    pages = "10862--10878"
}

@article{kang2025uncertainty,
      title={Uncertainty Quantification for Hallucination Detection in Large Language Models: Foundations, Methodology, and Future Directions}, 
      author={Sungmin Kang and Yavuz Faruk Bakman and Duygu Nur Yaldiz and Baturalp Buyukates and Salman Avestimehr},
      year={2025},
      journal={arXiv 2510.12040},
      archivePrefix={arXiv},
      primaryClass={cs.CL},
      url={https://arxiv.org/abs/2510.12040}, 
}

@article{fadeeva2023lm,
  title={LM-polygraph: Uncertainty estimation for language models},
  author={Fadeeva, Ekaterina and Vashurin, Roman and Tsvigun, Akim and Vazhentsev, Artem and Petrakov, Sergey and Fedyanin, Kirill and Vasilev, Daniil and Goncharova, Elizaveta and Panchenko, Alexander and Panov, Maxim and others},
  journal={Proceedings of the Conference on Empirical Methods in Natural Language
Processing: System Demonstrations},
  year={2023}
}

@article{fomicheva2020unsupervised,
  title={Unsupervised quality estimation for neural machine translation},
  author={Fomicheva, Marina and Sun, Shuo and Yankovskaya, Lisa and Blain, Fr{\'e}d{\'e}ric and Guzm{\'a}n, Francisco and Fishel, Mark and Aletras, Nikolaos and Chaudhary, Vishrav and Specia, Lucia},
  journal={Transactions of the Association for Computational Linguistics},
  volume={8},
  pages={539--555},
  year={2020},
  publisher={MIT Press One Rogers Street, Cambridge, MA 02142-1209, USA journals-info~…}
}

@article{malinin2020uncertainty,
  title={Uncertainty estimation in autoregressive structured prediction},
  author={Malinin, Andrey and Gales, Mark},
  journal={International Conference on Learning Representations},
  year={2021}
}

@article{kuhn2023semantic,
  title={Semantic uncertainty: Linguistic invariances for uncertainty estimation in natural language generation},
  author={Kuhn, Lorenz and Gal, Yarin and Farquhar, Sebastian},
  journal={International Conference on Learning Representations},
  year={2023}
}

@article{lin2023generating,
  title={Generating with confidence: Uncertainty quantification for black-box large language models},
  author={Lin, Zhen and Trivedi, Shubhendu and Sun, Jimeng},
  journal={Findings of ACL},
  year={2023}
}

@article{zha2023alignscore,
  title={AlignScore: Evaluating factual consistency with a unified alignment function},
  author={Zha, Yuheng and Yang, Yichi and Li, Ruichen and Hu, Zhiting},
  journal={arXiv preprint arXiv:2305.16739},
  year={2023}
}

@article{kadavath2022language,
  title={Language models (mostly) know what they know},
  author={Kadavath, Saurav and Conerly, Tom and Askell, Amanda and Henighan, Tom and Drain, Dawn and Perez, Ethan and Schiefer, Nicholas and Hatfield-Dodds, Zac and DasSarma, Nova and Tran-Johnson, Eli and others},
  journal={arXiv preprint arXiv:2207.05221},
  year={2022}
}

@inproceedings{darrin2023rainproof,
  title={RainProof: An umbrella to shield text generator from out-of-distribution data},
  author={Darrin, Maxime and Piantanida, Pablo and Colombo, Pierre},
  booktitle={Proceedings of the Conference on Empirical Methods in Natural Language Processing},
  pages={5831--5857},
  year={2023}
}

@article{lee2018simple,
  title={A simple unified framework for detecting out-of-distribution samples and adversarial attacks},
  author={Lee, Kimin and Lee, Kibok and Lee, Honglak and Shin, Jinwoo},
  journal={Advances in neural information processing systems},
  volume={31},
  year={2018}
}

@inproceedings{zhang_luq_2024,
    naddress = {Miami, Florida, USA},
    title = {{LUQ}: {Long}-text {Uncertainty} {Quantification} for {LLMs}},
    shorttitle = {{LUQ}},
    url = {https://aclanthology.org/2024.emnlp-main.299/},
    doi = {10.18653/v1/2024.emnlp-main.299},
    urldate = {2025-01-20},
    booktitle = {Proceedings of the {Conference} on {Empirical} {Methods} in {Natural} {Language} {Processing}},
    npublisher = {Association for Computational Linguistics},
    author = {Zhang, Caiqi and Liu, Fangyu and Basaldella, Marco and Collier, Nigel},
    editor = {Al-Onaizan, Yaser and Bansal, Mohit and Chen, Yun-Nung},
    month = nov,
    year = {2024},
    pages = {5244--5262},
}

@inproceedings{qiu_semantic_2024,
    title = {Semantic {Density}: {Uncertainty} {Quantification} for {Large} {Language} {Models} through {Confidence} {Measurement} in {Semantic} {Space}},
    shorttitle = {Semantic {Density}},
    url = {https://openreview.net/forum?id=LOH6qzI7T6},
    language = {en},
    urldate = {2025-01-24},
    booktitle = {Advances in {Neural} {Information} {Processing} {Systems}},
    author = {Qiu, Xin and Miikkulainen, Risto},
    month = nov,
    year = {2024},
}

@inproceedings{huang_uncertainty_2024,
    naddress = {Miami, Florida, USA},
    title = {Uncertainty in {Language} {Models}: {Assessment} through {Rank}-{Calibration}},
    shorttitle = {Uncertainty in {Language} {Models}},
    url = {https://aclanthology.org/2024.emnlp-main.18/},
    doi = {10.18653/v1/2024.emnlp-main.18},
    urldate = {2025-01-20},
    booktitle = {Proceedings of the {Conference} on {Empirical} {Methods} in {Natural} {Language} {Processing}},
    npublisher = {Association for Computational Linguistics},
    author = {Huang, Xinmeng and Li, Shuo and Yu, Mengxin and Sesia, Matteo and Hassani, Hamed and Lee, Insup and Bastani, Osbert and Dobriban, Edgar},
    neditor = {Al-Onaizan, Yaser and Bansal, Mohit and Chen, Yun-Nung},
    month = nov,
    year = {2024},
    pages = {284--312},
}

@misc{fan_halluhard_2026,
    title = {{HalluHard}: {A} {Hard} {Multi}-{Turn} {Hallucination} {Benchmark}},
    shorttitle = {{HalluHard}},
    url = {http://arxiv.org/abs/2602.01031},
    doi = {10.48550/arXiv.2602.01031},
    urldate = {2026-03-20},
    publisher = {arXiv},
    author = {Fan, Dongyang and Delsad, Sebastien and Flammarion, Nicolas and Andriushchenko, Maksym},
    month = feb,
    year = {2026},
    note = {arXiv:2602.01031 [cs]},
}

@misc{bengio_international_2025,
    title = {International {AI} {Safety} {Report}},
    url = {http://arxiv.org/abs/2501.17805},
    doi = {10.48550/arXiv.2501.17805},
    urldate = {2026-03-20},
    publisher = {arXiv},
    author = {Bengio, Yoshua and Mindermann, Sören and Privitera, Daniel and Besiroglu, Tamay and Bommasani, Rishi and Casper, Stephen and Choi, Yejin and Fox, Philip and Garfinkel, Ben and Goldfarb, Danielle and Heidari, Hoda and Ho, Anson and Kapoor, Sayash and Khalatbari, Leila and Longpre, Shayne and Manning, Sam and Mavroudis, Vasilios and Mazeika, Mantas and Michael, Julian and Newman, Jessica and Ng, Kwan Yee and Okolo, Chinasa T. and Raji, Deborah and Sastry, Girish and Seger, Elizabeth and Skeadas, Theodora and South, Tobin and Strubell, Emma and Tramèr, Florian and Velasco, Lucia and Wheeler, Nicole and Acemoglu, Daron and Adekanmbi, Olubayo and Dalrymple, David and Dietterich, Thomas G. and Felten, Edward W. and Fung, Pascale and Gourinchas, Pierre-Olivier and Heintz, Fredrik and Hinton, Geoffrey and Jennings, Nick and Krause, Andreas and Leavy, Susan and Liang, Percy and Ludermir, Teresa and Marda, Vidushi and Margetts, Helen and McDermid, John and Munga, Jane and Narayanan, Arvind and Nelson, Alondra and Neppel, Clara and Oh, Alice and Ramchurn, Gopal and Russell, Stuart and Schaake, Marietje and Schölkopf, Bernhard and Song, Dawn and Soto, Alvaro and Tiedrich, Lee and Varoquaux, Gaël and Yao, Andrew and Zhang, Ya-Qin and Albalawi, Fahad and Alserkal, Marwan and Ajala, Olubunmi and Avrin, Guillaume and Busch, Christian and Carvalho, André Carlos Ponce de Leon Ferreira de and Fox, Bronwyn and Gill, Amandeep Singh and Hatip, Ahmet Halit and Heikkilä, Juha and Jolly, Gill and Katzir, Ziv and Kitano, Hiroaki and Krüger, Antonio and Johnson, Chris and Khan, Saif M. and Lee, Kyoung Mu and Ligot, Dominic Vincent and Molchanovskyi, Oleksii and Monti, Andrea and Mwamanzi, Nusu and Nemer, Mona and Oliver, Nuria and Portillo, José Ramón López and Ravindran, Balaraman and Rivera, Raquel Pezoa and Riza, Hammam and Rugege, Crystal and Seoighe, Ciarán and Sheehan, Jerry and Sheikh, Haroon and Wong, Denise and Zeng, Yi},
    month = jan,
    year = {2025},
    note = {arXiv:2501.17805 [cs]},
}

@inproceedings{moskvoretskii_adaptive_2025,
    naddress = {Vienna, Austria},
    title = {Adaptive {Retrieval} {Without} {Self}-{Knowledge}? {Bringing} {Uncertainty} {Back} {Home}},
    isbn = {979-8-89176-251-0},
    shorttitle = {Adaptive {Retrieval} {Without} {Self}-{Knowledge}?},
    url = {https://aclanthology.org/2025.acl-long.319/},
    doi = {10.18653/v1/2025.acl-long.319},
    urldate = {2025-08-25},
    booktitle = {Proceedings of the 63rd {Annual} {Meeting} of the {Association} for {Computational} {Linguistics} ({Volume} 1: {Long} {Papers})},
    npublisher = {Association for Computational Linguistics},
    author = {Moskvoretskii, Viktor and Marina, Maria and Salnikov, Mikhail and Ivanov, Nikolay and Pletenev, Sergey and Galimzianova, Daria and Krayko, Nikita and Konovalov, Vasily and Nikishina, Irina and Panchenko, Alexander},
    neditor = {Che, Wanxiang and Nabende, Joyce and Shutova, Ekaterina and Pilehvar, Mohammad Taher},
    month = jul,
    year = {2025},
    pages = {6355--6384},
}

@inproceedings{yao_learning_2024,
    title = {Learning {From} {Correctness} {Without} {Prompting} {Makes} {LLM} {Efficient} {Reasoner}},
    url = {https://openreview.net/forum?id=dcbNzhVVQj#discussion},
    language = {en},
    urldate = {2026-03-20},
    author = {Yao, Yuxuan and Wu, Han and Guo, Zhijiang and Biyan, Zhou and Gao, Jiahui and Luo, Sichun and Hou, Hanxu and Fu, Xiaojin and Song, Linqi},
    month = aug,
    year = {2024},
    booktitle = {COLM},
}

@inproceedings{abbasi-yadkori_believe_2024,
    title = {To {Believe} or {Not} to {Believe} {Your} {LLM}: {Iterative} {Prompting} for {Estimating} {Epistemic} {Uncertainty}},
    shorttitle = {To {Believe} or {Not} to {Believe} {Your} {LLM}},
    nurl = {https://openreview.net/forum?id=k6iyUfwdI9&referrer=%5Bthe%20profile%20of%20Andr%C3%A1s%20Gy%C3%B6rgy%5D(%2Fprofile%3Fid%3D~Andr%C3%A1s_Gy%C3%B6rgy2)},
    language = {en},
    urldate = {2025-03-05},
    author = {Abbasi-Yadkori, Yasin and Kuzborskij, Ilja and György, András and Szepesvari, Csaba},
    month = nov,
    booktitle={NeurIPS},
    year = {2024},
}

@misc{yang_uncertainty-aware_2023,
    title = {Uncertainty-aware {Language} {Modeling} for {Selective} {Question} {Answering}},
    url = {http://arxiv.org/abs/2311.15451},
    doi = {10.48550/arXiv.2311.15451},
    urldate = {2026-03-20},
    publisher = {arXiv},
    author = {Yang, Qi and Ravikumar, Shreya and Schmitt-Ulms, Fynn and Lolla, Satvik and Demir, Ege and Elistratov, Iaroslav and Lavaee, Alex and Lolla, Sadhana and Ahmadi, Elaheh and Rus, Daniela and Amini, Alexander and Perez, Alejandro},
    month = nov,
    year = {2023},
    note = {arXiv:2311.15451 [cs]},
}

@inproceedings{chen_query-level_2025,
    title = {Query-{Level} {Uncertainty} in {Large} {Language} {Models}},
    url = {https://openreview.net/forum?id=11QZITAMUO},
    language = {en},
    urldate = {2026-03-20},
    author = {Chen, Lihu and Melo, Gerard de and Suchanek, Fabian M. and Varoquaux, Gaël},
    month = oct,
    year = {2026},
    booktitle = {ICLR},
}

@misc{zhao_survey_2025,
    title = {A {Survey} of {Large} {Language} {Models}},
    url = {http://arxiv.org/abs/2303.18223},
    doi = {10.48550/arXiv.2303.18223},
    urldate = {2025-03-19},
    publisher = {arXiv},
    author = {Zhao, Wayne Xin and Zhou, Kun and Li, Junyi and Tang, Tianyi and Wang, Xiaolei and Hou, Yupeng and Min, Yingqian and Zhang, Beichen and Zhang, Junjie and Dong, Zican and Du, Yifan and Yang, Chen and Chen, Yushuo and Chen, Zhipeng and Jiang, Jinhao and Ren, Ruiyang and Li, Yifan and Tang, Xinyu and Liu, Zikang and Liu, Peiyu and Nie, Jian-Yun and Wen, Ji-Rong},
    month = mar,
    year = {2025},
    note = {arXiv:2303.18223 [cs]},
}

@inproceedings{sahoo_comprehensive_2024,
    naddress = {Miami, Florida, USA},
    title = {A {Comprehensive} {Survey} of {Hallucination} in {Large} {Language}, {Image}, {Video} and {Audio} {Foundation} {Models}},
    url = {https://aclanthology.org/2024.findings-emnlp.685/},
    doi = {10.18653/v1/2024.findings-emnlp.685},
    urldate = {2025-01-22},
    booktitle = {Findings of the {Association} for {Computational} {Linguistics}: {EMNLP} 2024},
    npublisher = {Association for Computational Linguistics},
    author = {Sahoo, Pranab and Meharia, Prabhash and Ghosh, Akash and Saha, Sriparna and Jain, Vinija and Chadha, Aman},
    neditor = {Al-Onaizan, Yaser and Bansal, Mohit and Chen, Yun-Nung},
    month = nov,
    year = {2024},
    pages = {11709--11724},
}

@article{augenstein_factuality_2024,
    title = {Factuality challenges in the era of large language models and opportunities for fact-checking},
    volume = {6},
    copyright = {2024 Springer Nature Limited},
    issn = {2522-5839},
    url = {https://www.nature.com/articles/s42256-024-00881-z},
    doi = {10.1038/s42256-024-00881-z},
    language = {en},
    number = {8},
    urldate = {2026-03-20},
    journal = {Nature Machine Intelligence},
    author = {Augenstein, Isabelle and Baldwin, Timothy and Cha, Meeyoung and Chakraborty, Tanmoy and Ciampaglia, Giovanni Luca and Corney, David and DiResta, Renee and Ferrara, Emilio and Hale, Scott and Halevy, Alon and Hovy, Eduard and Ji, Heng and Menczer, Filippo and Miguez, Ruben and Nakov, Preslav and Scheufele, Dietram and Sharma, Shivam and Zagni, Giovanni},
    month = aug,
    year = {2024},
    nnote = {Publisher: Nature Publishing Group},
    pages = {852--863},
}

@inproceedings{tian_just_2023,
    naddress = {Singapore},
    title = {Just {Ask} for {Calibration}: {Strategies} for {Eliciting} {Calibrated} {Confidence} {Scores} from {Language} {Models} {Fine}-{Tuned} with {Human} {Feedback}},
    shorttitle = {Just {Ask} for {Calibration}},
    url = {https://aclanthology.org/2023.emnlp-main.330/},
    doi = {10.18653/v1/2023.emnlp-main.330},
    urldate = {2025-01-24},
    booktitle = {Proceedings of the {Conference} on {Empirical} {Methods} in {Natural} {Language} {Processing}},
    npublisher = {Association for Computational Linguistics},
    author = {Tian, Katherine and Mitchell, Eric and Zhou, Allan and Sharma, Archit and Rafailov, Rafael and Yao, Huaxiu and Finn, Chelsea and Manning, Christopher},
    neditor = {Bouamor, Houda and Pino, Juan and Bali, Kalika},
    month = dec,
    year = {2023},
    pages = {5433--5442},
}

@inproceedings{santilli_revisiting_2025,
    naddress = {Vienna, Austria},
    title = {Revisiting {Uncertainty} {Quantification} {Evaluation} in {Language} {Models}: {Spurious} {Interactions} with {Response} {Length} {Bias} {Results}},
    isbn = {979-8-89176-252-7},
    shorttitle = {Revisiting {Uncertainty} {Quantification} {Evaluation} in {Language} {Models}},
    url = {https://aclanthology.org/2025.acl-short.60/},
    doi = {10.18653/v1/2025.acl-short.60},
    urldate = {2025-08-25},
    booktitle = {Proceedings of the 63rd {Annual} {Meeting} of the {Association} for {Computational} {Linguistics} ({Volume} 2: {Short} {Papers})},
    npublisher = {Association for Computational Linguistics},
    author = {Santilli, Andrea and Golinski, Adam and Kirchhof, Michael and Danieli, Federico and Blaas, Arno and Xiong, Miao and Zappella, Luca and Williamson, Sinead},
    neditor = {Che, Wanxiang and Nabende, Joyce and Shutova, Ekaterina and Pilehvar, Mohammad Taher},
    month = jul,
    year = {2025},
    pages = {743--759},
}

@inproceedings{wang_ubench_2025,
    naddress = {Vienna, Austria},
    title = {{UBench}: {Benchmarking} {Uncertainty} in {Large} {Language} {Models} with {Multiple} {Choice} {Questions}},
    isbn = {979-8-89176-256-5},
    shorttitle = {{UBench}},
    url = {https://aclanthology.org/2025.findings-acl.423/},
    doi = {10.18653/v1/2025.findings-acl.423},
    urldate = {2025-10-16},
    booktitle = {Findings of the {Association} for {Computational} {Linguistics}},
    npublisher = {Association for Computational Linguistics},
    author = {Wang, Xunzhi and Zhang, Zhuowei and Chen, Gaonan and Li, Qiongyu and Luo, Bitong and Han, Zhixin and Wang, Haotian and Li, Zhiyu and Gao, Hang and Hu, Mengting},
    neditor = {Che, Wanxiang and Nabende, Joyce and Shutova, Ekaterina and Pilehvar, Mohammad Taher},
    month = jul,
    year = {2025},
    pages = {8076--8107},
}

@inproceedings{nikitin_kernel_2024,
    title = {Kernel {Language} {Entropy}: {Fine}-grained {Uncertainty} {Quantification} for {LLMs} from {Semantic} {Similarities}},
    shorttitle = {Kernel {Language} {Entropy}},
    url = {https://openreview.net/forum?id=j2wCrWmgMX},
    language = {en},
    urldate = {2025-07-01},
    booktitle = {Advances in {Neural} {Information} {Processing} {Systems}},
    author = {Nikitin, Alexander V. and Kossen, Jannik and Gal, Yarin and Marttinen, Pekka},
    month = nov,
    year = {2024},
}

@inproceedings{vashurin_cocoa_2025,
    title = {{CoCoA}: {A} {Minimum} {Bayes} {Risk} {Framework} {Bridging} {Confidence} and {Consistency} for {Uncertainty} {Quantification} in {LLMs}},
    shorttitle = {{CoCoA}},
    url = {https://openreview.net/forum?id=H1NGlLNaVC},
    language = {en},
    urldate = {2026-04-08},
    author = {Vashurin, Roman and Goloburda, Maiya and Ilina, Albina and Rubashevskii, Aleksandr and Nakov, Preslav and Shelmanov, Artem and Panov, Maxim},
    month = oct,
    booktitle = {NeurIPS},
    year = {2025},
}

@inproceedings{duan_shifting_2024,
    naddress = {Bangkok, Thailand},
    title = {Shifting {Attention} to {Relevance}: {Towards} the {Predictive} {Uncertainty} {Quantification} of {Free}-{Form} {Large} {Language} {Models}},
    shorttitle = {Shifting {Attention} to {Relevance}},
    url = {https://aclanthology.org/2024.acl-long.276/},
    doi = {10.18653/v1/2024.acl-long.276},
    urldate = {2025-07-03},
    booktitle = {Proceedings of the 62nd {Annual} {Meeting} of the {Association} for {Computational} {Linguistics} ({Volume} 1: {Long} {Papers})},
    publisher = {Association for Computational Linguistics},
    author = {Duan, Jinhao and Cheng, Hao and Wang, Shiqi and Zavalny, Alex and Wang, Chenan and Xu, Renjing and Kailkhura, Bhavya and Xu, Kaidi},
    neditor = {Ku, Lun-Wei and Martins, Andre and Srikumar, Vivek},
    month = aug,
    year = {2024},
    pages = {5050--5063},
}

@inproceedings{van_der_poel_mutual_2022,
    naddress = {Abu Dhabi, United Arab Emirates},
    title = {Mutual {Information} {Alleviates} {Hallucinations} in {Abstractive} {Summarization}},
    url = {https://aclanthology.org/2022.emnlp-main.399/},
    doi = {10.18653/v1/2022.emnlp-main.399},
    urldate = {2026-04-08},
    booktitle = {Proceedings of the {Conference} on {Empirical} {Methods} in {Natural} {Language} {Processing}},
    npublisher = {Association for Computational Linguistics},
    author = {van der Poel, Liam and Cotterell, Ryan and Meister, Clara},
    neditor = {Goldberg, Yoav and Kozareva, Zornitsa and Zhang, Yue},
    month = dec,
    year = {2022},
    pages = {5956--5965},
}

@inproceedings{sriramanan_llm-check_2024,
    title = {{LLM}-{Check}: {Investigating} {Detection} of {Hallucinations} in {Large} {Language} {Models}},
    shorttitle = {{LLM}-{Check}},
    url = {https://openreview.net/forum?id=LYx4w3CAgy},
    language = {en},
    urldate = {2025-01-15},
    booktitle = {NeurIPS},
    author = {Sriramanan, Gaurang and Bharti, Siddhant and Sadasivan, Vinu Sankar and Saha, Shoumik and Kattakinda, Priyatham and Feizi, Soheil},
    month = nov,
    year = {2024},
}

@article{vazhentsev_uncertainty-aware_2025,
    title = {Uncertainty-{Aware} {Attention} {Heads}: {Efficient} {Unsupervised} {Uncertainty} {Quantification} for {LLMs}},
    copyright = {Creative Commons Attribution 4.0 International},
    shorttitle = {Uncertainty-{Aware} {Attention} {Heads}},
    url = {https://arxiv.org/abs/2505.20045},
    doi = {10.48550/ARXIV.2505.20045},
    urldate = {2026-04-08},
    journal = {arXiv 2505.20045},
    author = {Vazhentsev, Artem and Rvanova, Lyudmila and Kuzmin, Gleb and Fadeeva, Ekaterina and Lazichny, Ivan and Panchenko, Alexander and Panov, Maxim and Baldwin, Timothy and Sachan, Mrinmaya and Nakov, Preslav and Shelmanov, Artem},
    year = {2025},
    nnote = {Version Number: 2},
}

@inproceedings{lin_contextualized_2024,
    naddress = {Miami, Florida, USA},
    title = {Contextualized {Sequence} {Likelihood}: {Enhanced} {Confidence} {Scores} for {Natural} {Language} {Generation}},
    shorttitle = {Contextualized {Sequence} {Likelihood}},
    url = {https://aclanthology.org/2024.emnlp-main.578/},
    doi = {10.18653/v1/2024.emnlp-main.578},
    urldate = {2025-01-20},
    booktitle = {Proceedings of the {Conference} on {Empirical} {Methods} in {Natural} {Language} {Processing}},
    npublisher = {Association for Computational Linguistics},
    author = {Lin, Zhen and Trivedi, Shubhendu and Sun, Jimeng},
    neditor = {Al-Onaizan, Yaser and Bansal, Mohit and Chen, Yun-Nung},
    month = nov,
    year = {2024},
    pages = {10351--10368},
}

@inproceedings{vazhentsev_hybrid_2023,
    naddress = {Toronto, Canada},
    title = {Hybrid {Uncertainty} {Quantification} for {Selective} {Text} {Classification} in {Ambiguous} {Tasks}},
    url = {https://aclanthology.org/2023.acl-long.652/},
    doi = {10.18653/v1/2023.acl-long.652},
    urldate = {2026-04-08},
    booktitle = {Proceedings of the 61st {Annual} {Meeting} of the {Association} for {Computational} {Linguistics} ({Volume} 1: {Long} {Papers})},
    npublisher = {Association for Computational Linguistics},
    author = {Vazhentsev, Artem and Kuzmin, Gleb and Tsvigun, Akim and Panchenko, Alexander and Panov, Maxim and Burtsev, Mikhail and Shelmanov, Artem},
    neditor = {Rogers, Anna and Boyd-Graber, Jordan and Okazaki, Naoaki},
    month = jul,
    year = {2023},
    pages = {11659--11681},
}

@article{bouchard_uqlm_2026,
    title = {{UQLM}: {A} {Python} {Package} for {Uncertainty} {Quantification} in {Large} {Language} {Models}},
    volume = {27},
    issn = {1533-7928},
    shorttitle = {{UQLM}},
    url = {http://jmlr.org/papers/v27/25-1557.html},
    number = {13},
    urldate = {2026-04-08},
    journal = {Journal of Machine Learning Research},
    author = {Bouchard, Dylan and Chauhan, Mohit Singh and Skarbrevik, David and Ra, Ho-Kyeong and Bajaj, Viren and Ahmad, Zeya},
    year = {2026},
    pages = {1--10},
}

@inproceedings{fadeeva_fact-checking_2024,
    naddress = {Bangkok, Thailand},
    title = {Fact-{Checking} the {Output} of {Large} {Language} {Models} via {Token}-{Level} {Uncertainty} {Quantification}},
    url = {https://aclanthology.org/2024.findings-acl.558/},
    doi = {10.18653/v1/2024.findings-acl.558},
    urldate = {2025-03-12},
    booktitle = {Findings of the {Association} for {Computational} {Linguistics}},
    npublisher = {Association for Computational Linguistics},
    author = {Fadeeva, Ekaterina and Rubashevskii, Aleksandr and Shelmanov, Artem and Petrakov, Sergey and Li, Haonan and Mubarak, Hamdy and Tsymbalov, Evgenii and Kuzmin, Gleb and Panchenko, Alexander and Baldwin, Timothy and Nakov, Preslav and Panov, Maxim},
    neditor = {Ku, Lun-Wei and Martins, Andre and Srikumar, Vivek},
    month = aug,
    year = {2024},
    pages = {9367--9385},
}

@inproceedings{chen_inside_2023,
    title = {{INSIDE}: {LLMs}' {Internal} {States} {Retain} the {Power} of {Hallucination} {Detection}},
    shorttitle = {{INSIDE}},
    url = {https://openreview.net/forum?id=Zj12nzlQbz},
    language = {en},
    urldate = {2025-01-20},
    booktitle = {International {Conference} on {Learning} {Representations}},
    author = {Chen, Chao and Liu, Kai and Chen, Ze and Gu, Yi and Wu, Yue and Tao, Mingyuan and Fu, Zhihang and Ye, Jieping},
    month = oct,
    year = {2023},
}

@misc{ren_simple_2021,
    title = {A {Simple} {Fix} to {Mahalanobis} {Distance} for {Improving} {Near}-{OOD} {Detection}},
    url = {http://arxiv.org/abs/2106.09022},
    doi = {10.48550/arXiv.2106.09022},
    urldate = {2026-04-08},
    publisher = {arXiv},
    author = {Ren, Jie and Fort, Stanislav and Liu, Jeremiah and Roy, Abhijit Guha and Padhy, Shreyas and Lakshminarayanan, Balaji},
    month = jun,
    year = {2021},
    note = {arXiv:2106.09022 [cs]},
}

@inproceedings{yoo_detection_2022,
    naddress = {Dublin, Ireland},
    title = {Detection of {Adversarial} {Examples} in {Text} {Classification}: {Benchmark} and {Baseline} via {Robust} {Density} {Estimation}},
    shorttitle = {Detection of {Adversarial} {Examples} in {Text} {Classification}},
    url = {https://aclanthology.org/2022.findings-acl.289/},
    doi = {10.18653/v1/2022.findings-acl.289},
    urldate = {2026-04-08},
    booktitle = {Findings of the {Association} for {Computational} {Linguistics}},
    npublisher = {Association for Computational Linguistics},
    author = {Yoo, KiYoon and Kim, Jangho and Jang, Jiho and Kwak, Nojun},
    neditor = {Muresan, Smaranda and Nakov, Preslav and Villavicencio, Aline},
    month = may,
    year = {2022},
    pages = {3656--3672},
}

@inproceedings{
he2021deberta,
title={{DeBERTa}: Decoding-enhanced {BERT} with Disentangled Attention},
author={Pengcheng He and Xiaodong Liu and Jianfeng Gao and Weizhu Chen},
booktitle={International Conference on Learning Representations},
year={2021},
url={https://openreview.net/forum?id=XPZIaotutsD}
}

@inproceedings{lin-2004-rouge,
    title = "{ROUGE}: A Package for Automatic Evaluation of Summaries",
    author = "Lin, Chin-Yew",
    booktitle = "Text Summarization Branches Out",
    month = jul,
    year = "2004",
    naddress = "Barcelona, Spain",
    publisher = "Association for Computational Linguistics",
    url = "https://aclanthology.org/W04-1013/",
    pages = "74--81"
}

@inproceedings{papineni-etal-2002-bleu,
    title = "{BLEU}: a Method for Automatic Evaluation of Machine Translation",
    author = "Papineni, Kishore  and
      Roukos, Salim  and
      Ward, Todd  and
      Zhu, Wei-Jing",
    neditor = "Isabelle, Pierre  and
      Charniak, Eugene  and
      Lin, Dekang",
    booktitle = "Proceedings of the 40th Annual Meeting of the Association for Computational Linguistics",
    month = jul,
    year = "2002",
    naddress = "Philadelphia, Pennsylvania, USA",
    npublisher = "Association for Computational Linguistics",
    url = "https://aclanthology.org/P02-1040/",
    doi = "10.3115/1073083.1073135",
    pages = "311--318"
}

@article{lin_generating_2024,
    title = {Generating with {Confidence}: {Uncertainty} {Quantification} for {Black}-box {Large} {Language} {Models}},
    issn = {2835-8856},
    shorttitle = {Generating with {Confidence}},
    url = {https://openreview.net/forum?id=DWkJCSxKU5},
    language = {en},
    urldate = {2026-04-23},
    journal = {Transactions on Machine Learning Research},
    author = {Lin, Zhen and Trivedi, Shubhendu and Sun, Jimeng},
    month = feb,
    year = {2024},
}

@article{farquhar_detecting_2024,
    title = {Detecting hallucinations in large language models using semantic entropy},
    volume = {630},
    copyright = {2024 The Author(s)},
    issn = {1476-4687},
    url = {https://www.nature.com/articles/s41586-024-07421-0},
    doi = {10.1038/s41586-024-07421-0},
    language = {en},
    number = {8017},
    urldate = {2025-01-15},
    journal = {Nature},
    author = {Farquhar, Sebastian and Kossen, Jannik and Kuhn, Lorenz and Gal, Yarin},
    month = jun,
    year = {2024},
    nnote = {Publisher: Nature Publishing Group},
    pages = {625--630},
}

@book{cover_elements_2001,
    address = {Hoboken, NJ},
    title = {Elements of information theory},
    isbn = {978-0-471-06259-2 978-0-471-20061-1},
    language = {en},
    publisher = {Wiley-Interscience},
    author = {Cover, Thomas M. and Thomas, Joy A.},
    year = {2001},
}

@article{ji_survey_2023,
    title = {Survey of {Hallucination} in {Natural} {Language} {Generation}},
    volume = {55},
    issn = {0360-0300, 1557-7341},
    url = {http://arxiv.org/abs/2202.03629},
    doi = {10.1145/3571730},
    number = {12},
    urldate = {2025-01-22},
    journal = {ACM Computing Surveys},
    author = {Ji, Ziwei and Lee, Nayeon and Frieske, Rita and Yu, Tiezheng and Su, Dan and Xu, Yan and Ishii, Etsuko and Bang, Yejin and Chen, Delong and Dai, Wenliang and Chan, Ho Shu and Madotto, Andrea and Fung, Pascale},
    month = dec,
    year = {2023},
    pages = {1--38},
}

@misc{wu_uncertaintyzoo_2025,
    title = {{UncertaintyZoo}: {A} {Unified} {Toolkit} for {Quantifying} {Predictive} {Uncertainty} in {Deep} {Learning} {Systems}},
    shorttitle = {{UncertaintyZoo}},
    url = {http://arxiv.org/abs/2512.06406},
    doi = {10.48550/arXiv.2512.06406},
    urldate = {2026-04-28},
    publisher = {arXiv},
    author = {Wu, Xianzong and Li, Xiaohong and Quan, Lili and Hu, Qiang},
    month = dec,
    year = {2025},
    note = {arXiv:2512.06406 [cs]},
}

@inproceedings{ielanskyi_addressing_2025,
    title = {Addressing {Pitfalls} in the {Evaluation} of {Uncertainty} {Estimation} {Methods} for {Natural} {Language} {Generation}},
    url = {https://openreview.net/forum?id=hbIpwrdfE1},
    language = {en},
    urldate = {2025-04-09},
    booktitle = {{ICLR} {Workshop}: {Quantify} {Uncertainty} and {Hallucination} in {Foundation} {Models}: {The} {Next} {Frontier} in {Reliable} {AI}},
    author = {Ielanskyi, Mykyta and Schweighofer, Kajetan and Aichberger, Lukas and Hochreiter, Sepp},
    month = mar,
    year = {2025},
}

@misc{devic_calibration_2025,
    title = {From {Calibration} to {Collaboration}: {LLM} {Uncertainty} {Quantification} {Should} {Be} {More} {Human}-{Centered}},
    shorttitle = {From {Calibration} to {Collaboration}},
    url = {http://arxiv.org/abs/2506.07461},
    doi = {10.48550/arXiv.2506.07461},
    urldate = {2026-04-28},
    publisher = {arXiv},
    author = {Devic, Siddartha and Srinivasan, Tejas and Thomason, Jesse and Neiswanger, Willie and Sharan, Vatsal},
    month = jun,
    year = {2025},
    note = {arXiv:2506.07461 [cs]
version: 1},
}

@misc{tomov_illusion_2026,
    title = {The {Illusion} of {Certainty}: {Uncertainty} {Quantification} for {LLMs} {Fails} under {Ambiguity}},
    shorttitle = {The {Illusion} of {Certainty}},
    url = {http://arxiv.org/abs/2511.04418},
    doi = {10.48550/arXiv.2511.04418},
    urldate = {2026-02-10},
    publisher = {arXiv},
    author = {Tomov, Tim and Fuchsgruber, Dominik and Wollschläger, Tom and Günnemann, Stephan},
    month = jan,
    year = {2026},
    note = {arXiv:2511.04418 [cs]},
}

@article{huang_survey_2024,
    title = {A {Survey} on {Hallucination} in {Large} {Language} {Models}: {Principles}, {Taxonomy}, {Challenges}, and {Open} {Questions}},
    issn = {1046-8188},
    shorttitle = {A {Survey} on {Hallucination} in {Large} {Language} {Models}},
    url = {https://dl.acm.org/doi/10.1145/3703155},
    doi = {10.1145/3703155},
    urldate = {2025-01-17},
    journal = {ACM Transactions on Information Systems},
    author = {Huang, Lei and Yu, Weijiang and Ma, Weitao and Zhong, Weihong and Feng, Zhangyin and Wang, Haotian and Chen, Qianglong and Peng, Weihua and Feng, Xiaocheng and Qin, Bing and Liu, Ting},
    month = nov,
    year = {2024},
}

@article{kwiatkowski_natural_2019,
    title = {Natural {Questions}: {A} {Benchmark} for {Question} {Answering} {Research}},
    volume = {7},
    shorttitle = {Natural {Questions}},
    url = {https://aclanthology.org/Q19-1026/},
    doi = {10.1162/tacl_a_00276},
    urldate = {2025-03-05},
    journal = {Transactions of the Association for Computational Linguistics},
    author = {Kwiatkowski, Tom and Palomaki, Jennimaria and Redfield, Olivia and Collins, Michael and Parikh, Ankur and Alberti, Chris and Epstein, Danielle and Polosukhin, Illia and Devlin, Jacob and Lee, Kenton and Toutanova, Kristina and Jones, Llion and Kelcey, Matthew and Chang, Ming-Wei and Dai, Andrew M. and Uszkoreit, Jakob and Le, Quoc and Petrov, Slav},
    neditor = {Lee, Lillian and Johnson, Mark and Roark, Brian and Nenkova, Ani},
    year = {2019},
    nnote = {Place: Cambridge, MA
Publisher: MIT Press},
    pages = {452--466},
}

@inproceedings{joshi_triviaqa_2017,
    naddress = {Vancouver, Canada},
    title = {{TriviaQA}: {A} {Large} {Scale} {Distantly} {Supervised} {Challenge} {Dataset} for {Reading} {Comprehension}},
    shorttitle = {{TriviaQA}},
    url = {https://aclanthology.org/P17-1147/},
    doi = {10.18653/v1/P17-1147},
    urldate = {2026-04-28},
    booktitle = {Proceedings of the 55th {Annual} {Meeting} of the {Association} for {Computational} {Linguistics} ({Volume} 1: {Long} {Papers})},
    npublisher = {Association for Computational Linguistics},
    author = {Joshi, Mandar and Choi, Eunsol and Weld, Daniel and Zettlemoyer, Luke},
    neditor = {Barzilay, Regina and Kan, Min-Yen},
    month = jul,
    year = {2017},
    pages = {1601--1611},
}

@misc{wei_measuring_2024,
    title = {Measuring short-form factuality in large language models},
    url = {http://arxiv.org/abs/2411.04368},
    doi = {10.48550/arXiv.2411.04368},
    urldate = {2025-03-04},
    publisher = {arXiv},
    author = {Wei, Jason and Karina, Nguyen and Chung, Hyung Won and Jiao, Yunxin Joy and Papay, Spencer and Glaese, Amelia and Schulman, John and Fedus, William},
    month = nov,
    year = {2024},
    note = {arXiv:2411.04368 [cs]},
}

@inproceedings{cruz_evaluating_2024,
    naddress = {Red Hook, NY, USA},
    nseries = {{NIPS} '24},
    title = {Evaluating language models as risk scores},
    volume = {37},
    isbn = {979-8-3313-1438-5},
    urldate = {2026-04-29},
    booktitle = {Proceedings of the 38th {International} {Conference} on {Neural} {Information} {Processing} {Systems}},
    npublisher = {Curran Associates Inc.},
    author = {Cruz, André F. and Hardt, Moritz and Mendler-Dünner, Celestine},
    year = {2024},
    pages = {97378--97407},
}

@article{williams_learning_1989,
    title = {A {Learning} {Algorithm} for {Continually} {Running} {Fully} {Recurrent} {Neural} {Networks}},
    volume = {1},
    issn = {0899-7667, 1530-888X},
    url = {https://direct.mit.edu/neco/article/1/2/270-280/5490},
    doi = {10.1162/neco.1989.1.2.270},
    language = {en},
    number = {2},
    urldate = {2026-05-06},
    journal = {Neural Computation},
    author = {Williams, Ronald J. and Zipser, David},
    month = jun,
    year = {1989},
    pages = {270--280},
}

@inproceedings{davis2006relationship,
    title = {The Relationship Between Precision-Recall and {ROC} Curves},
    author = {Davis, Jesse and Goadrich, Mark},
    booktitle = {Proceedings of the 23rd International Conference on Machine Learning},
    pages = {233--240},
    year = {2006},
    publisher = {ACM},
    doi = {10.1145/1143844.1143874}
}
